\documentclass[]{template}

\usepackage[utf8]{inputenc}             % allow utf-8 input
\usepackage[T1]{fontenc}                % use 8-bit T1 fonts
\usepackage{url}                        % simple URL typesetting
\usepackage{booktabs}                   % professional-quality tables
\usepackage{multirow}
\usepackage{colortbl}
\usepackage{multicol}
\usepackage{amsfonts}                   % blackboard math symbols
\usepackage{nicefrac}                   % compact symbols for 1/2, etc.
\usepackage{microtype}                  % microtypography
\usepackage[dvipsnames]{xcolor}         % colors
\usepackage{tocloft}
\usepackage{setspace}

\usepackage{latexsym}

\usepackage{graphicx}
\usepackage{float}
\usepackage{subcaption}
\usepackage{wrapfig}
\usepackage{lipsum}

\usepackage{adjustbox}
\usepackage{makecell}

\usepackage{bm}

\usepackage{tabularx} 
\usepackage{ragged2e} 
\newcolumntype{L}{>{\RaggedRight\hangafter=1\hangindent=0em}X}

\usepackage{enumitem}

\usepackage{amsmath}
\usepackage{amssymb}
\usepackage{mathtools}
\usepackage{amsthm}
\usepackage{arydshln}

\setboolean{logo}{true}    % 设为 true 表示显示 logo；设为 false 则不显示

\usepackage[linesnumbered,ruled,vlined]{algorithm2e}

\hypersetup{
    colorlinks=true,
    linkcolor=red,
    citecolor=Cerulean,
    filecolor=magenta,      
    urlcolor=magenta,
}

\usepackage[capitalize,noabbrev]{cleveref}
\crefname{section}{§}{§§}
\Crefname{section}{§}{§§}

\usepackage{calligra}
\DeclareMathAlphabet{\mathcalligra}{T1}{calligra}{m}{n}

\usepackage{pifont}
\newcommand{\cmark}{\ding{51}} % ✓
\newcommand{\xmark}{\ding{55}} % ✗

\theoremstyle{plain}

\theoremstyle{definition}

\theoremstyle{remark}

\renewcommand{\paragraph}[1]{\vspace{1mm}\noindent\textbf{#1}}

\DeclareCaptionLabelFormat{cont}{#1~#2\alph{ContinuedFloat}}
\usepackage[most]{tcolorbox}
\tcbset{
  promptbox/.style={
    top=10pt,
    colback=lightgray!20,
    colframe=Black,
    colbacktitle=NavyBlue,
    enhanced,
    center,
    attach boxed title to top center={yshift=-0.1in,xshift=0.0in},
    boxed title style={boxrule=0pt,colframe=white,},
  }
}
\newtcolorbox{promptbox}[2][]{promptbox, title=#2,#1}
\tcbset{
  takeawaybox/.style={
    top=10pt,
    colback=lightgray!20,
    colframe=Black,
    colbacktitle=BurntOrange,
    enhanced,
    center,
    attach boxed title to top center={yshift=-0.1in,xshift=0.0in},
    boxed title style={boxrule=0pt,colframe=white,},
  }
}
\newtcolorbox{takeawaybox}[2][]{takeawaybox, title=#2,#1}
\tcbset{
  observationbox/.style={
    top=10pt,
    colback=lightgray!20,
    colframe=Black,
    colbacktitle=YellowGreen,
    enhanced,
    center,
    attach boxed title to top center={yshift=-0.1in,xshift=0.0in},
    boxed title style={boxrule=0pt,colframe=white,},
  }
}
\newtcolorbox{observationbox}[2][]{observationbox, title=#2,#1}

\usepackage{xspace}

\newcommand\blfootnote[1]{%
  \begingroup
  \renewcommand\thefootnote{}\footnote{#1}%
  \addtocounter{footnote}{-1}%
  \endgroup
}

\usepackage{twemojis}
\usepackage{fontawesome}
\newcommand{\huggingface}{\raisebox{-1.5pt}{\includegraphics[height=1.05em]{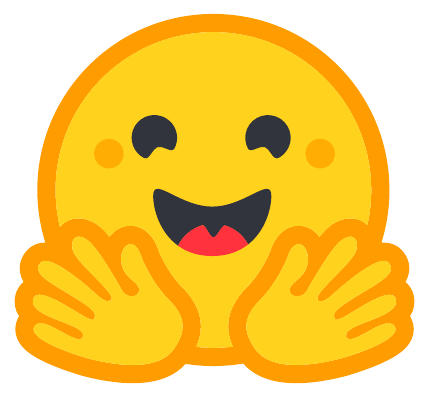}}\xspace}

\newcommand{\github}{\raisebox{-1.5pt}{\includegraphics[height=1.05em]{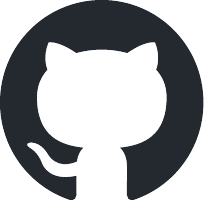}}\xspace}

\usepackage{CJK}

\usepackage{duckuments} 
\usepackage{marvosym}
\usepackage{authblk}

\newcommand{\dataset}{ScaleEdit\xspace}
\newcommand{\method}{ScaleEditor\xspace}

\input{preamble}

\title{ScaleEdit-12M: Scaling Open-Source Image Editing Data Generation via Multi-Agent Framework}

\author[1,2*]{Guanzhou Chen}
\author[1,2*]{Erfei Cui}
\author[3,2*$^\dagger$]{Changyao Tian}
\author[4,2*]{Danni Yang}

\author[5,2]{Ganlin Yang}

\author[2]{Yu Qiao}
\author[3]{Hongsheng Li}
\author[6]{Gen Luo}
\author[2\textsuperscript{\Letter}]{Hongjie Zhang}

\affil[1]{Shanghai Jiao Tong University}
\affil[2]{Shanghai AI Laboratory}
\affil[3]{CUHK MMLab}
\affil[4]{Fudan University}

\affil[5]{University of Science and Technology of China}
\affil[6]{Xiamen University}
\begin{abstract}
Instruction-based image editing has emerged as a key capability for unified multimodal models (UMMs), yet constructing large-scale, diverse, and high-quality editing datasets without costly proprietary APIs remains challenging.  Previous image editing datasets either rely on closed-source models for annotation, which prevents cost-effective scaling, or employ fixed synthetic editing pipelines, which suffer from limited quality and generalizability. 
To address these challenges, we propose \textbf{\method}, a fully open-source hierarchical multi-agent framework for end-to-end construction of large-scale, high-quality image editing datasets. Our pipeline consists of three key components: source image expansion with world-knowledge infusion, adaptive multi-agent editing instruction-image synthesis, and a task-aware data quality verification mechanism. 
Using \method, we curate \textbf{\dataset-12M}, the largest open-source image editing dataset to date, spanning 23 task families across diverse real and synthetic domains.
Fine-tuning UniWorld-V1 and Bagel on \dataset yields consistent gains, improving performance by up to 10.4\% on ImgEdit and 35.1\% on GEdit for general editing benchmarks and by up to 150.0\% on RISE and 26.5\% on KRIS-Bench for knowledge-infused benchmarks.
These results demonstrate that open-source, agentic pipelines can approach commercial-grade data quality while retaining cost-effectiveness and scalability.
Both the framework and dataset will be open-sourced.
\end{abstract}

\begin{document}

\blfootnote{* Equal Contribution. \textsuperscript{\Letter} Corresponding Author.  $^\dagger$ Project Lead. This work was done when Changyao Tian was an intern at Shanghai AI Laboratory. }

\maketitle

\begin{center}
    \renewcommand{\arraystretch}{1.5}
    \vspace{1em}
    \begin{tabular}{rll}
        \github{} & \textbf{GitHub Repo} & \url{https://github.com/gzchen4ai/ScaleEdit-12M} \\
        \huggingface{} & \textbf{HuggingFace Dataset} & \url{https://huggingface.co/datasets/InternVL-U/ScaleEdit-12M} \\
    \end{tabular}
\end{center}

\clearpage

\begin{figure*}[t!]
  \centering
   \includegraphics[width=0.98\linewidth]{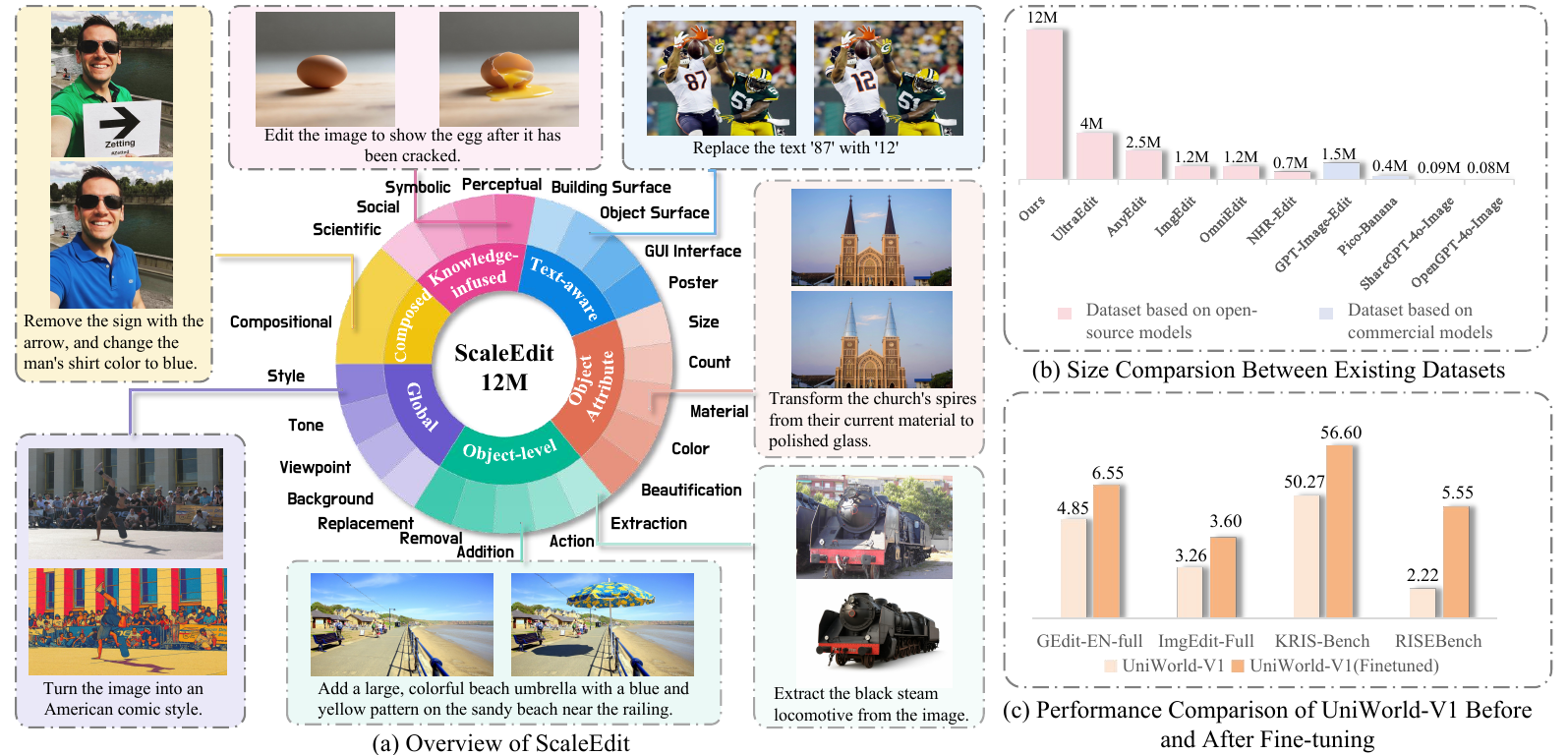}
   \caption{
  \textbf{Overview of \dataset.}
  (a) Examples of instruction-based image editing in \dataset, covering diverse local and global editing types. The central wheel illustrates the taxonomy of editing categories supported by \dataset. 
  (b) Size comparison between \dataset and existing instruction-based image editing datasets constructed from open-source and commercial models.
  (c) Performance of UniWorld-V1~\cite{lin2025uniworld} before and after fine-tuning on \dataset across multiple benchmarks (GEdit-EN-full~\cite{liu2025step1x}, ImgEdit-Full~\cite{ye2025imgedit}, KRIS-Bench~\cite{wu2025kris}, RISEBench~\cite{zhao2025envisioning}), where \dataset consistently brings significant improvements.
   }
   \label{fig:intro_v1}
\end{figure*}

\section{Introduction}

Instruction-based image editing has become a core capability of Unified Multimodal Models (UMMs)~\cite{lin2025uniworld, deng2025bagel, xiao2024omnigen}, enabling models to interpret natural-language instructions and perform precise edits in an end-to-end fashion~\cite{shi2024seededit}. Recent commercial systems such as GPT-4o-Image~\cite{openai2025gpt4oimage} and Nano-Banana~\cite{google2025nanobanana} demonstrate strong instruction-following and visual consistency on complex edits, pushing image editing from eye-catching demos toward production-grade tools for real-world applications.

Inspired by this, the community has introduced numerous image editing datasets and generation pipelines~\cite{zhao2024ultraedit,yu2025anyedit,wei2024omniedit}, aiming to enhance the editing capabilities of open-source UMMs. A common approach applies a predefined compositional synthetic pipeline with fixed editing operators (\eg, mask-guided inpainting, background replacement, style transfer) to large image collections. While scalable, this strategy biases datasets toward narrow edit types and often introduces noise, artifacts, and text-image misalignment~\cite{chen2025instruct,yu2025anyedit}. 
Another line of work directly queries leading proprietary commercial models (\eg, GPT-4o-Image and Nano-Banana) to synthesize high-quality image editing samples~\cite{wang2025gpt,chen2025opengpt,qian2025pico}. While effective, this strategy quickly becomes economically prohibitive as the dataset scale grows. Such limitations naturally raise the following question: \textit{Is it possible to build large-scale, diverse, and high-quality image editing datasets using open-source and cost-effective agentic toolkits?}

To answer this question, we first revisit why existing open-source editing corpora still lag behind datasets synthesized via frontier commercial image editors in terms of diversity and quality.
First, their source images often come from a narrow domain, restricted to a limited set of specific categories, or weakly curated synthetic collections, with limited coverage of real-world scenes, styles, and object compositions. Second, many current pipelines rely on rigid edit templates or rule-based instruction generation, which constrains the diversity and semantic richness of instruction-image pairs and hinders content-adaptive editing behavior for each input image. 
Third, their simple heuristic filters inadequately detect misalignment and artifacts at scale. 
These observations suggest that achieving GPT-level editing data in a purely open-source regime requires simultaneously expanding the source image distribution, adopting flexible agentic editing synthesis pipelines, and deploying a multi-dimensional quality verification mechanism.

Driven by these observations, we propose \textbf{\method}, a novel hierarchical multi-agent framework built on open-source toolkits for synthesizing large-scale, high-quality image editing datasets. \method comprises three workflows: (1) \textit{source image expansion} with world-knowledge infusion, employing web-search retrieval, captioning, and text-to-image agents to diversify the image pool; (2) \textit{adaptive multi-agent editing synthesis}, routing each image to appropriate editing tasks and workflows, with corresponding specialized agents to synthesize the editing instruction and edited image; and (3) \textit{task-aware quality verification}, assessing samples across multiple dimensions using specialized agents for different tasks. The framework enables efficient construction of large-scale editing datasets from a limited source pool.

Based on \method, we curate a large-scale, high-quality image-editing dataset named \textbf{\dataset}. As illustrated in \cref{fig:intro_v1}, \dataset comprises 12 million editing samples across 23 editing tasks, covering diverse visual domains including natural landscapes, urban environments, and human-centric daily scenes. To the best of our knowledge, \dataset is the largest instruction-based image editing dataset to date. 

To validate the effectiveness and generality of \dataset, we finetune two representative UMMs, UniWorld-V1~\cite{lin2025uniworld} and Bagel~\cite{deng2025bagel}, and evaluate them across multiple benchmarks. On general instruction-based editing benchmarks, including GEdit~\cite{liu2025step1x} and ImgEdit~\cite{ye2025imgedit}, the finetuned UniWorld-V1 achieves substantial improvements of 35.1\% and 10.4\% over its original baseline, respectively, while Bagel also delivers clear gains of 10.0\% and 7.8\%.
On knowledge-infused editing benchmarks such as RISE~\cite{zhao2025envisioning} and KRIS-Bench~\cite{wu2025kris}, both models obtain remarkable performance leaps: UniWorld-V1 improves by 150.0\% and 12.6\%, while Bagel gains 23.0\% and 26.5\%, respectively. 
Notably, both fine-tuned models consistently outperform counterparts trained on other open-source datasets, validating the immense value of \dataset and the efficacy of our \method framework.

In summary, our contributions are threefold:
\begin{itemize}
    \item We present \method, a fully open-source, multi-agent framework tailored for the cost-effective construction of large-scale, high-quality image editing datasets. It seamlessly integrates source image expansion, adaptive instruction-image synthesis, and rigorous multi-dimensional quality verification.    
    \item We introduce \dataset-12M, the largest high-quality, open-source image editing dataset to date. Comprising 12 million rigorously verified instruction-image pairs, it encompasses a wide spectrum of local and global editing tasks across diverse real and synthetic visual domains.
    \item We demonstrate the broad generalization of \dataset by fine-tuning leading foundation models (\eg, UniWorld-V1 and Bagel). The resulting models consistently surpass those trained on other open-source datasets across diverse benchmarks, proving that our open-source pipeline can rival commercial APIs.
\end{itemize}

\section{Related Works}
\paragraph{Text-Guided Image Editing Models.} 
With the advent of large-scale diffusion models~\cite{rombach2022high, podell2023sdxl}, text-guided image editing task has been extensively explored in recent years~\cite{liu2020open, ling2021editgan, zhang2023adding, crowson2022vqgan, huihq, wasserman2025paint,labs2025flux,wu2025qwen}.
InstructPix2Pix~\cite{brooks2023instructpix2pix} pioneered this direction by fine-tuning Stable Diffusion~\cite{rombach2022high} on an instruction-based image editing dataset. 
Building on this, UltraEdit~\cite{zhao2024ultraedit} and MagicBrush~\cite{zhang2024magicbrushmanuallyannotateddataset} focus on improving the quality and diversity of training datasets, while OmniEdit~\cite{wei2024omniedit} enhances generalization across tasks by incorporating supervision from multiple specialist models. 
Further, AnyEdit~\cite{yu2024anyedit} and ImgEdit~\cite{ye2025imgedit} enhance editing capabilities through broader coverage of complex editing tasks. 
Recent advancements in image editing~\cite{labs2025flux, wu2025qwen, liu2025step1x}, fueled by large-scale, high-quality datasets and represented by models such as Step1X-Edit~\cite{liu2025step1x} and InternVL-U~\cite{tian2026internvl}, have substantially improved editing fidelity.
However, despite these gains, existing models still struggle to integrate broad world knowledge and follow instructions reliably, highlighting the need for more high-quality, knowledge-rich editing datasets.

\paragraph{Instruction-Based Image Editing Datasets.}
Recent instruction-based image editing datasets show a clear trend toward scaling~\cite{ma2025x2edit,kuprashevich2025nohumansrequired,ge2024seeddataedittechnicalreporthybrid,wang2025gpt}. Early manually curated sets, such as MagicBrush~\cite{zhang2024magicbrushmanuallyannotateddataset}, have evolved into million-scale collections like OmniEdit~\cite{wei2024omniedit}, AnyEdit~\cite{yu2024anyedit}, ImgEdit~\cite{ye2025imgedit}, and UltraEdit~\cite{zhao2024ultraedit}. Recently, closed-source commercial models have enabled the synthesis of high-quality datasets, such as OpenGPT-4o-Image~\cite{openai2025gpt4oimage} (40K edit pairs), ShareGPT-4o-Image~\cite{chen2025sharegpt} (46K edit pairs), Pico-Banana-400K~\cite{qian2025pico}, and Nano-consistent-150K~\cite{ye2025echo}, yet their scale remains limited. Despite these advances, the community still lacks large-scale (\eg, 10M-level), high-quality datasets enriched with broad world knowledge, which is crucial for building more reliable and capable instruction-based image editing systems.
\begin{figure*}[!t]
  \centering
   \includegraphics[width=\linewidth]{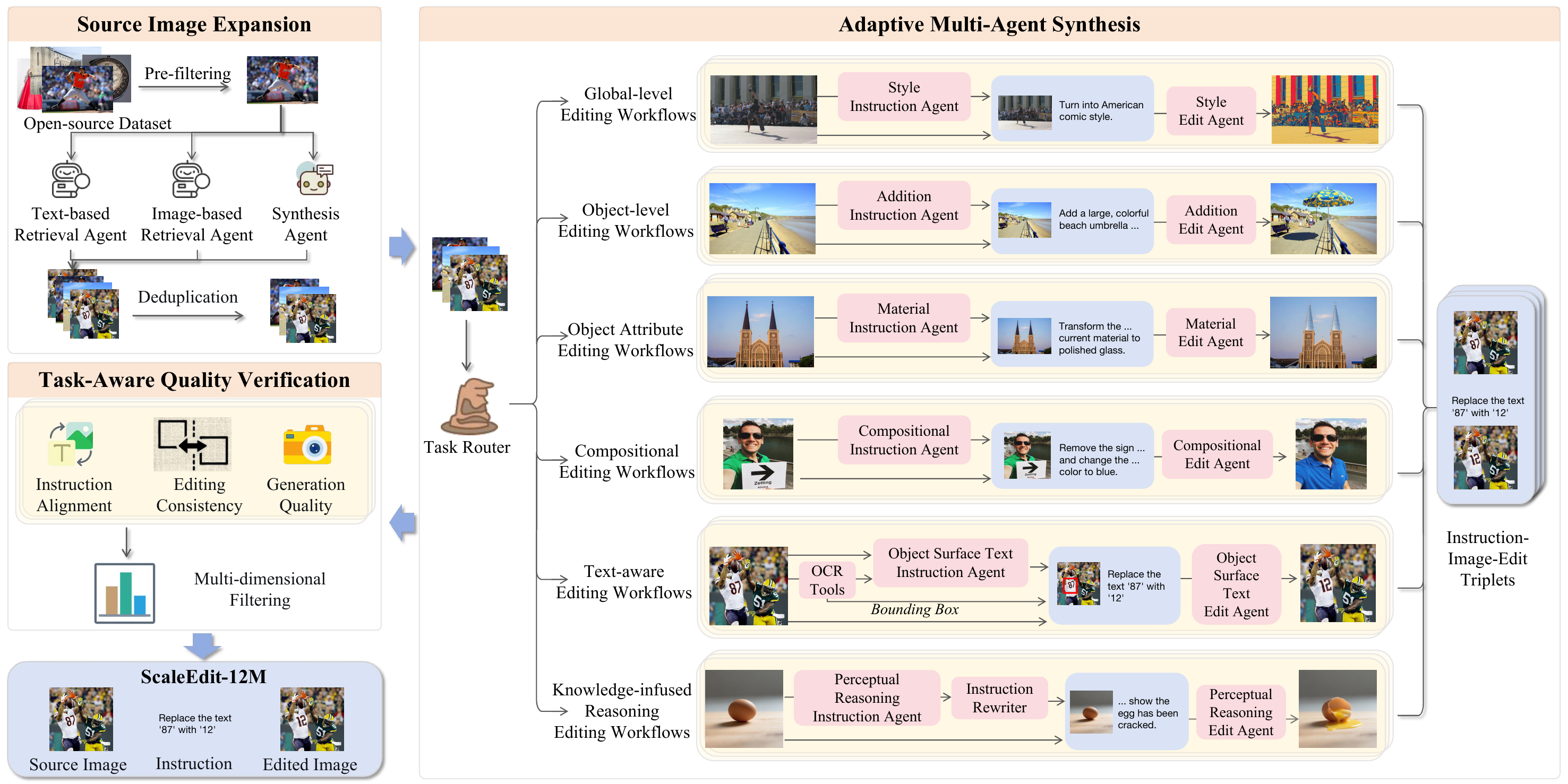}

   \caption{\textbf{Overview of \method.} The framework consists of three main stages: (1) Source Image Expansion, which retrieves and synthesizes diverse images while filtering and deduplicating noisy samples; (2) Adaptive Multi-Agent Synthesis, where a task router dispatches each image to appropriate editing workflows, with specialized agents to synthesize corresponding editing instruction and images; and (3) Task-Aware Quality Verification, which assesses instruction alignment, editing consistency, and generation quality. These high-quality instruction-image-edit triplets form our \dataset.}
   \label{fig:fig2}
\end{figure*}

\section{\method}
\label{sec:method}
As shown in Fig.~\ref{fig:fig2}, \method decomposes large-scale editing data construction into three components: \textit{Source Image Expansion}, \textit{Adaptive Multi-Agent Synthesis}, and \textit{Task-Aware Quality Verification}.

\subsection{Source Image Expansion}
\label{sec:source_image_expansion}
We first collect images from several open-source datasets, including COCO~\cite{lin2014microsoft}, OpenImages~\cite{kuznetsova2020openimages}, SA-1B~\cite{kirillov2023segment}, \etc. 
During dataset curation, we conduct rule-based pre-filtering, retaining images with a shorter side exceeding 512 pixels and an aspect ratio between 0.5 and 2.
To further enrich the diversity and domain coverage of the source image pool, we design a world-knowledge-enhanced expansion workflow with multiple branches: retrieval-based and synthesis-based.

\noindent\textbf{Retrieval-Based Expansion.} Beyond standard open-source image datasets, we further incorporate web-scale visual knowledge via large-scale search engines. Our retrieval-based expansion has two branches: (1) image-based retrieval, where representative domain images are used as visual queries to retrieve semantically and stylistically related samples; and (2) text-based retrieval, where domain-specific captions serve as queries to collect contextually aligned images. Combining image- and text-driven search introduces real-world variations and long-tail visual concepts, yielding a more comprehensive and knowledge-grounded source image pool for downstream editing tasks.

\noindent\textbf{Synthesis-based Expansion.}
To further increase intra-domain diversity, we adopt a synthesis-based expansion strategy that leverages generative models to produce realistic yet semantically coherent variants. Concretely, we first obtain detailed captions for each source image using  MetaCaptioner~\cite{lei2025metacaptionergeneralistvisualcaptioning}, which provides rich, fine-grained descriptions of scene context, object attributes, and style cues. 
We then utilize Qwen-Image to generate multiple variants by introducing element-aware modifications along specific dimensions while preserving the core semantics of the original image. This synthesis-based branch complements retrieval-based expansion by densifying the source image manifold within each domain.

Finally, we remove near-duplicate samples based on perceptual hashes, yielding a highly diverse expanded source pool of over 10M unique images.

\subsection{Adaptive Multi-Agent Editing Synthesis}
\label{sec:Workflow}
After obtaining the expanded source image pool (Section~\ref{sec:source_image_expansion}), we generate editing instructions and corresponding edited images through an adaptive multi-agent framework. We categorize all editing tasks into 23 predefined types and employ a task router powered by Qwen2.5-VL-72B~\cite{bai2025qwen2} to determine which tasks are suitable for each image. Unlike prior single-task approaches~\cite{yu2018generative,zhang2016colorful}, our router uses a rejection-based strategy that explicitly excludes unsuitable tasks while treating the remaining ones as applicable. This selective mechanism enables each image to support multiple content-appropriate editing tasks, thereby enhancing task coverage, data diversity, and dataset scalability.

Based on routing results, each image is dispatched to specialized agents that generate editing instructions and outputs tailored to specific task requirements.
To accommodate heterogeneous editing demands, which often require capturing distinct levels of visual semantics, we instantiate a modular pool of 24 dedicated instruction agents, including the rewriter agent for reasoning workflows. All these instruction agents are driven by Qwen2.5-VL-72B and are equipped with task-specific guidelines.
For edit agents, we collect state-of-the-art open-source models, including Qwen-Image-Edit~\cite{wu2025qwen}, FLUX.1 Kontext~\cite{labs2025flux}, Step1X-Edit~\cite{liu2025step1x}, and Flux-Text~\cite{lan2025fluxtext}. These agents produce instruction-image-edit triplets that yield higher-quality edits and greater dataset diversity.

\noindent\textbf{Text-aware Editing Workflows}. Text-aware editing remains underserved in existing datasets due to scarce resources and low-resolution limitations. We address this through a three-stage pipeline. First, we employ PaddleOCR~\cite{cui2025paddleocr30technicalreport} to detect text regions and extract content, confidence scores, and bounding polygons. Second, a text instruction agent filters candidate regions, validates textual relevance and visual consistency, and generates semantically meaningful editing instructions. Finally, a text edit agent produces masked images and glyph-rendered overlays as inputs for specialist models (\eg, Flux-Text~\cite{lan2025flux}) to perform precise, context-aware text editing. This workflow enables high-quality, semantically aligned text-image editing pairs suitable for text-aware applications.

\noindent\textbf{Knowledge-infused Reasoning Editing Workflows.}
For tasks demanding logical reasoning or world knowledge, we propose reasoning workflows based on an instruction decoupling strategy. Specifically, an instruction agent first constructs complex, reasoning-rich user queries (\eg, embedding explicit reasoning chains or knowledge cues), and a rewriter agent then distills these into concise, executable commands. During data synthesis, these rewritten commands are used to generate images, while the original complex queries are retained as the final user inputs. This decoupling elegantly bridges the gap between complex human intents and the execution limits of current editing models.

\subsection{Task-Aware Quality Verification}
\label{sec:filter}
To control the quality of our large-scale constructed editing corpus in a way that is sensitive to both task type and semantic alignment, we introduce a task-aware verification module built on Qwen2.5-VL-72B. 
For each of the 23 predefined editing tasks, we define a three-dimensional evaluation protocol that assesses: (1) \textit{Instruction Following}, which measures whether the edited image faithfully executes the editing prompt; (2) \textit{Editing Consistency}, which evaluates semantic and structural coherence between the edited output and the original image; and (3) \textit{Generation Quality}, which focuses on visual fidelity, realism, and the suppression of artifacts. Each task is paired with a task-specific evaluation prompt that precisely captures its editing intent, enabling fine-grained and task-aware assessment across diverse manipulation categories.

During filtering, each image-editing pair is routed to its corresponding task branch and automatically evaluated on a 1-to-3 scale along the three dimensions. Specifically, we only retain samples that achieve a perfect score of 3 for Instruction Following, and a score of at least 2 for both Editing Consistency and Generation Quality. 
This adaptive score-based enhancement process effectively removes low-quality or misaligned examples and significantly improves data reliability and diversity. As a result, the refined dataset exhibits stronger instruction alignment, visual coherence, and generalization performance, providing a more robust foundation for training and evaluating UMMs. 

\section{\dataset}
We introduce \textbf{\dataset}, a \textit{large-scale} and \textit{high-quality} image editing dataset with diverse editing tasks, constructed through our \method pipeline described in Sec.~\ref{sec:method}.
\subsection{Edit Type Definition}

To enhance the model's ability for image editing, we organize tasks into six categories: (1) \textit{global-level editing}:  modifying overall style, tone, and background while preserving structure; (2) \textit{object-level editing}: adding, removing, replacing objects or extracting parts with precise boundary handling; (3) \textit{object attribute editing}: adjusting properties like color, material, size, and count while maintaining scene coherence; (4) \textit{text-aware editing}: manipulating textual elements in posters, GUIs, and signage with visual-linguistic understanding; (5) \textit{knowledge-infused reasoning editing}: incorporating domain-specific knowledge including perceptual, symbolic, and scientific reasoning for logically consistent modifications; and (6) \textit{compositional editing}: executing multiple compound instructions coherently in a single operation.
This comprehensive taxonomy covers diverse instruction-based scenarios from global transformations to localized, context-sensitive modifications.

\subsection{Data Analysis}

\begin{figure*}[!th]
  \centering
   \includegraphics[width=1.0\linewidth]{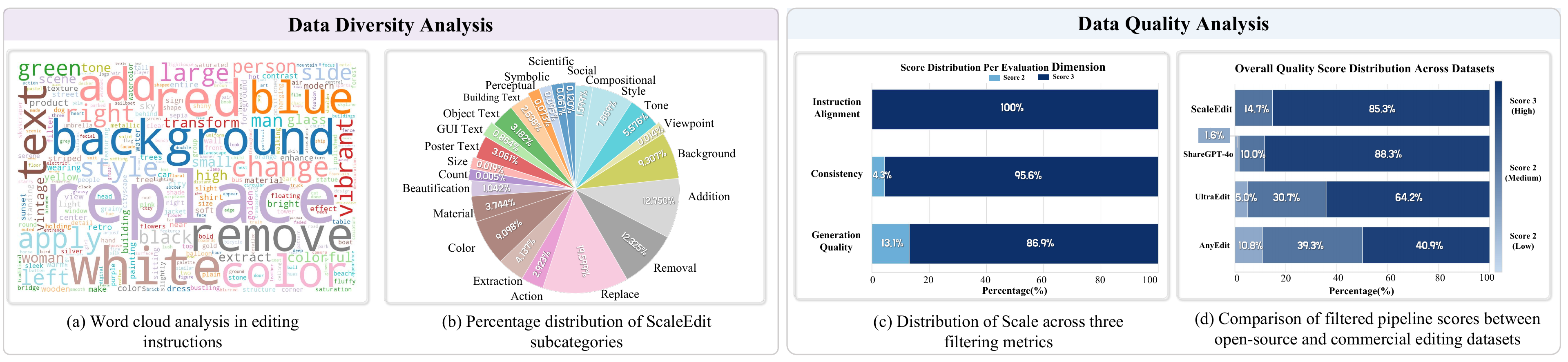}
    \vspace{-1em}
    \caption{ \textbf{Diversity and quality analysis of \dataset.}
    (a) Distribution of editing subcategories demonstrates the broad coverage of editing behaviors in \dataset.
    (b) Word cloud of editing instructions highlights the linguistic richness in edit semantics.
    (c) Multi-dimensional filtering score distribution across three evaluation metrics.
    (d) Comparison of filtering scores across representative datasets (ShareGPT-4o~\cite{chen2025sharegpt}, UltraEdit~\cite{zhao2024ultraedit}, AnyEdit~\cite{yu2024anyedit}), where each sample\textquotesingle s final score is defined as the minimum of its three evaluation metrics.
    }
   \label{fig:data_analysis}
\end{figure*}

\noindent\textbf{Data Diversity Analysis.}
Our ScaleEdit demonstrates exceptional diversity across multiple dimensions of image editing tasks. As shown in Fig.~\ref{fig:data_analysis}(a), the dataset encompasses a comprehensive range of editing categories, with the most prominent being Action (12.3\%), followed by Background, Addition, and Removal operations, ensuring balanced representation across different editing types. The subcategories span from low-level manipulations (Color, Style, Material) to high-level semantic edits (Object Text, Building Text, Compositional changes), covering both local and global image modifications. The word cloud in Fig.~\ref{fig:data_analysis}(b) further illustrates the rich vocabulary used in editing instructions, with frequently appearing terms such as ``background'', ``replace'', ``remove'', and ``white'' indicating that the dataset captures diverse editing intentions. This linguistic diversity, combined with the wide distribution of editing operations, ensures that models trained on \dataset can handle a broad spectrum of real-world editing scenarios.

\noindent\textbf{Data Quality Analysis.}
The meticulously designed filtering pipeline detailed in Sec.~\ref{sec:filter} serves as a robust quality assurance mechanism for our ScaleEdit dataset. As evidenced in Fig.~\ref{fig:data_analysis}(c), the dataset achieves remarkable performance across three critical filtering dimensions: Instruction Alignment, Editing Consistency, and Generation Quality. Notably, 85.3\% of the data instances attain the minimum score of 3 across all metrics, demonstrating exceptional baseline quality from the initial construction phase. Through comprehensive benchmarking, as shown in Fig.~\ref{fig:data_analysis}(d), our filtered results not only substantially surpass current state-of-the-art open-source datasets (UltraEdit~\cite{zhao2024ultraedit} and AnyEdit~\cite{yu2024anyedit}) but also maintain competitive parity with the commercially annotated ShareGPT-4o~\cite{chen2025sharegpt} dataset that leverages GPT-4o for labeling. This high-quality standard ensures that \dataset can serve as a reliable resource for training advanced image editing models.

\section{Experiment}
\label{sec:exp}
\subsection{Experimental Setup}

\noindent\textbf{Settings.}
For the main experiments, we employ UniWorld-V1~\cite{lin2025uniworld} and Bagel~\cite{deng2025bagel} as the baseline unified generative models and finetune them using data only from our dataset.
We trained the model using the entire dataset with a learning rate of 1e-5. Please refer to the appendix for more training details.

\noindent\textbf{Benchmarks.}
We evaluate our models across 4 widely adopted image editing benchmarks, including general editing benchmarks such as GEdit-EN-full~\cite{liu2025step1x} and ImgEdit-Full~\cite{ye2025imgedit}, and knowledge-infused editing benchmarks KRIS-Bench~\cite{wu2025kris}, and RISEBench~\cite{zhao2025envisioning}.

\noindent\textbf{Baselines.}
To further demonstrate the effectiveness of \dataset, we also finetune UniWorld-V1 and Bagel on other existing datasets, including both commercial datasets and open-source datasets, including OmniEdit \cite{wei2024omniedit}, NHR-Edit \cite{kuprashevich2025nohumansrequiredautonomoushighqualityimage} , ImgEdit \cite{ye2025imgedit}, AnyEdit \cite{yu2024anyedit}, and UltraEdit \cite{zhao2024ultraedit}.
These datasets are all finetuned for a single epoch under the same training configurations for fair comparison. 

\subsection{Quantitative Evaluations}

\noindent\textbf{General Editing Performance.}
As shown in Tab.~\ref{tab:main_gedit} and Tab.~\ref{tab:main_imgedit}, the models fine-tuned on our dataset consistently outperform those trained on existing open-source editing datasets across most evaluation dimensions. 
Specifically for the UniWorld-V1 baseline, ScaleEdit reaches an average score of 6.55 on GEdit-EN-Full, substantially surpassing all open-source datasets and exceeding commercial-model-generated datasets, while also improving upon the baseline by an average margin of 0.34 on ImgEdit-Bench, showing enhanced robustness in challenging categories such as \emph{Hybrid} and \emph{Action}. 
Similarly, fine-tuning the Bagel baseline on ScaleEdit yields an impressive average score of 7.17 on GEdit-EN-Full and 3.45 on ImgEdit-Bench, which again consistently outperforms all open-source dataset counterparts. 
These results indicate that the design of our dataset effectively enhances the editing capability of the models and enables stronger generalization across diverse editing types and base architectures.

\begin{table*}[!thbp]
\centering
\small
\caption{
    \textbf{Comparison of fine-tuning results on GEdit-EN-Full~\cite{liu2025step1x}.}
    For each baseline, the best score is shown in \textbf{bold}.
    }
\label{tab:main_gedit}
\scriptsize

\resizebox{0.9\linewidth}{!}{
\begin{tabular}{llcccccccccccc}
\toprule
& \textbf{Model} &
\textbf{Background} & \textbf{Color} & \textbf{Material} & \textbf{Motion} &
\textbf{Portrait} & \textbf{Style} & \textbf{Add} & \textbf{Remove} &
\textbf{Replace} & \textbf{Text} & \textbf{Tone} & \textbf{Avg} \\
\midrule

& GPT-4o \cite{openai2024gpt4o} & 6.96 & 6.85 & 7.10 & 5.41 & 6.74 & 7.44 & 7.51 & 8.73 & 8.55 & 8.45 & 8.69 & 7.49 \\
& OmniGen  \cite{xiao2024omnigen}  & 5.23 & 5.93 & 5.44 & 3.12 & 3.17 & 4.88 & 6.33 & 6.35 & 5.34 & 4.31 & 4.96 & 5.01 \\
& Step1X-Edit \cite{liu2025step1x}   & 7.03 & 6.26 & 6.46 & 3.66 & 5.23 & 7.24 & 7.17 & 6.42 & 7.39 & 7.40 & 6.62 & 6.44 \\
\midrule

\multirow{16}{*}{\rotatebox{90}{\small \emph{Finetuning on UniWorld-V1}}}

& \multicolumn{13}{>{\columncolor{gray!15}}l}{$\blacktriangledown$ \textit{Baseline}} \\
& UniWorld-V1 \cite{lin2025uniworld}  & 4.92 & 6.37 & 4.79 & 1.85 & 4.03 & 5.64 & 7.23 & 6.17 & 5.70 & 1.15 & 5.54 & 4.85 \\
\cmidrule{2-14}

& \multicolumn{13}{>{\columncolor{gray!15}}l}{$\blacktriangledown$ \textit{w/ Commercial Datasets}} \\
& OpenGPT-4o-Image \cite{chen2025opengpt} & 5.94 & 7.99 & 5.76 & 6.13 & 6.51 & 6.19 & 7.64 & 4.84 & 5.84 & 1.28 & 7.27 & 5.95 \\
& ShareGPT-4o-Image \cite{chen2025sharegpt} & 4.93 & 7.94 & 5.54 & 5.84 & \textbf{6.61} & 6.23 & 7.27 & 5.24 & 5.63 & 1.33 & 7.27 & 5.80 \\
& Nano-consistent \cite{ye2025echo}& 5.29 & 7.76 & 4.31 & 4.90 & 6.15 & 3.97 & 6.98 & 3.91 & 5.40 & 1.21 & 6.81 & 5.10 \\
& Pico-Banana   \cite{qian2025pico}    & 6.33 & 7.93 & 5.74 & 6.69 & 6.48 & 5.61 & \textbf{7.66} & 6.00 & 6.16 & 1.84 & 6.98 & 6.13 \\
& GPT-Image-Edit  \cite{wang2025gpt} & 7.24 & 6.94 & \textbf{6.41} & 6.60 & 5.85 & \textbf{7.40} & 7.16 & 6.45 & 6.59 & \textbf{2.49} & 6.18 & 6.30 \\
\cmidrule{2-14}

& \multicolumn{13}{>{\columncolor{gray!15}}l}{$\blacktriangledown$ \textit{w/ Open-source Datasets}} \\
& OmniEdit \cite{wei2024omniedit}  & 5.47 & 7.42 & 5.21 & 5.74 & 6.19 & 6.58 & 6.78 & 4.51 & 4.99 & 1.61 & 6.54 & 5.55 \\
& NHR-Edit \cite{kuprashevich2025nohumansrequiredautonomoushighqualityimage}    & 7.12 & 6.88 & 5.97 & 4.67 & 5.80 & 6.34 & 6.52 & 7.22 & \textbf{6.99} & 1.97 & 5.92 & 5.95 \\
& ImgEdit  \cite{ye2025imgedit} & 5.95 & 7.08 & 4.86 & 5.16 & 5.85 & 6.36 & 6.50 & 4.34 & 5.46 & 1.78 & 5.71 & 5.37 \\
& AnyEdit \cite{yu2025anyedit}   & 4.23 & 4.55 & 4.69 & 4.64 & 4.30 & 5.27 & 4.74 & 3.50 & 5.30 & 1.80 & 3.57 & 4.23 \\
& UltraEdit  \cite{zhao2024ultraedit} & 3.32 & 3.14 & 3.41 & 3.46 & 1.93 & 4.70 & 2.26 & 0.99 & 4.05 & 0.88 & 1.80 & 2.72 \\
\cmidrule{2-14}

& \multicolumn{13}{>{\columncolor{gray!15}}l}{$\blacktriangledown$ \textit{w/ Our Dataset}} \\
& \textbf{ScaleEdit} & \textbf{7.42} & \textbf{8.18} & 5.76 & \textbf{7.07} &
6.51 & 7.09 & 7.39 & \textbf{7.24} &
5.96 & 1.77 & \textbf{7.64} & \textbf{6.55} \\

\midrule

\multirow{16}{*}{\rotatebox{90}{\small \emph{Finetuning on Bagel}}}

& \multicolumn{13}{>{\columncolor{gray!15}}l}{$\blacktriangledown$ \textit{Baseline}} \\
& Bagel \cite{lin2025uniworld}  & 6.73 & 6.84  & 6.33  & 6.86  & 5.49  & 5.91  & 7.81  & 6.60  & 7.35  & 6.34  & 5.56  & 6.52 \\
\cmidrule{2-14}

& \multicolumn{13}{>{\columncolor{gray!15}}l}{$\blacktriangledown$ \textit{w/ Commercial Datasets}} \\
& OpenGPT-4o-Image \cite{chen2025opengpt} & 7.15 & 6.92 & 6.15 & 6.28 & 5.10 & 5.65 & 7.88 & 6.81 & 7.09 & 6.79 & 5.81 & 6.54 \\
& ShareGPT-4o-Image \cite{chen2025sharegpt} & 7.17  & 6.91  & 6.30  & 6.56  & 5.22  & 5.74  & 7.79  & 6.72  & 7.17  & 6.91  & 5.99  & 6.59 \\
& Nano-consistent \cite{ye2025echo} & 7.13  & 6.93  & 6.27  & 6.47  & 5.08  & 5.75  & 7.89  & 6.76  & 7.15  & 6.86  & 6.03  & 6.57 \\
& Pico-Banana   \cite{qian2025pico} & 7.22  & 6.78  & 6.40  & 6.29  & 5.25  & 5.95  & 8.17  & 6.89  & 6.99  & 6.93  & 5.98  & 6.62 \\
& GPT-Image-Edit  \cite{wang2025gpt} & \textbf{7.81}  & 7.15  & \textbf{6.89}  & 7.30  & \textbf{6.60}  & 6.61  & 8.12  & 7.30  & 7.48  & 6.90  & 5.78  & 7.09 \\
\cmidrule{2-14}

& \multicolumn{13}{>{\columncolor{gray!15}}l}{$\blacktriangledown$ \textit{w/ Open-source Datasets}} \\
& OmniEdit \cite{wei2024omniedit} & 7.17  & 6.94  & 6.35  & 6.40  & 5.16  & 5.60  & 8.03  & 6.85  & 7.01  & 6.87  & 5.95  & 6.57 \\
& NHR-Edit \cite{kuprashevich2025nohumansrequiredautonomoushighqualityimage} & 7.20  & 6.96  & 6.62  & 6.98  & 5.95  & 5.89  & 8.07  & 7.03  & 7.51  & 6.60  & 6.03  & 6.80 \\
& ImgEdit \cite{ye2025imgedit} & 7.30  & 7.31  & 6.33  & 6.15  & 5.39  & 6.08  & 8.07  & 7.04  & 7.11  & 6.71  & 6.40  & 6.72 \\
& AnyEdit \cite{yu2025anyedit} & 6.70  & 7.10  & 6.26  & 5.21  & 4.88  & 5.61  & 8.13  & 7.25  & 6.76  & 6.72  & 6.31  & 6.45 \\
& UltraEdit  \cite{zhao2024ultraedit} & 7.44  & 6.31  & 6.68  & 5.93  & 5.83  & \textbf{6.80}  & 7.15  & \textbf{7.93}  & 7.51  & 6.32  & 6.05  & 6.72 \\
\cmidrule{2-14}

& \multicolumn{13}{>{\columncolor{gray!15}}l}{$\blacktriangledown$ \textit{w/ Our Dataset}} \\
& \textbf{ScaleEdit}  & 7.43  & \textbf{7.59}  & 6.43  & \textbf{7.73}  & 6.49  & 6.35  & \textbf{8.27}  & 7.66  & \textbf{7.54}  & \textbf{6.93}  & \textbf{6.45}  & \textbf{7.17} \\

\bottomrule
\end{tabular}}
\end{table*}

\begin{table*}[!thbp]
\centering
\small
\caption{
    \textbf{Comparison of fine-tuning results on ImgEdit-Bench~\cite{ye2025imgedit}.}
    For each baseline, the best score is shown in \textbf{bold}.
}
\scriptsize
\label{tab:main_imgedit}
\resizebox{0.85\linewidth}{!}{
\begin{tabular}{llcccccccccc}
\toprule
& \textbf{Model} &
\textbf{Add} & \textbf{Adjust} & \textbf{Extract} & \textbf{Replace} &
\textbf{Remove} & \textbf{Background} & \textbf{Style} &
\textbf{Hybrid} & \textbf{Action} & \textbf{Avg} \\
\midrule

\scriptsize
& GPT-4o \cite{openai2024gpt4o} & 4.61 & 4.33 & 2.90 & 4.35 & 3.66 & 4.57 & 4.93 & 3.96 & 4.89 & 4.20 \\
& OmniGen \cite{xiao2024omnigen}          & 3.47 & 3.04 & 1.71 & 2.94 & 2.43 & 3.21 & 4.19 & 2.24 & 3.38 & 2.96 \\
& Step1X-Edit \cite{liu2025step1x}       & 3.88 & 3.14 & 1.76 & 3.40 & 2.41 & 3.16 & 4.63 & 2.64 & 2.52 & 3.06 \\
\midrule

\multirow{16}{*}{\rotatebox{90}{\small \emph{Finetuning on UniWorld-V1}}}

& \multicolumn{11}{>{\columncolor{gray!15}}l}{$\blacktriangledown$ \textit{Baseline}} \\
& UniWorld-V1 \cite{lin2025uniworld} & 3.82 & 3.64 & \textbf{2.27} & 3.47 & 3.24 & 2.99 & 4.21 & 2.96 & 2.74 & 3.26 \\
\cmidrule{2-12}

& \multicolumn{11}{>{\columncolor{gray!15}}l}{$\blacktriangledown$ \textit{w/ Commercial Datasets}} \\
& OpenGPT-4o-Image \cite{chen2025opengpt}   & \textbf{4.18} & 3.96 & 1.99 & 3.44 & 2.62 & 3.67 & 4.65 & 2.74 & 3.07 & 3.37 \\
& ShareGPT-4o-Image \cite{chen2025sharegpt}  & 4.03 & \textbf{4.01} & 1.83 & 3.46 & 2.84 & 3.58 & 4.79 & 2.75 & 3.09 & 3.38 \\
& Nano-consistent \cite{ye2025echo}    & 3.96 & 3.62 & 1.90 & 3.41 & 2.45 & 3.15 & 4.14 & 2.80 & \textbf{3.97} & 3.27 \\
& Pico-Banana \cite{qian2025pico}        & 4.07 & 3.99 & 1.83 & 3.59 & 3.44 & 3.49 & 4.23 & 3.01 & 3.42 & 3.45 \\
& GPT-Image-Edit \cite{wang2025gpt}     & 3.97 & 3.16 & 1.92 & 3.55 & 3.52 & 3.36 & \textbf{4.80} & 3.00 & 3.44 & 3.41 \\
\cmidrule{2-12}

& \multicolumn{11}{>{\columncolor{gray!15}}l}{$\blacktriangledown$ \textit{w/ Open-source Datasets}} \\
& OmniEdit \cite{wei2024omniedit}  & 3.78 & 3.39 & 2.11 & 3.02 & 2.53 & 3.12 & 4.46 & 2.71 & 2.86 & 3.11 \\
& NHR-Edit \cite{kuprashevich2025nohumansrequiredautonomoushighqualityimage}  & 3.85 & 3.04 & 1.75 & 3.74 & \textbf{3.87} & 2.90 & 4.44 & \textbf{3.36} & 2.53 & 3.28 \\
& ImgEdit \cite{ye2025imgedit}   & 3.70 & 3.34 & 2.11 & 3.29 & 2.53 & 3.06 & 4.59 & 2.55 & 2.84 & 3.11 \\
& AnyEdit \cite{yu2025anyedit}   & 3.71 & 2.48 & 1.82 & 3.20 & 2.86 & 2.04 & 3.82 & 2.62 & 2.94 & 2.83 \\
& UltraEdit \cite{zhao2024ultraedit} & 2.14 & 1.59 & 1.98 & 2.62 & 1.23 & 1.80 & 4.01 & 1.07 & 1.95 & 2.04 \\
\cmidrule{2-12}

& \multicolumn{11}{>{\columncolor{gray!15}}l}{$\blacktriangledown$ \textit{w/ Our Dataset}} \\
& \textbf{ScaleEdit}      & 4.05 & 3.88 & 2.15 & \textbf{3.77} & 2.95 & \textbf{3.93} & 4.71 & 3.33 & 3.67 & \textbf{3.60} \\

\midrule

\multirow{16}{*}{\rotatebox{90}{\small \emph{Finetuning on Bagel}}}

& \multicolumn{11}{>{\columncolor{gray!15}}l}{$\blacktriangledown$ \textit{Baseline}} \\
& Bagel \cite{lin2025uniworld} & 3.56 & 3.31 & 1.7 & 3.30 & 2.62 & 3.24 & 4.49 & 2.38 & 4.17 & 3.20 \\
\cmidrule{2-12}

& \multicolumn{11}{>{\columncolor{gray!15}}l}{$\blacktriangledown$ \textit{w/ Commercial Datasets}} \\
& OpenGPT-4o-Image \cite{chen2025opengpt}   & 3.48 & 3.23 & 1.63 & 3.27 & 2.60 & 3.24 & 4.29 & 2.48 & 3.77 & 3.11 \\
& ShareGPT-4o-Image \cite{chen2025sharegpt} & 3.47 & 3.32  & 1.67  & 3.34  & 2.68  & 3.23  & 4.31  & 2.55  & 3.70  & 3.14 \\
& Nano-consistent \cite{ye2025echo} & 3.51 & 3.25 & 1.77  & 3.32  & 2.64  & 3.20  & 4.32  & 2.74  & 3.70  & 3.16 \\
& Pico-Banana \cite{qian2025pico} & 3.52 & 3.15 & 1.67  & 3.35  & 2.70  & 3.19  & 4.37  & 2.40  & 3.66  & 3.11 \\
& GPT-Image-Edit \cite{wang2025gpt} & 3.80 & 3.25 & 1.94  & 3.82  & 3.02  & 3.67  & \textbf{4.63}  & 2.48  & 4.03  & 3.40 \\
\cmidrule{2-12}

& \multicolumn{11}{>{\columncolor{gray!15}}l}{$\blacktriangledown$ \textit{w/ Open-source Datasets}} \\
& OmniEdit \cite{wei2024omniedit} & 3.48 & 3.15 & 1.65  & 3.29  & 2.68  & 3.24  & 4.34  & 2.55  & 3.86  & 3.14 \\
& NHR-Edit \cite{kuprashevich2025nohumansrequiredautonomoushighqualityimage} & \textbf{4.19} & \textbf{3.48} & 1.65 & 3.51 & 3.12 & 3.31 & 4.28 & 2.99 & 3.81 & 3.33 \\
& ImgEdit \cite{ye2025imgedit} & 3.52 & 3.29 & 1.61  & 3.47  & 2.77  & 3.40  & 4.40  & 2.52  & 3.68  & 3.18 \\
& AnyEdit \cite{yu2025anyedit} & 3.42 & 3.13 & 1.69  & 3.24  & 2.86  & 3.18  & 4.38  & 2.51  & 3.69  & 3.12 \\
& UltraEdit \cite{zhao2024ultraedit} & 3.67  & 3.26 & 1.82  & 3.17  & 3.13  & 3.32  & 4.59 & \textbf{3.22}  & 2.66  & 3.20 \\
\cmidrule{2-12}

& \multicolumn{11}{>{\columncolor{gray!15}}l}{$\blacktriangledown$ \textit{w/ Our Dataset}} \\
& \textbf{ScaleEdit}      & 3.68 & 2.97 & \textbf{2.12}  & \textbf{3.83}  & \textbf{3.15}  & \textbf{3.72}  & 4.43  & 2.91  & \textbf{4.20}  & \textbf{3.45} \\
\bottomrule
\end{tabular}
}
\end{table*}

\begin{table*}[htbp]
\centering
\small
\caption{
    \textbf{Comparison of fine-tuning results on RISEBench~\cite{zhao2025envisioning} and KRIS Bench ~\cite{wu2025kris}.}
    For each baseline, the best score is shown in \textbf{bold}.
    }
\label{tab:main_reason}
\scriptsize
\resizebox{0.85\linewidth}{!}{
\begin{tabular}{llcccccccc}
\toprule
& \multirow[c]{2}{*}{\textbf{Model}}
& \multicolumn{4}{c}{\textbf{RISEBench \cite{zhao2025envisioning}}} 
& \multicolumn{4}{c}{\textbf{KRIS Bench \cite{wu2025kris}}} \\
\cmidrule(lr){3-6} \cmidrule(lr){7-10}
&  & \textbf{Reasoning} & \textbf{ApprConsistency} & \textbf{VisualPlausibility} & \textbf{Overall} 
& \textbf{Factual} & \textbf{Conceptual} & \textbf{Procedural} & \textbf{Overall} \\
\midrule

& GPT-4o \cite{openai2024gpt4o}             & 62.80 & 80.20 & 94.90 & 28.90 & 79.80 & 81.37 & 78.32 & 80.09 \\

& OmniGen \cite{xiao2024omnigen}          & 22.00 & 32.60 & 55.30 & 0.80  & 33.11 & 28.02 & 23.89 & 28.85  \\
& Step1X-Edit \cite{liu2025step1x}       & 25.10 & 41.50 & 73.50 & 1.90  & 45.52 & 48.01 & 31.82 & 43.29 \\

\midrule

\multirow[c]{16}{*}{\rotatebox{90}{\small \emph{Finetuning on UniWorld-V1}}}

& \multicolumn{9}{>{\columncolor{gray!15}}l}{$\blacktriangledown$ \textit{Baseline}} \\
& UniWorld-V1 \cite{lin2025uniworld}      & 18.33 & \textbf{65.79} & 86.63 & 2.22 & 47.71 & 44.80 & \textbf{47.92} & 50.27 \\
\cmidrule{2-10}

& \multicolumn{9}{>{\columncolor{gray!15}}l}{$\blacktriangledown$ \textit{w/ Commercial Datasets}} \\
& OpenGPT-4o-Image \cite{chen2025opengpt}  & 25.07 & 52.62 & 89.72 & 2.50 & 53.49 & 60.03 & 39.51 & 53.22 \\
& ShareGPT-4o-Image \cite{chen2025sharegpt} & 26.32 & 60.84 & 89.08 & 5.00 & 56.07 & \textbf{64.29} & 36.15 & 55.24 \\
& Nano-consistent \cite{ye2025echo}   & 22.71 & 56.16 & 90.36 & 3.06 & 54.92 & 57.88 & 36.64 & 52.03 \\
& Pico-Banana \cite{qian2025pico}       & 22.15 & 65.44 & 91.09 & 3.61 & \textbf{58.86} & 63.87 & 38.61 & 56.51 \\
& GPT-Image-Edit \cite{wang2025gpt}    & \textbf{31.01} & 40.65 & 88.27 & 2.22 & 51.62 & 60.71 & 33.73 & 51.60 \\
\cmidrule{2-10}

& \multicolumn{9}{>{\columncolor{gray!15}}l}{$\blacktriangledown$ \textit{w/ Open-source Datasets}} \\
& OmniEdit \cite{wei2024omniedit}         & 23.26 & 49.29 & 88.09 & 4.17 & 51.71 & 57.85 & 27.50 & 48.87 \\
& NHR-Edit \cite{kuprashevich2025nohumansrequiredautonomoushighqualityimage}          & 25.25 & 43.06 & 86.64 & 3.33 & 50.65 & 60.57 & 28.65 & 50.07 \\
& ImgEdit \cite{ye2025imgedit}           & 23.75 & 48.30 & 84.91 & 3.05 & 51.15 & 58.68 & 28.92 & 49.44 \\
& AnyEdit \cite{yu2025anyedit}           & 24.65 & 39.80 & 79.18 & 2.50 & 47.81 & 54.35 & 23.24 & 45.09 \\
& UltraEdit \cite{zhao2024ultraedit}         & 24.38 & 22.24 & 81.81 & 0.83 & 36.61 & 43.59 & 20.72 & 36.17 \\
\cmidrule{2-10}

& \multicolumn{9}{>{\columncolor{gray!15}}l}{$\blacktriangledown$ \textit{w/ Our Dataset}} \\
& \textbf{ScaleEdit} & 26.18 & 57.64 & \textbf{91.18} & \textbf{5.55} &
57.76 & \textbf{64.29} & 39.95 & \textbf{56.60} \\

\midrule

\multirow[c]{16}{*}{\rotatebox{90}{\small \emph{Finetuning on Bagel}}}
& \multicolumn{9}{>{\columncolor{gray!15}}l}{$\blacktriangledown$ \textit{Baseline}} \\
& Bagel \cite{lin2025uniworld}      & 36.50 & 53.50 & 73.00 & 6.10 & 47.71 & 44.80 & 47.92 & 50.27 \\
\cmidrule{2-10}

& \multicolumn{9}{>{\columncolor{gray!15}}l}{$\blacktriangledown$ \textit{w/ Commercial Datasets}} \\
& OpenGPT-4o-Image \cite{chen2025opengpt}  & 36.70 & 55.54 & 71.08 & 5.80 & 67.6 & 57.59 & 59.58 & 61.09 \\ 
& ShareGPT-4o-Image \cite{chen2025sharegpt} & 36.67 & 56.59 & 72.09 & 7.20 & 66.92 & 58.31 & 59.85 & 61.28 \\
& Nano-consistent \cite{ye2025echo}   & 35.56 & 58.07 & 71.73 & 6.70 & 66.21 & 58.17 & 58.61 & 60.71 \\
& Pico-Banana \cite{qian2025pico}       & 35.97 & 57.15 & 71.91 & 6.10 & 67.38 & 57.96 & 58.91 & 61.04 \\
& GPT-Image-Edit \cite{wang2025gpt}    & \textbf{37.64} & 57.86 & 78.73 & 7.20 & 66.86 & 44.32 & \textbf{60.70} & 62.92 \\
\cmidrule{2-10}

& \multicolumn{9}{>{\columncolor{gray!15}}l}{$\blacktriangledown$ \textit{w/ Open-source Datasets}} \\
& OmniEdit \cite{wei2024omniedit}         & 36.94 & 57.01 & 70.73 & 7.20 & 66.04 & 57.59 & 59.46 & 60.59 \\
& NHR-Edit \cite{kuprashevich2025nohumansrequiredautonomoushighqualityimage}  & 35.62 & 56.66 & 72.45 & 6.40 & 65.74 & 57.93 & 58.04 & 60.31 \\
& ImgEdit \cite{ye2025imgedit}           & 35.42 & 59.07 & 74.27 & 6.40 & 69.61 & 57.28 & 58.82 & 62.01 \\
& AnyEdit \cite{yu2025anyedit}           & 35.83 & \textbf{60.20} & 69.91 & 6.90 & 66.96 & 38.89 & 56.71 & 61.20 \\
& UltraEdit \cite{zhao2024ultraedit}         & 30.21 & 59.63 & \textbf{80.00} & 3.90 & 66.26 & 56.55 & 52.91 & 58.64 \\
\cmidrule{2-10}

& \multicolumn{9}{>{\columncolor{gray!15}}l}{$\blacktriangledown$ \textit{w/ Our Dataset}} \\
& \textbf{ScaleEdit} & 36.18 & 59.07 & 72.36 & \textbf{7.50} & \textbf{70.24} & \textbf{60.78} & 60.54 & \textbf{63.58} \\
\bottomrule
\end{tabular}}
\end{table*}

\noindent\textbf{Knowledge-infused Editing Performance.}
As shown in Tab.~\ref{tab:main_reason}, overall performance on these reasoning benchmarks remains limited for all models, reflecting the inherent difficulty of reasoning-informed editing tasks. While Bagel demonstrates top-level reasoning capabilities among open-source systems, it still lags significantly behind closed-source commercial models like GPT-4o~\cite{openai2024gpt4o}. 
Despite these challenges, fine-tuning UniWorld-V1 on ScaleEdit achieves an overall score of 5.55 on RISEBench and 56.60 on KRIS-Bench, matching leading open-source models and exhibiting reliable performance in dimensions such as \emph{Factual}, and \emph{Conceptual}. 
Similarly, applying ScaleEdit to the stronger Bagel baseline further elevates performance, reaching an overall score of 7.50 on RISEBench and 63.58 on KRIS Bench. 
These results further validate that our data construction pipeline, \method, consistently enhances reasoning-based editing capabilities across different base architectures under a fully open-source and reproducible setup.

\subsection{Ablation Study}

\noindent\textbf{Equal-scale Comparison.}
To isolate the effect of data quality from dataset scale, we conduct an equal-scale ablation by resampling each existing editing dataset to a consistent scale of 1M instruction-image-edit triplets, and subsequently fine-tuning UniWorld-V1 \cite{lin2025uniworld} and Janus-Pro \cite{chen2025janus} on them respectively.
As shown in Tab.~\ref{tab:ablationi_gedit_imgedit}, \method demonstrates leading performance on most of the evaluation metrics compared with current open-source and commercial datasets.
These results indicate that our dataset is not only larger in scale but also superior in data quality compared to previous datasets.

\begin{table}[!thbp]
\centering
\scriptsize
\caption{
    \textbf{Ablation study of different datasets with equal scale. }
    All experiments are conducted under the exact same settings and data size for a fair comparison.
    The best score is shown in \textbf{bold}, and the second best is \underline{underlined}.
    }
\label{tab:ablationi_gedit_imgedit}
\resizebox{0.75\linewidth}{!}{
\begin{tabular}{lcccc}
\toprule
\multirow{2}{*}{\textbf{ Training Data}}& \multicolumn{2}{c}{\textbf{UniWorld-V1 \cite{lin2025uniworld}}} & \multicolumn{2}{c}{\textbf{Janus-Pro \cite{chen2025janus}}} \\
\cmidrule(lr){2-3} \cmidrule(lr){4-5}
& \textbf{ImgEdit \cite{ye2025imgedit}} & \textbf{GEdit \cite{liu2025step1x}} & \textbf{ImgEdit \cite{ye2025imgedit} } & \textbf{GEdit \cite{liu2025step1x}} \\
\midrule
\multicolumn{5}{>{\columncolor{gray!15}}l}{$\blacktriangledown$ \textit{Baseline (No Fine-tuning)}} \\
Original & 3.26 & 4.85 & -- & -- \\
\midrule

\multicolumn{5}{>{\columncolor{gray!15}}l}{$\blacktriangledown$ \textit{w/ Commercial Datasets}} \\
OpenGPT-4o-Image \cite{chen2025opengpt}   & 3.37 & 6.06 & \underline{3.08} & 4.52 \\
ShareGPT-4o-Image \cite{chen2025sharegpt} & 3.49 & 6.07 & 2.96 & 4.47 \\
Nano-consistent \cite{ye2025echo}       & 3.21 & 5.02 & 2.56 & 3.14 \\
Pico-Banana \cite{qian2025pico}           & 3.42 & 5.99 & 2.01 & 2.03 \\
GPT-Image-Edit \cite{wang2025gpt}         & \textbf{3.51} & \underline{6.12} & 3.03 & \underline{4.87} \\
\midrule

\multicolumn{5}{>{\columncolor{gray!15}}l}{$\blacktriangledown$ \textit{w/ Open-source Datasets}} \\
OmniEdit \cite{wei2024omniedit}  & 3.08 & 5.25 & 1.92 & 2.22 \\
ImgEdit \cite{ye2025imgedit}     & 2.99 & 4.90 & 1.99 & 1.98 \\
NHR-Edit \cite{kuprashevich2025nohumansrequiredautonomoushighqualityimage} & 3.19 & 5.66 & 2.00 & 2.02 \\
AnyEdit \cite{yu2025anyedit}     & 2.97 & 5.19 & 2.13 & 2.47 \\
UltraEdit \cite{zhao2024ultraedit} & 2.79 & 3.92 & 2.27 & 3.12 \\
\midrule

\multicolumn{5}{>{\columncolor{gray!15}}l}{$\blacktriangledown$ \textit{w/ Our Dataset}} \\
\textbf{ScaleEdit} & \underline{3.50} & \textbf{6.15} & \textbf{3.17} & \textbf{4.92} \\
\bottomrule
\end{tabular}
}
\end{table}

\noindent\textbf{Effect of Task Router and Data Filtering.}
As shown in Tab.~\ref{tab:abaltion_filter}, disabling the task router leads to consistent drops across GEdit and ImgEdit for Janus-Pro and UniWorld-V1, highlighting that appropriate task assignment is crucial. In addition, training on the filtered subset also outperforms the unfiltered counterpart: UniWorld-V1 improves from 6.06 to 6.15 on GEdit and from 3.36 to 3.50 on ImgEdit; Janus-Pro improves from 4.82 to 4.92 and from 3.09 to 3.17, respectively. Such results indicate that both our task router and filtering mechanism refine the data distribution towards higher consistency and quality.

\begin{table*}[!thbp]
\centering
\caption{
    \textbf{Ablation study on Data Filtering and Task Router.} 
    Our routing and filtering strategy improves performance on general editing tasks across different models.
    }
\resizebox{0.75\linewidth}{!}{
\begin{tabular}{ccccc}
\toprule
\textbf{Model} & \textbf{Task Router} & \textbf{Data Filtering} & \textbf{ImgEdit \cite{ye2025imgedit} } & \textbf{GEdit \cite{liu2025step1x} } \\
\midrule
\multirow{3}{*}{Janus-Pro \cite{chen2025janus}} & \cmark & \xmark & 3.09 & 4.82 \\
                       & \xmark & \cmark & 3.14 & 4.85 \\
                       & \cmark & \cmark & 3.17 & 4.92 \\
\midrule
\multirow{3}{*}{UniWorld-V1 \cite{lin2025uniworld}} & \cmark & \xmark & 3.36 & 6.06 \\
                        & \xmark & \cmark & 3.43 & 6.07 \\
                          & \cmark & \cmark & 3.50 & 6.15 \\
\bottomrule
\end{tabular}
}
\label{tab:abaltion_filter}

\end{table*}

\noindent\textbf{Impact of Instruction Rewriting.}
To further evaluate the effectiveness of the instruction rewriting agent on knowledge-intensive tasks, we ablate this module using a 1M subset of our dataset.
As shown in Tab.~\ref{tab:abaltion_rewrite}, incorporating instruction rewriting leads to a notable improvement on the RISEBench, with UniWorld-V1 score increasing from 3.05 to 4.17, indicating that the rewritten instructions better support the interpretation and execution of reasoning-based edits. 

\begin{table*}[!thbp]
\centering
\caption{
    \textbf{Ablation study on instruction rewriting.} 
    VisPlaus. denotes the Visual Plausibility sub-dimension.
    }
\renewcommand{\arraystretch}{0.9} 
\scriptsize
\setlength{\tabcolsep}{3pt}
\resizebox{0.75\linewidth}{!}{
\begin{tabular}{clcccc}
\toprule
\multirow{2}{*}{\textbf{Model}} & \multirow{2}{*}{\textbf{Training Data}} & \multicolumn{4}{c}{\textbf{RISEBench \cite{zhao2025envisioning}}} \\
\cmidrule(lr){3-6}
 &  & Reasoning & Consistency & VisPlaus. & Overall \\
\midrule
\multirow{2}{*}{UniWorld-V1 \cite{lin2025uniworld}} 
    & w/o rewrite   &  22.75 & 57.35 & 90.26 & 3.05 \\
    & w/ rewrite  & 24.38 & 64.24 & 90.64 & 4.17 \\
\bottomrule
\end{tabular}
}
\label{tab:abaltion_rewrite}

\end{table*}

\subsection{Reliability of Quality Verification}

To establish a robust and cost-effective open-source evaluation pipeline, we sought to identify a reliable surrogate for proprietary models. To this end, we evaluated two leading open-source MLLMs, Qwen2.5-VL-72B~\cite{bai2025qwen2} and InternVL3-72B~\cite{zhu2025internvl3exploringadvancedtraining}. We iteratively refined our prompts and then assessed their alignment with GPT-4o's judgments on a diverse set of 10k instances. To comprehensively measure this alignment, we report two metrics: Accuracy ($\uparrow$), which calculates the exact agreement rate between the evaluated model and the reference, and Mean Absolute Error (MAE, $\downarrow$), which quantifies the average magnitude of score deviations. As shown in \cref{tab:consistency_open_gpt4o}, Qwen2.5-VL-72B demonstrates superior alignment with GPT-4o, consistently achieving higher accuracy and lower MAE across all three evaluation dimensions compared to InternVL3-72B. 

\begin{table}[htbp]
\centering
\caption{
    \textbf{Alignment of open-source MLLM judges with GPT-4o.}
    We compare the scoring consistency of different models against GPT-4o across three key quality dimensions. Higher Accuracy and lower MAE indicate better alignment with GPT-4o. Qwen2.5-VL-72B consistently outperforms InternVL3-72B across all metrics.
}
\resizebox{0.85\linewidth}{!}{
\begin{tabular}{ccccccc}
\toprule
\multirow{3}{*}{\textbf{Models}} & 
\multicolumn{3}{c}{\textbf{Accuracy}} & 
\multicolumn{3}{c}{\textbf{MAE}} \\
\cmidrule(lr){2-4} \cmidrule(lr){5-7}
 & \makecell{\textbf{Instruction}\\\textbf{Following}} 
 & \makecell{\textbf{Editing}\\\textbf{Consistency}} 
 & \makecell{\textbf{Generation}\\\textbf{Quality}} 
 & \makecell{\textbf{Instruction}\\\textbf{Following}} 
 & \makecell{\textbf{Editing}\\\textbf{Consistency}} 
 & \makecell{\textbf{Generation}\\\textbf{Quality}} \\
\midrule
InternVL3-72B~\cite{zhu2025internvl3exploringadvancedtraining} & 0.81 & 0.61 & 0.85 & 0.22 & 0.34 & 0.22 \\
Qwen2.5-VL-72B~\cite{bai2025qwen2} & 0.82 & 0.63 & 0.89 & 0.17 & 0.28 & 0.17 \\
\bottomrule
\end{tabular}
}
\label{tab:consistency_open_gpt4o}
\end{table}

To further validate its practical reliability against human perception, we conducted a human study involving 20 domain experts on 1,000 samples. The results in \cref{tab:consistency_human} demonstrate that Qwen2.5-VL-72B yields highly competitive accuracy and error margins compared with GPT-4o~\cite{openai2024gpt4o}, justifying its effectiveness as a reliable, scalable, and fully open-source judge for large-scale quality verification.

\begin{table*}[!htbp]
\centering
\caption{
    \textbf{Alignment of MLLM judges with human preference.}
    Higher Accuracy and lower MAE indicate better alignment with human preference. Qwen2.5-VL-72B shows highly competitive alignment with human evaluators, demonstrating its reliability at scale compared to GPT-4o.
}
\resizebox{0.85\linewidth}{!}{
\begin{tabular}{ccccccc}
\toprule
\multirow{3}{*}{\textbf{Models}} & 
\multicolumn{3}{c}{\textbf{Accuracy}} & 
\multicolumn{3}{c}{\textbf{MAE}} \\
\cmidrule(lr){2-4} \cmidrule(lr){5-7}
 & \makecell{\textbf{Instruction}\\\textbf{Following}} 
 & \makecell{\textbf{Editing}\\\textbf{Consistency}} 
 & \makecell{\textbf{Generation}\\\textbf{Quality}} 
 & \makecell{\textbf{Instruction}\\\textbf{Following}} 
 & \makecell{\textbf{Editing}\\\textbf{Consistency}} 
 & \makecell{\textbf{Generation}\\\textbf{Quality}} \\
\midrule
Qwen2.5-VL-72B~\cite{bai2025qwen2}  & 0.78 & 0.67 & 0.78 & 0.24 & 0.31 & 0.26 \\
GPT-4o~\cite{openai2024gpt4o} & 0.86 & 0.75 & 0.82 & 0.16 & 0.22 & 0.20 \\
\bottomrule
\end{tabular}}
\label{tab:consistency_human}
\end{table*}

\subsection{More Results}

\subsubsection{Generalization across Different Models} 
We further validated \dataset on more representative models, and the results in \cref{tab:extend_exp} show consistent performance gains across different architectures, indicating the value of \dataset as a general high-quality editing dataset. 

\begin{table}[htbp]
\centering
\caption{
    \textbf{Finetuned results across different models.} All experiments are conducted using the same settings as the equal-scale comparison.
}
\begin{tabular}{ccccc}
\toprule
\multirow{2}{*}{\textbf{Models}} & \multicolumn{2}{c}{\textbf{GEdit}~\cite{liu2025step1x}} & \multicolumn{2}{c}{\textbf{ImgEdit}~\cite{ye2025imgedit}} \\
\cmidrule(lr){2-3} \cmidrule(lr){4-5}
                        & \textbf{Baseline}    & \textbf{Finetuned}    & \textbf{Baseline}     & \textbf{Finetuned}    \\ \midrule
InstructPix2Pix~\cite{brooks2023instructpix2pix}         & 3.68        & 3.78        & 1.88         & 2.06         \\
OmniGen~\cite{xiao2024omnigen}                 & 5.06        & 5.49        & 2.96         & 3.15         \\
Step1X-Edit~\cite{liu2025step1x}             & 6.70        & 7.14        & 3.06         & 3.21         \\
\bottomrule
\end{tabular}
\label{tab:extend_exp}
\end{table}

\subsubsection{Qualitative Results} 
\begin{figure*}[!thbp]
  \centering
   \includegraphics[width=1.0\linewidth]{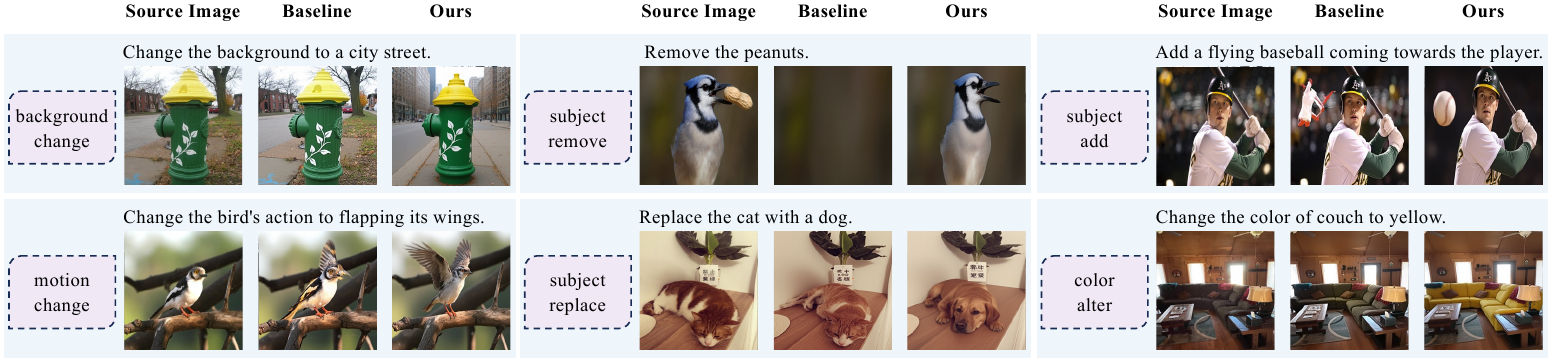}
    \vspace{-1em}
    \caption{ \textbf{Qualitative comparison of UniWorld-V1 before and after fine-tuning on \dataset.}
    }
   \label{fig:example_final}
\end{figure*}

Fig.~\ref{fig:example_final} presents qualitative comparisons between the baseline UniWorld-V1 and the model fine-tuned on \dataset. Across various editing types, the fine-tuned model more faithfully follows the editing instructions and better preserves the original image structure, whereas the baseline often fails to complete the edits or introduces noticeable artifacts. These results indicate that \method and the resulting dataset \dataset substantially improve the performance of the model on visual image-editing tasks.

\section{Conclusion}
This paper introduces \method, a hierarchical framework bridging commercial and open-source datasets through world-knowledge enhanced image expansion, adaptive multi-agent editing workflows, and multi-dimensional quality verification. Using this pipeline, we construct \dataset, a 12M dataset spanning diverse editing tasks and visual domains.
Fine-tuned on \dataset, both UniWorld-V1 and Bagel achieve competitive performance on general and knowledge-infused editing benchmarks, demonstrating that open-source agentic pipelines can match commercial-level quality while remaining cost-efficient and scalable.
We believe this framework and dataset will advance image editing capabilities in UMMs.

\noindent\textbf{Limitations and Future work}.
While \method yields high-quality data, relying on off-the-shelf open-source generators inherently caps the visual quality ceiling. Although targeted fine-tuning mitigates this, systematically training expert models across 23 diverse tasks requires prohibitive costs. Furthermore, iterative multi-turn editing remains underexplored. Future work will explore efficient task-specific fine-tuning to push visual boundaries and extend \dataset to support complex multi-turn conversational editing.

\clearpage
\bibliographystyle{plain}
\bibliography{refs}

%%%%%%%%%%%%%%%%%%%%%%%%%%%%%%%%%%%%%%%%%%%%%%%%%%%%%%%%%%%%

\clearpage
\appendix
\section{Implementation Details of \method}

\subsection{Source Image Expansion}

The expansion mechanism populates our source pool with approximately 1.5M high-quality samples, ensuring a diverse foundation for subsequent editing.

\paragraph{Retrieval-based Expansion.}
Our retrieval pipeline bifurcates into image-based and text-based branches to capture both visual and semantic diversity. For the image-based branch, we first employ representative domain images as visual queries and collect candidate results from large-scale search engines (\eg Google Search). All retrieved images undergo automatic validity filtering to remove corrupted or non-loadable samples, followed by perceptual hashing (pHash) based near-duplicate removal to ensure dataset diversity. For the text-based branch, we generate precise and subject-focused captions using the Qwen2.5-VL-72B model~\cite{bai2025qwen2} (the captioning prompt is provided in \cref{sec:prompts_3_1}). We then use these captions as textual queries to retrieve semantically aligned images. Retrieved results are processed with the same filtering and pHash-based deduplication pipeline. This combined retrieval process guarantees that both visual-query and text-query expansions contribute clean, diverse, and complementary samples for building a comprehensive source image pool.

\paragraph{Synthesis-based Expansion.}
The synthesis-based expansion can be broadly divided into three steps: detailed captioning, variant caption generation, and image synthesis. In the detailed captioning step, we designed a hierarchical description of the image, encompassing seven aspects: \textit{foreground, midground, background, style, lighting and atmosphere, composition and relationships, and visual focus and perspective}. This decomposition allows for fine-grained control over individual image attributes. 
Utilizing Qwen3-8B~\cite{yang2025qwen3}, we then perform ``attribute-swapping'' on the hierarchical descriptions to generate variant captions systematically.
These variant captions are then used by the Qwen-Image~\cite{wu2025qwen} for subsequent image synthesis. 
The detailed prompts are listed in \cref{sec:prompts_3_1}.

\begin{figure*}[!htbp]
  \centering
   \includegraphics[width=0.9\linewidth]{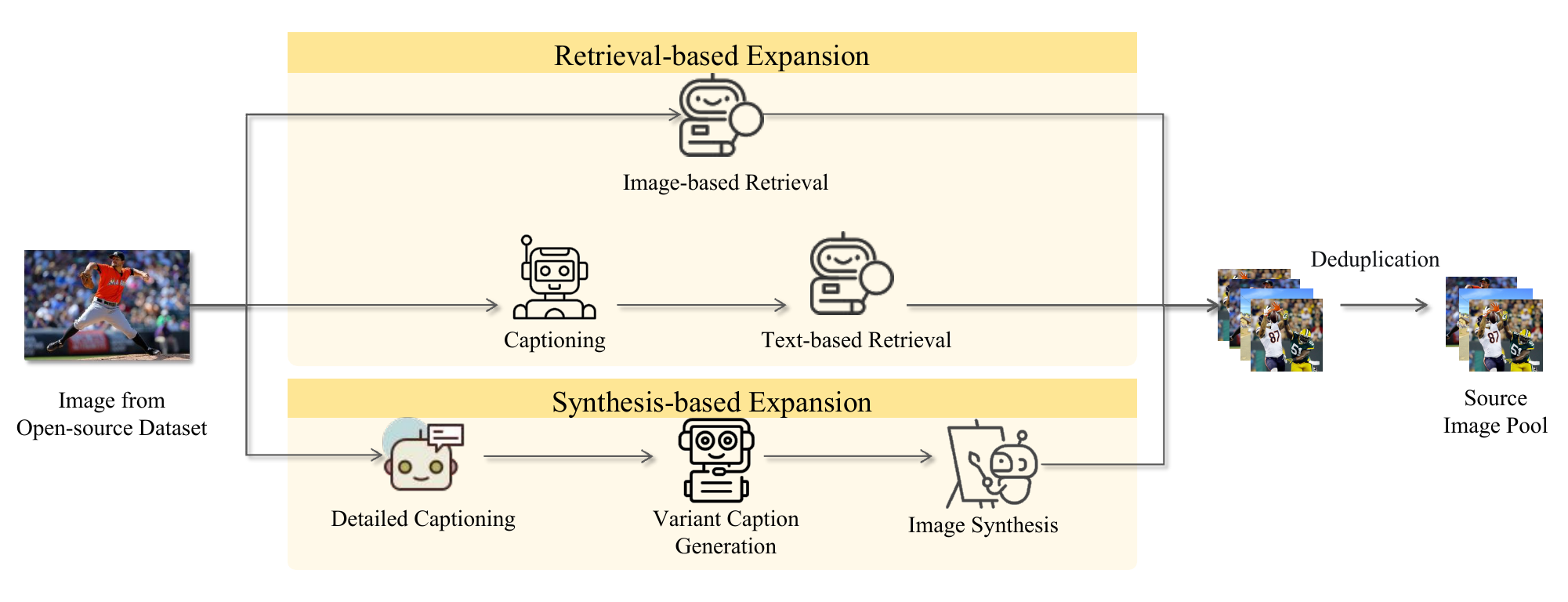}

   \caption{\textbf{Overview of the Source Image Expansion stage in \method.} Retrieval-based expansion performs image-based and text-based retrieval to collect semantically related images, while synthesis-based expansion produces high-fidelity variants, guided by detailed image captions. }
   \label{fig:figA}
\end{figure*}

\subsection{Adaptive Multi-Agent Editing Synthesis}

\paragraph{Task Router.}
The task router, based on Qwen2.5VL-72B, is responsible for assigning images to appropriate workflows. The prompt designed in \cref{sec:prompts_3_2} is to instruct the task router to remove tasks that are unsuitable for the current image.

\paragraph{Instruction Agents.}
A series of instruction agents are designed for task-specific instruction generation, including task-specific instruction agents and an instruction rewrite agent. For the task-specific instruction agents, the image, the detailed definition of the task, guidelines and in-context learning examples are provided, enabling diverse instruction generation. We further analysis the semantic diversity in \cref{sec:diversity_analysis}. Please refer to the prompts for each agent in \cref{sec:prompts_3_2}.

\begin{table*}[htbp]
\centering
\caption{
    \textbf{Detailed definition of editing tasks}.
    }
\label{tab:taxonomy}
% \resizebox{0.95\linewidth}{!}{
\small
\begin{tabularx}{\textwidth}{ccX}
\toprule
\textbf{Category} & \textbf{Editing Task} & \textbf{Definition} \\
\midrule
\multirow{8}{*}{\rotatebox{90}{Global-level Editing}} & Style Transfer & Converts the overall artistic appearance of an image into a target style while preserving its structural and semantic content. \\

& Tone Adjustment  & Adjusts global tonal parameters such as brightness, contrast, saturation, or color temperature. \\

& Viewpoint Transformation & Modifies the camera viewpoint or perspective geometry to present the scene from a new spatial angle. \\

& Background Replacement & Replaces the entire background of an image while maintaining the integrity of the foreground objects. \\
\midrule
\multirow{10}{*}{\rotatebox{90}{Object-level Editing}} & Object Addition & Inserts new object instances into the scene while maintaining coherent spatial relationships and lighting conditions. \\
& Object Removal & Removes designated objects and reconstructs the occluded background to preserve scene realism and continuity. \\

& Object Replacement & Substitutes an existing object with another object of similar semantics, ensuring consistency in scale, pose, and contextual relevance. \\

& Action Editing & Modifies the pose, action, or behavioral state of animate subjects (\textit{e.g.}, humans or animals). \\

& Part Extraction & Extracts specific parts or sub-regions from a complex object, enabling fine-grained manipulation or recomposition. \\
\midrule
\multirow{10}{*}{\rotatebox{90}{Object Attribute Editing}}
& Color Change & Alters the color attributes of a specific object or region while preserving shading and material coherence. \\
& Material Change & Modifies or replaces the surface texture or material properties of an object to achieve a different visual appearance. \\
& Visual Beautification & Enhances or stylizes the appearance of animate subjects while maintaining identity consistency and structural realism. \\
& Count Change & Adjusts the number of primary objects in the scene, including duplication, reduction, or redistribution. \\
& Size Change & Manipulates the scale or relative size of an object while preserving its geometric proportion and contextual alignment. \\
\midrule
\multirow{10}{*}{\rotatebox{90}{Text-aware Editing}} 
& Movie Poster Text Editing & Replaces textual content appearing in movie posters while preserving stylistic coherence and typography consistency. \\
& GUI Interface Text Editing & Modifies textual elements in application interfaces—such as labels or button names while maintaining layout integrity and interaction semantics. \\
& Object Surface Text Editing & Alters text printed on object surfaces (\textit{e.g.}, daily goods, clothing) while preserving material properties and surface curvature. \\
& Building Surface Text Editing & Edits text on architectural structures (\textit{e.g.}, road signs, or billboards) while ensuring geometric alignment and integration with the built environment. \\
\midrule
\multirow{8}{*}{\rotatebox{90}{\makecell{Knowledge-infused\\Reasoning Editing}}} 
& Perceptual Reasoning 
& Performs logically consistent modifications to natural images based on causal, spatial, or functional relationships inferred from the scene. \\

& Symbolic Reasoning 
& Applies reasoning-driven edits to abstract, symbolic, or synthetic visual scenes, ensuring internal logical consistency. \\

& Social Knowledge Reasoning 
& Conducts edits guided by cultural norms, social semantics, or commonsense human conventions to ensure socially coherent outcomes. \\

& Scientific Knowledge Reasoning 
& Produces scientifically valid edits constrained by physical, biological, or chemical principles, ensuring adherence to real-world scientific laws. \\
\midrule
\multirow{2}{*}{-} 
& Compositional Editing 
& Complex edits composed of multiple atomic editing instructions (\textit{e.g.}, Object Addition, Color Change, \etc). \\
\bottomrule
\end{tabularx}
% }
\end{table*}

\section{Supplementary Analysis of \dataset}
\subsection{Construction Efficiency}
The entire construction process, including source expansion, multi-agent synthesis, and multi-dimensional verification, consumed approximately 125k H800 GPU hours. While the operational cost is estimated at \$250k, this represents a significant reduction (over 75\%) compared to the projected cost of using proprietary commercial APIs (\eg, GPT-4o), which would exceed \$1M for a dataset of this scale. This efficiency validates \method as a scalable and cost-effective paradigm for open-source data generation.

\subsection{Detailed Edit Type Definition}
To ensure comprehensive coverage of editing scenarios, we define 23 fine-grained tasks across 6 functional categories. Detailed definitions and their corresponding scopes are provided in \cref{tab:taxonomy}. For qualitative visualizations of each task, please refer to \cref{sec:visualization_task}.

\subsection{Filtering Efficacy}
To demonstrate the robustness of our open-source agentic pipeline, we report the score distribution for the three evaluation dimensions, \textit{Instruction Following} ($F$), \textit{Edit Consistency} ($C$), and \textit{Generation Quality} ($Q$), both before and after our rigorous filtering process.

As shown in \cref{tab:score-dist}, \dataset (pre-filtered) already exhibits a high baseline quality, with 82.8\% of samples achieving the maximum score in instruction alignment. After applying our 3/2/2 threshold ($F=3, C\geq2, Q\geq2$), the final \dataset achieves a near-perfect alignment profile. More importantly, \cref{tab:joint-dist-transpose} highlights the joint distribution $(F, C, Q)$. Notably, \textbf{85.3\%} of our final samples attain the ``perfect triplet'' (3, 3, 3), surpassing even the GPT-4o-based ShareGPT-4o-Image (81.3\%) and significantly outperforming other open-source alternatives like OmniEdit (63.5\%) and UltraEdit (44.7\%). These statistics underscore that our multi-agent framework, coupled with task-aware verification, can yield data that rivals or exceeds the quality of proprietary commercial pipelines.

\begin{table*}[ht]
\centering
\caption{\textbf{Score distribution of each dimension across different datasets.}}
\label{tab:score-dist}
\resizebox{\linewidth}{!}{
\begin{tabular}{l|ccc|ccc|ccc}
\hline % \toprule
\multirow{2}{*}{\textbf{Dataset}} 
  & \multicolumn{3}{c|}{\textbf{Instruction Alignment}} 
  & \multicolumn{3}{c|}{\textbf{Edit Consistency}} 
  & \multicolumn{3}{c}{\textbf{Generation Quality}} \\
 & 1 & 2 & 3 & 1 & 2 & 3 & 1 & 2 & 3 \\ \hline
% \midrule
ScaleEdit                 & 0.0\%  & 0.0\%  & \textbf{100.0\%} & 0.0\%  & 14.2\% & \underline{85.8\%} & 0.0\%  & 23.5\% & \underline{76.5\%} \\
ScaleEdit (pre-filtered)                & 7.5\%  & 9.7\%  & 82.8\% & 1.6\%  & 14.0\% & 84.4\% & 1.2\%  & 23.2\% & 75.6\% \\ \hline
% \midrule
ShareGPT-4o-Image~\cite{chen2025sharegpt}
                         & 3.9\%  & 8.3\%  & \underline{87.8\%} & 0.4\%  & 9.7\%  & \textbf{89.9\%} & 0.5\%  & 12.1\% & \textbf{87.4\%} \\
OmniEdit~\cite{wei2024omniedit}                 & 5.8\%  & 13.0\% & 81.2\% & 0.4\%  & 24.8\% & 74.8\% & 0.5\%  & 25.3\% & 74.2\% \\
ImgEdit~\cite{ye2025imgedit}                 & 19.9\% & 25.6\% & 54.5\% & 3.7\%  & 38.4\% & 58.0\% & 2.4\%  & 49.2\% & 48.4\% \\
UltraEdit~\cite{zhao2024ultraedit}                & 10.7\% & 12.1\% & 77.2\% & 2.5\%  & 30.8\% & 66.7\% & 1.9\%  & 49.3\% & 48.8\% \\
SEED-Data-Edit~\cite{ge2024seeddataedittechnicalreporthybrid}           & 48.0\% & 10.5\% & 41.6\% & 6.8\%  & 43.6\% & 49.7\% & 8.4\%  & 66.0\% & 25.6\% \\
AnyEdit~\cite{yu2025anyedit}                  & 20.5\% & 17.2\% & 62.4\% & 6.6\%  & 45.9\% & 47.6\% & 5.3\%  & 55.0\% & 39.7\% \\ \hline
% \bottomrule
\end{tabular}
}
\end{table*}

\begin{table*}[ht]
\centering
\caption{
\textbf{The distribution of joint score combinations across datasets.}
Each tuple denotes \textit{(F, C, Q)}.
}
\label{tab:joint-dist-transpose}
\resizebox{\linewidth}{!}{
\begin{tabular}{c|cccccccccc}
\hline % \toprule
\textbf{Dataset}
& \textbf{(3,3,3)}
& \textbf{(3,3,2)}
& \textbf{(2,2,2)}
& \textbf{(1,2,2)}
& \textbf{(3,2,2)}
& \textbf{(2,3,2)}
& \textbf{(2,3,3)}
& \textbf{(3,2,3)}
& \textbf{(2,2,3)}
& \textbf{(1,1,1)} \\ \hline
% \midrule
ScaleEdit
& 85.3\% & 10.4\% & 0.0\% & 0.0\% & 2.7\% & 0.0\% & 0.0\% & 1.7\% & 0.0\% & 0.0\% \\

ScaleEdit (pre-filtered) 
& 70.6\% & 8.6\% & 4.5\% & 4.2\% & 2.2\% & 2.4\% & 1.8\% & 1.4\% & 1.1\% & 0.9\% \\ \hline

% \midrule

ShareGPT-4o-Image~\cite{chen2025sharegpt}
& 81.3\% & 3.7\% & 3.3\% & 2.1\% & 1.2\% & 1.4\% & 2.5\% & 1.6\% & 1.1\% & 0.3\% \\

OmniEdit~\cite{wei2024omniedit}
& 63.5\% & 7.3\% & 7.6\% & 4.1\% & 4.4\% & 1.4\% & 1.9\% & 6.0\% & 2.1\% & 0.3\% \\

ImgEdit~\cite{ye2025imgedit}
& 37.5\% & 9.0\% & 15.3\% & 11.5\% & 6.1\% & 4.2\% & 3.8\% & 1.9\% & 2.3\% & 2.0\% \\

UltraEdit~\cite{zhao2024ultraedit}
& 44.7\% & 19.2\% & 9.1\% & 7.1\% & 11.4\% & 1.3\% & 1.0\% & 1.9\% & 0.8\% & 1.6\% \\

SEED-Data-Edit~\cite{ge2024seeddataedittechnicalreporthybrid}
& 22.0\% & 14.6\% & 6.6\% & 28.8\% & 4.3\% & 2.9\% & 0.8\% & 0.6\% & 0.2\% & 4.8\% \\

AnyEdit~\cite{yu2025anyedit}
& 32.2\% & 12.0\% & 13.7\% & 11.7\% & 13.3\% & 1.5\% & 0.7\% & 4.8\% & 1.3\% & 4.5\% \\ \hline

% \bottomrule
\end{tabular}}
\end{table*}

\subsection{Semantic Diversity Analysis}
\label{sec:diversity_analysis}

To mitigate potential mode collapse, our instruction agents are explicitly prompted to generate content-adaptive instructions based on the visual context of each input image (\cref{sec:prompts_3_2}). We evaluate the resulting diversity through both fine-grained entity distribution and global semantic manifold analysis.

\paragraph{Target Entity Distribution.}
For each task, we uniformly sampled 10k instances and extracted the target entities (\eg, objects to add/remove, the target colors/materials/styles) using Part-of-Speech (POS) tagging combined with a suite of task-specific syntactic rules. We then computed the vocabulary size and the concentration ratios of the Top-1, Top-5, Top-10, and Top-20 most frequent entities. As summarized in \cref{tab:diversity_stats}, the distribution of target entities follows three distinct patterns: 
(1) Open-Ended Editing. For tasks like \textit{Object Addition} and \textit{Object Replacement}, the vocabulary size is large (up to 8k unique entities), with common objects forming a small head. The majority of entities exhibit a long-tail distribution, rigorously proving that the agents generate highly diverse, non-repetitive instructions.
(2) Context-Driven Editing. Tasks such as \textit{Text-aware} and \textit{Knowledge-infused Reasoning} show high entity diversity, with vocabulary sizes often surpassing 6k. These tasks have a low concentration in the Top-10 (below 10\%), indicating that instruction generation is highly dependent on the specific input context rather than parametric priors. 
(3) Attribute-Constrained Editing. For tasks like \textit{Color Change} and \textit{Material Change}, the vocabulary is smaller, and the Top-10 concentration is higher (exceeding 50\%). This reflects the limited range of primary attributes used in the real world, though the agents still generate diverse variations within these constraints.

\begin{table}[htbp]
\centering
\caption{\textbf{Target Entity Distribution.} 
    The Top-$K$ columns indicate the cumulative percentage of the $K$ most frequent entities. \textit{Vocab Size} represents the total number of unique core entities extracted for that task. 
    The data demonstrates a healthy long-tail distribution for open-ended tasks and extreme diversity for context-driven tasks.
    }
\label{tab:diversity_stats}
\resizebox{0.75\linewidth}{!}{
% \small
\begin{tabular}{lccccc}
\toprule
\textbf{Edit Subtask} & \textbf{Top-1 (\%)} & \textbf{Top-5 (\%)} & \textbf{Top-10 (\%)} & \textbf{Top-20 (\%)} & \textbf{Vocab Size} \\
\midrule
\multicolumn{6}{l}{\cellcolor{gray!15} $\blacktriangledown$ \textbf{Open-Ended Semantic Editing}} \\
Background Replacement & 9.81  & 22.21 & 29.58 & 38.81 & 2456 \\
Object Addition        & 1.48  & 5.53  & 8.86  & 12.50 & 6222 \\
Object Removal         & 10.19 & 21.41 & 25.44 & 30.03 & 3987 \\
Object Replacement     & 5.17  & 10.19 & 12.60 & 16.01 & 5590 \\
Action Editing         & 2.54  & 10.12 & 14.61 & 21.09 & 3764 \\
Part Extraction        & 1.33  & 3.07  & 3.96  & 5.46  & 7926 \\
\midrule
\multicolumn{6}{l}{\cellcolor{gray!15} $\blacktriangledown$ \textbf{Context-Driven \& Complex Editing}} \\
Movie Poster Text Editing      & 0.52  & 2.27  & 3.43  & 5.10  & 7035 \\
GUI Interface Text Editing     & 2.00  & 7.22  & 11.05 & 16.42 & 5799 \\
Object Surface Text Editing   & 0.53  & 2.43  & 3.95  & 6.19  & 6331 \\
Building Surface Text Editing  & 0.44  & 1.86  & 3.32  & 5.30  & 6847 \\
Perceptual Reasoning   & 1.68  & 6.58  & 9.31  & 13.35 & 4608 \\
Symbolic Reasoning     & 3.82  & 11.38 & 18.45 & 26.97 & 4198 \\
Social Knowledge Reasoning       & 2.07  & 6.46  & 9.55  & 14.60 & 3226 \\
Scientific Knowledge Reasoning   & 1.18  & 3.71  & 5.53  & 8.43  & 5546 \\
Compositional Editing  & 2.33  & 6.02  & 7.68  & 9.19  & 8202 \\
\midrule
\multicolumn{6}{l}{\cellcolor{gray!15} $\blacktriangledown$ \textbf{Attribute-Constrained Editing }} \\
Style Transfer         & 8.47  & 15.37 & 21.40 & 28.54 & 1820 \\
Tone Adjustment        & 4.32  & 18.57 & 26.92 & 35.17 & 1956 \\
Viewpoint Transformation    & 13.87 & 29.20 & 35.86 & 41.45 & 972  \\
Color Change           & 12.60 & 38.90 & 55.52 & 72.62 & 605  \\
Material Change        & 16.18 & 41.17 & 53.25 & 67.07 & 753  \\
Visual Beautification  & 10.60 & 25.01 & 31.74 & 38.97 & 2926 \\
Count Change           & 3.65  & 12.13 & 19.44 & 29.68 & 332  \\
Size Change            & 7.10  & 16.70 & 20.07 & 23.14 & 1950 \\
\bottomrule
\end{tabular}
}
\end{table}

\begin{figure*}[!htbp]
  \centering
   \includegraphics[width=0.8\linewidth]{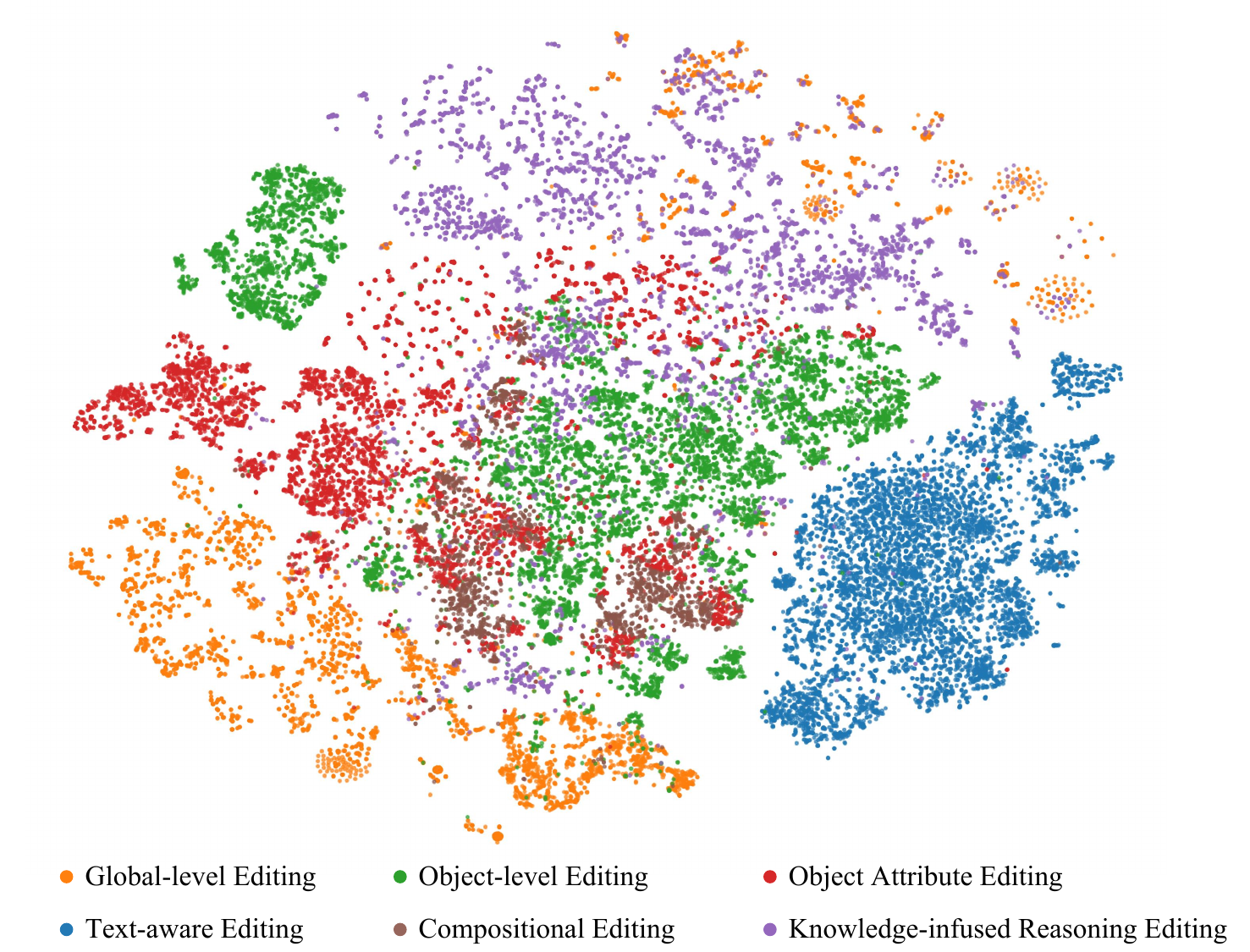}

   \caption{\textbf{T-SNE projection of semantic diversity across editing tasks.}  
   The absence of highly isolated or overly dense clusters demonstrates our dataset’s semantic richness and avoids repetitive or formulaic instruction generation.
   }
   \label{fig:tsne}
\end{figure*}

\paragraph{Semantic Manifold Analysis.}
To evaluate the global semantic diversity in \dataset, we randomly sampled 2k instances from each of the 23 tasks and extracted their sentence embeddings using All-Mpnet-Base-V2~\cite{song2020mpnet}. The resulting T-SNE projection (\cref{fig:tsne}) offers a panoramic view of the dataset's semantic coverage, characterized by the following observations:
(1) Context-Sensitive Manifolds. Tasks like \textit{Knowledge-infused Reasoning Editing} (purple) and \textit{Global-level Editing} (orange) are widely dispersed in the semantic space. These instructions are dynamically tailored to the unique visual contexts of source images rather than being confined to repetitive linguistic templates.
(2) Thematic Clustering and Convergence. Conversely, we observe meaningful overlap among \textit{Object-level} (green), \textit{Object Attribute} (red), and \textit{Compositional} (brown) editing. This semantic convergence is physically grounded since \textit{Compositional} tasks naturally bridge the gap by simultaneously manipulating objects and their properties. Notably, \textit{Part Extraction} (top-left) emerges as a distinct semantic island, which is consistent with its specialized focus on fine-grained decomposition rather than generic object manipulation.
(3) Absence of Mode Collapse. Crucially, the projection is devoid of isolated or hyper-dense clusters, which are often signatures of formulaic or template-based generation. The continuous and broad distribution across the entire manifold suggests that our framework effectively mitigates mode collapse, yielding a diverse corpus of context-aware instructions.

\subsection{Detailed Image Aspect Ratio}

To better characterize the resolution properties of our dataset, we computed the distribution of image aspect ratios, defined as the ratio between width and height for each sample (using the first resolution entry when multiple resolutions are provided). 
As summarized in \cref{tab:aspect_ratio_percentage}, the dataset exhibits a strong concentration around standard photographic formats, with the 1.50 (3:2) ratio representing the largest proportion, followed by 0.67 (2:3) and 1.33 (4:3). The diversity in aspect ratios ensures that models trained on \dataset remain robust to various input resolutions and compositions, while the exclusion of extreme outliers prevents training instability.

\begin{table}[htbp]
\centering
\caption{\textbf{Aspect Ratio Distribution.}}
\label{tab:aspect_ratio_percentage}
\small
\begin{tabular}{cc|cc|cc|cc}
    \hline
    \textbf{Ratio} & \textbf{Percentage} &
    \textbf{Ratio} & \textbf{Percentage} &
    \textbf{Ratio} & \textbf{Percentage} &
    \textbf{Ratio} & \textbf{Percentage} \\ \hline
    0.56 & 0.57\%  & 0.65 & 0.39\%  & 0.66 & 1.36\%  & 0.67 & 10.47\% \\
    0.70 & 0.42\%  & 0.71 & 0.61\%  & 0.80 & 0.77\%  & 1.00 & 8.81\% \\
    1.25 & 0.41\%  & 1.33 & 8.92\%  & 1.34 & 0.44\%  & 1.36 & 0.40\% \\
    1.39 & 0.59\%  & 1.40 & 0.38\%  & 1.49 & 1.13\%  & 1.50 & 34.67\% \\
    1.51 & 3.00\%  & 1.60 & 0.40\%  & 1.78 & 4.56\%  & --   & -- \\ \hline
\end{tabular}
\end{table}

\section{Implementation Details of Experiments}

\subsection{Training Details}

\paragraph{Main Experiments.}
We trained UniWorld-V1~\cite{lin2025uniworld} using 32 A100 GPUs with a default resolution of 512$\times$512. The entire training process took approximately 100 hours. Similarly, Bagel~\cite{deng2025bagel} was trained on 32 A100 GPUs with the default resolution (longest side $\leq$ 1024), spanning roughly 350 hours.
Additional training hyperparameters are provided in \cref{tab:training_cfg}. For the other datasets, we exclusively trained UniWorld-V1 and Bagel on the editing part to ensure a fair comparison.

\begin{table}[ht]
\caption{\textbf{Training Parameters for UniWorld-V1~\cite{lin2025uniworld} and Bagel~\cite{deng2025bagel}.} LR scheduler denotes learning rate scheduler.}
\label{tab:training_cfg}
\centering
\small
\begin{tabular}{ccc}
\toprule
\textbf{Parameters}        & \textbf{UniWorld-V1~\cite{lin2025uniworld}}                          & \textbf{Bagel~\cite{deng2025bagel}}                          \\ \midrule
batch size        & 128                                     & 256                                      \\ 
iterations        & 89640                                  & 44820              \\
learning rate     & $10^{-5}$                               & $2\times 10^{-5}$                                    \\ 
warmup rate       & 0.05                                    & 0.05                                    \\ 
optimizer         & AdamW                                   & AdamW                                   \\ 
LR scheduler      & Const                       & Const            \\ 
\bottomrule
\end{tabular}
\end{table}

\paragraph{Ablation Studies.}
We conducted a series of ablation studies on both UniWorld-V1 and Janus-Pro~\cite{chen2025janus}, each utilizing distinct training configurations, as detailed in \cref{tab:ablation_training_cfg}.

\begin{table}[ht]
\small
\caption{\textbf{Training Parameters for Ablation Studies.} LR scheduler denotes learning rate scheduler.}
\label{tab:ablation_training_cfg}
\centering
\begin{tabular}{ccc}
\toprule
\textbf{Parameters}        & \textbf{UniWorld-V1~\cite{lin2025uniworld}}                          & \textbf{Janus-Pro~\cite{chen2025janus}}                          \\ \midrule
batch size        & 128                                     & 32                                      \\ 
iterations        & 7,812                                   & 31,250                   \\
gradient accumulation & 1 & 4 \\
learning rate     & $10^{-5}$                                    & $2\times 10^{-6}$                               \\ 
warmup rate       & 0.05                                    & 0.05                                    \\ 
optimizer         & AdamW                                   & AdamW                                   \\ 
LR scheduler      & Const                       & Cosine Annealing            \\ 
GPUs              & 32 A100 GPUs                            & 8 A100 GPUs \\
\bottomrule
\end{tabular}
\end{table}

\subsection{More Results}

\paragraph{Equal-scale Comparison.}
In the main paper, we have provided the overall scores of ImgEdit~\cite{ye2025imgedit} and GEdit~\cite{liu2025step1x} in the equal-scale ablation study. Here, we further list the scores of each subtask in \cref{tab:appendix_ablation_gedit} and \cref{tab:appendix_ablation_imgedit}.

\begin{table*}[ht]
\centering
\caption{\textbf{Ablation on instruction rewriting.}
ApprCons. denotes Appearance Consistency sub-dimension, while VisPlausi. represents Visual Plausibility sub-dimension.}
\label{tab:ablation_rewrite}
\resizebox{\linewidth}{!}{
\begin{tabular}{clcccccccc}
\toprule
\multirow{2}{*}{\textbf{Model}} & \multirow{2}{*}{\textbf{Training Dataset}}
& \multicolumn{4}{c}{\textbf{RISEBench~\cite{zhao2025envisioning}}}
& \multicolumn{4}{c}{\textbf{KRIS Bench~\cite{wu2025kris}}} \\
\cmidrule(lr){3-6} \cmidrule(lr){7-10}
 & & \textbf{Reasoning} & \textbf{ApprCons.} & \textbf{VisPlausi.} & \textbf{Overall}
 & \textbf{Factual} & \textbf{Conceptual} & \textbf{Procedural} & \textbf{Overall} \\
\midrule
\multirow{2}{*}{UniWorld-V1~\cite{lin2025uniworld}}
& w/o rewrite & 22.75 & 57.35 & 90.26 & 3.05  & 54.86 & 59.65 & 34.73 & 52.19 \\
& w/~rewrite  & 24.38 & 64.24 & 90.64 & 4.17  & 56.83 & 63.15 & 35.99 & 54.88 \\
\bottomrule
\end{tabular}
}
\end{table*}

\begin{table*}[htbp]
\centering
\small
\caption{
    \textbf{Ablation study of different datasets with equal scale on GEdit-EN-Full~\cite{liu2025step1x}}
    The best score is shown in \textbf{bold}, and the second best is \underline{underlined}, for UniWorld-V1 and Janus-Pro, respectively.
}
\label{tab:appendix_ablation_gedit}
\resizebox{0.9\linewidth}{!}{
\begin{tabular}{llcccccccccccc}
\toprule
& \textbf{Dataset} & \textbf{Background} & \textbf{Color} & \textbf{Material} & \textbf{Motion} & \textbf{Portrait} & \textbf{Style} & \textbf{Add} & \textbf{Remove} & \textbf{Replace} & \textbf{Text} & \textbf{Tone} & \textbf{Avg} \\
\midrule
\multirow{15}{*}{\rotatebox{90}{\textbf{UniWorld-V1} \cite{lin2025uniworld}}} & Pre-trained & 4.92 & 6.37 & 4.79 & 1.85 & 4.03 & 5.64 & 7.23 & 6.17 & 5.70 & 1.15 & 5.54 & 4.85 \\
& \multicolumn{13}{>{\columncolor{gray!15}}l}{ $\blacktriangledown$ \emph{w/ Commercial Datasets}} \\
& OpenGPT-4o-Image \cite{chen2025opengpt} & \textbf{6.88} & 6.53 & \underline{6.24} & 6.48 & 6.07 & \textbf{7.11} & 7.01 & \textbf{6.19} & 5.81 & \underline{2.23} & 6.09 & 6.06 \\
& ShareGPT-4o-Image \cite{chen2025sharegpt} & 6.42 & 7.60 & \textbf{6.43} & 6.16 & 6.26 & 6.62 & 7.26 & 5.85 & 5.31 & 2.26 & 6.60 & 6.07 \\
& Nano-consistent  \cite{ye2025echo} & \underline{6.76} & 6.10 & 4.23 & 5.55 & 6.31 & 5.41 & 6.44 & 1.44 & 5.24 & 1.36 & 6.39 & 5.02 \\
& Pico-Banana \cite{qian2025pico} & 6.58 & \underline{7.66} & 5.25 & \textbf{6.93} & 6.42 & 5.36 & 6.57 & 6.10 & \textbf{6.01} & \textbf{2.30} & 6.77 & 5.99 \\
& GPT-Image-Edit \cite{wang2025gpt} & 6.12 & \textbf{7.96} & 5.34 & 6.47 & \underline{6.68} & 6.56 & \textbf{7.36} & \underline{6.17} & 5.66 & 1.59 & \underline{7.15} & \underline{6.12} \\
& \multicolumn{13}{>{\columncolor{gray!15}}l}{ $\blacktriangledown$ \emph{w/ Open-source Datasets}} \\
& OmniEdit \cite{wei2024omniedit} & 5.69 & 6.37 & 5.61 & 4.93 & 5.34 & 6.91 & 5.82 & 5.04 & 5.17 & 1.61 & 5.29 & 5.25 \\
& ImgEdit \cite{ye2025imgedit} & 4.32 & 6.47 & 4.96 & 4.27 & 5.33 & 6.20 & 5.61 & 4.47 & 5.12 & 1.64 & 5.53 & 4.90 \\
& NHR-Edit \cite{kuprashevich2025nohumansrequiredautonomoushighqualityimage} & 6.24 & 6.39 & 5.70 & 5.13 & 5.45 & 6.25 & 6.41 & 6.76 & 6.11 & \underline{2.23} & 5.54 & 5.66 \\
& AnyEdit \cite{yu2025anyedit} & 4.56 & 6.54 & 5.53 & 4.89 & 5.44 & 5.86 & 6.25 & 4.99 & \underline{5.99} & 1.72 & 5.32 & 5.19 \\
& UltraEdit \cite{zhao2024ultraedit} & 4.77 & 4.09 & 4.90 & 4.31 & 4.13 & 5.81 & 4.19 & 1.48 & 4.87 & 1.50 & 3.11 & 3.92 \\
& \multicolumn{13}{>{\columncolor{gray!15}}l}{ $\blacktriangledown$ \emph{w/ Our Datasets}} \\
& \textbf{ScaleEdit} & 6.33 & 7.53 & 5.73 & \underline{6.66} & \textbf{6.69} & \underline{6.94} & \underline{7.32} & 6.10 & 5.45 & 1.52 & \textbf{7.40} & \textbf{6.15} \\
\midrule
\multirow{14}{*}{\rotatebox{90}{\textbf{Janus-Pro} \cite{chen2025janus}}} & \multicolumn{13}{>{\columncolor{gray!15}}l}{ $\blacktriangledown$ \emph{w/ Commercial Datasets}} \\
& OpenGPT-4o-Image \cite{chen2025opengpt} & 4.76 & 5.42 & 4.71 & 5.23 & \textbf{5.13} & 5.33 & 4.84 & \underline{2.49} & 4.34 & \underline{3.17} & 4.46 & 4.52 \\
& ShareGPT-4o-Image  \cite{chen2025sharegpt} & 4.47 & 5.47 & \underline{4.94} & 5.64 & 4.40 & \textbf{6.03} & 4.89 & 1.57 & 4.50 & 2.93 & 4.38 & 4.47 \\
& Nano-consistent \cite{ye2025echo} & 3.60 & 2.79 & 3.12 & 4.99 & 3.57 & 3.54 & 3.71 & 0.74 & 2.98 & 2.45 & 3.10 & 3.14 \\
& Pico-Banana \cite{qian2025pico} & 2.17 & 2.09 & 1.86 & 2.91 & 3.43 & 2.67 & 1.39 & 0.80 & 1.81 & 0.88 & 2.36 & 2.03 \\
& GPT-Image-Edit \cite{wang2025gpt} & \underline{5.02} & \underline{5.99} & \textbf{5.67} & \underline{5.67} & \underline{5.10} & \underline{5.62} & \textbf{5.34} & 2.43 & \underline{5.01} & 3.04 & \underline{4.64} & \underline{4.87} \\
& \multicolumn{13}{>{\columncolor{gray!15}}l}{ $\blacktriangledown$ \emph{w/ Open-source Datasets}} \\
& OmniEdit \cite{wei2024omniedit} & 2.48 & 2.22 & 2.10 & 3.19 & 3.53 & 3.51 & 1.44 & 0.89 & 1.12 & 0.83 & 3.15 & 2.22 \\
& ImgEdit \cite{ye2025imgedit} & 2.75 & 1.99 & 2.51 & 2.18 & 2.49 & 3.20 & 1.28 & 0.65 & 1.72 & 0.95 & 1.99 & 1.98 \\
& NHR-Edit \cite{kuprashevich2025nohumansrequiredautonomoushighqualityimage} & 2.62 & 1.69 & 1.58 & 2.77 & 3.44 & 2.49 & 2.04 & 1.14 & 1.11 & 0.98 & 2.40 & 2.02 \\
& AnyEdit \cite{yu2025anyedit} & 2.73 & 2.18 & 2.16 & 3.51 & 3.68 & 3.38 & 2.11 & 0.72 & 1.63 & 1.57 & 3.50 & 2.47 \\
& UltraEdit \cite{zhao2024ultraedit} & 2.86 & 3.54 & 3.53 & 3.97 & 4.64 & 3.86 & 3.17 & 0.82 & 3.27 & 1.67 & 2.98 & 3.12 \\
& \multicolumn{13}{>{\columncolor{gray!15}}l}{ $\blacktriangledown$ \emph{w/ Our Datasets}} \\
& \textbf{ScaleEdit} & \textbf{5.42} & \textbf{6.21} & 4.80 & \textbf{6.28} & 4.88 & 5.59 & \underline{5.04} & \textbf{2.55} & \textbf{5.29} & \textbf{3.27} & \textbf{4.83} & \textbf{4.92} \\
\bottomrule
\end{tabular}
}
\end{table*}

\begin{table*}[htbp]
\centering
\caption{
    \textbf{Ablation study of different datasets with equal scale on ImgEdit-Bench.~\cite{ye2025imgedit}} The best score is shown in \textbf{bold}, and the second best is \underline{underlined}, for UniWorld-V1 and Janus-Pro, respectively.
}
\label{tab:appendix_ablation_imgedit}
\resizebox{0.9\linewidth}{!}{
\begin{tabular}{llcccccccccc}
\toprule
& \textbf{Dataset} & \textbf{Add} & \textbf{Adjust} & \textbf{Extract} & \textbf{Replace} & \textbf{Remove} & \textbf{Background} & \textbf{Style} & \textbf{Hybrid} & \textbf{Action} & \textbf{Avg} \\
\midrule
\multirow{14}{*}{\rotatebox{90}{\textbf{UniWorld-V1} \cite{lin2025uniworld}}} & Pre-trained & 3.82 & 3.64 & 2.27 & 3.47 & 3.24 & 2.99 & 4.21 & 2.96 & 2.74 & 3.26 \\
& \multicolumn{11}{>{\columncolor{gray!15}}l}{ $\blacktriangledown$ \emph{w/ Commercial Datasets}} \\
& OpenGPT-4o-Image \cite{chen2025opengpt} & \textbf{4.18} & \textbf{3.96} & 1.99 & 3.44 & 2.62 & \underline{3.67} & 4.65 & 2.74 & 3.07 & 3.37 \\
& ShareGPT-4o-Image \cite{chen2025sharegpt} & 4.11 & 3.67 & \underline{2.04} & 3.53 & 2.90 & \textbf{3.83} & \underline{4.81} & 2.94 & \textbf{3.62} & 3.49 \\
& Nano-consistent \cite{ye2025echo} & 3.88 & 3.52 & 1.91 & 3.23 & 2.21 & 3.11 & 4.07 & 2.83 & 4.10 & 3.21 \\
& Pico-Banana \cite{qian2025pico} & 4.08 & \underline{3.79} & 1.86 & 3.71 & 3.56 & 3.56 & 4.11 & 2.85 & 3.26 & 3.42 \\
& GPT-Image-Edit \cite{wang2025gpt} & \underline{4.16} & 3.27 & 1.98 & \textbf{3.81} & \underline{3.78} & 3.55 & \textbf{4.87} & \underline{3.16} & 3.04 & \textbf{3.51} \\
& \multicolumn{11}{>{\columncolor{gray!15}}l}{ $\blacktriangledown$ \emph{w/ Open-source Datasets}} \\
& OmniEdit \cite{wei2024omniedit} & 3.80 & 3.71 & 2.00 & 3.02 & 2.96 & 2.95 & 4.33 & 2.46 & 2.50 & 3.08 \\
& ImgEdit \cite{ye2025imgedit} & 3.78 & 2.91 & \textbf{2.05} & 3.44 & 2.84 & 2.56 & 4.57 & 2.58 & 2.20 & 2.99 \\
& NHR-Edit \cite{kuprashevich2025nohumansrequiredautonomoushighqualityimage} & 3.87 & 3.07 & 1.81 & \underline{3.77} & \textbf{3.82} & 3.00 & 4.31 & 2.75 & 2.33 & 3.19 \\
& AnyEdit \cite{yu2025anyedit} & 3.85 & 3.00 & 1.96 & 3.50 & 2.94 & 2.52 & 4.06 & 2.54 & 2.33 & 2.97 \\
& UltraEdit \cite{zhao2024ultraedit} & 3.95 & 3.58 & 1.84 & 2.59 & 1.31 & 2.72 & 3.91 & 2.29 & 2.92 & 2.79 \\
& \multicolumn{11}{>{\columncolor{gray!15}}l}{ $\blacktriangledown$ \emph{w/ Our Datasets}} \\
& \textbf{ScaleEdit} & 4.13 & 3.76 & 1.99 & 3.74 & 2.99 & 3.54 & 4.71 & \textbf{3.29} & \underline{3.32} & \underline{3.50} \\
\midrule
\multirow{13}{*}{\rotatebox{90}{\textbf{Janus-Pro}}} & \multicolumn{11}{>{\columncolor{gray!15}}l}{$\blacktriangledown$ \emph{w/ Commercial Datasets}} \\
& OpenGPT-4o-Image \cite{chen2025opengpt} & \underline{3.83} & 2.89 & \underline{2.39} & \underline{2.82} & \underline{2.31} & 3.25 & 4.32 & \textbf{2.46} & 3.41 & \underline{3.08} \\
& ShareGPT-4o-Image \cite{chen2025sharegpt} & 3.69 & 2.89 & 2.21 & 2.74 & 1.54 & 3.10 & \textbf{4.64} & 2.16 & \textbf{3.70} & 2.96 \\
& Nano-consistent \cite{ye2025echo} & 3.76 & 2.28 & 1.98 & 2.45 & 1.09 & 2.50 & 3.79 & 1.54 & \underline{3.66} & 2.56 \\
& Pico-Banana \cite{qian2025pico} & 2.49 & 2.31 & 1.97 & 1.42 & 1.10 & 1.66 & 3.24 & 1.33 & 2.56 & 2.01 \\
& GPT-Image-Edit \cite{wang2025gpt} & 4.01 & \underline{3.04} & 2.14 & 2.63 & 2.04 & \textbf{3.36} & \underline{4.46} & 2.17 & 3.44 & 3.03 \\
& \multicolumn{11}{>{\columncolor{gray!15}}l}{ $\blacktriangledown$ \emph{w/ Open-source Datasets}} \\
& OmniEdit \cite{wei2024omniedit} & 1.76 & 2.37 & 1.99 & 1.30 & 1.11 & 1.42 & 3.43 & 1.26 & 2.63 & 1.92 \\
& ImgEdit \cite{ye2025imgedit} & 2.33 & 1.49 & 1.67 & 2.12 & 1.20 & 1.78 & 3.76 & 1.45 & 2.13 & 1.99 \\
& NHR-Edit \cite{kuprashevich2025nohumansrequiredautonomoushighqualityimage} & 2.47 & 2.33 & 1.99 & 1.40 & 1.14 & 1.93 & 3.01 & 1.39 & 2.37 & 2.00 \\
& AnyEdit \cite{yu2025anyedit} & 3.01 & 2.27 & 1.88 & 2.01 & 1.28 & 1.72 & 3.35 & 1.13 & 2.52 & 2.13 \\
& UltraEdit \cite{zhao2024ultraedit} & 3.28 & 2.56 & 1.94 & 1.71 & 1.09 & 2.10 & 3.63 & 1.40 & 2.71 & 2.27 \\
& \multicolumn{11}{>{\columncolor{gray!15}}l}{ $\blacktriangledown$ \emph{w/ Our Datasets}} \\
& \textbf{ScaleEdit} & \textbf{3.85} & \textbf{3.09} & \textbf{2.41} & \textbf{3.35} & \textbf{2.63} & \underline{3.33} & 4.20 & \underline{2.19} & 3.50 & \textbf{3.17} \\
\bottomrule
\end{tabular}}
\end{table*}

\paragraph{Impact of Instruction Rewriting.} In the main paper, we reported the effect of instruction rewriting on the RISE benchmark~\cite{zhao2025envisioning}, demonstrating that rewritten prompts substantially enhance multi-step visual reasoning. For completeness, Tab.~\ref{tab:ablation_rewrite} presents the full ablation results across both RISEBench and KRIS Bench~\cite{wu2025kris}. These results include all sub-dimensions Reasoning, Appearance Consistency, Visual Plausibility, as well as the benchmark-specific criteria in KRIS Bench (Factual, Conceptual, and Procedural). The complete evaluation confirms the consistent benefits of instruction rewriting across datasets and task types, reinforcing its effectiveness beyond the RISE benchmark alone.

\begin{figure*}[!htbp]
  \centering
   \includegraphics[width=\linewidth]{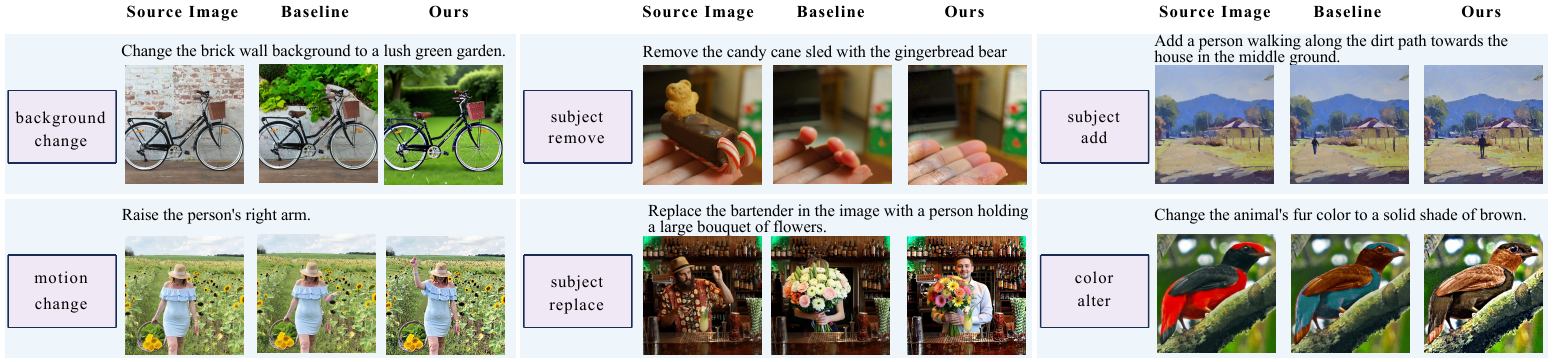}

   \caption{\textbf{Qualitative comparison of Bagel~\cite{deng2025bagel} before and after fine-tuning on \dataset.}  }
   \label{fig:example_final_bagel}
\end{figure*}

\paragraph{Qualitative Results.}
Besides the qualitative comparisons on UniWorld-V1 in the main paper, we additionally provide the qualitative comparisons between baseline Bagel and the model fine-tuned on ScaleEdit in \cref{fig:example_final_bagel}.

\subsection{User Study}
To more accurately evaluate the practical effectiveness of different training datasets in real-world image editing scenarios, we conduct a human evaluation. We randomly sample 50 test cases from general editing benchmarks~\cite{liu2025step1x, ye2025imgedit} and use the models trained on four representative datasets: GPT-Image-Edit~\cite{wang2025gpt}, OpenGPT-4o-Image~\cite{chen2025opengpt}, ImgEdit~\cite{ye2025imgedit}, and ScaleEdit. 

\begin{figure*}[htbp]
    \centering
    \includegraphics[width=0.8\linewidth]{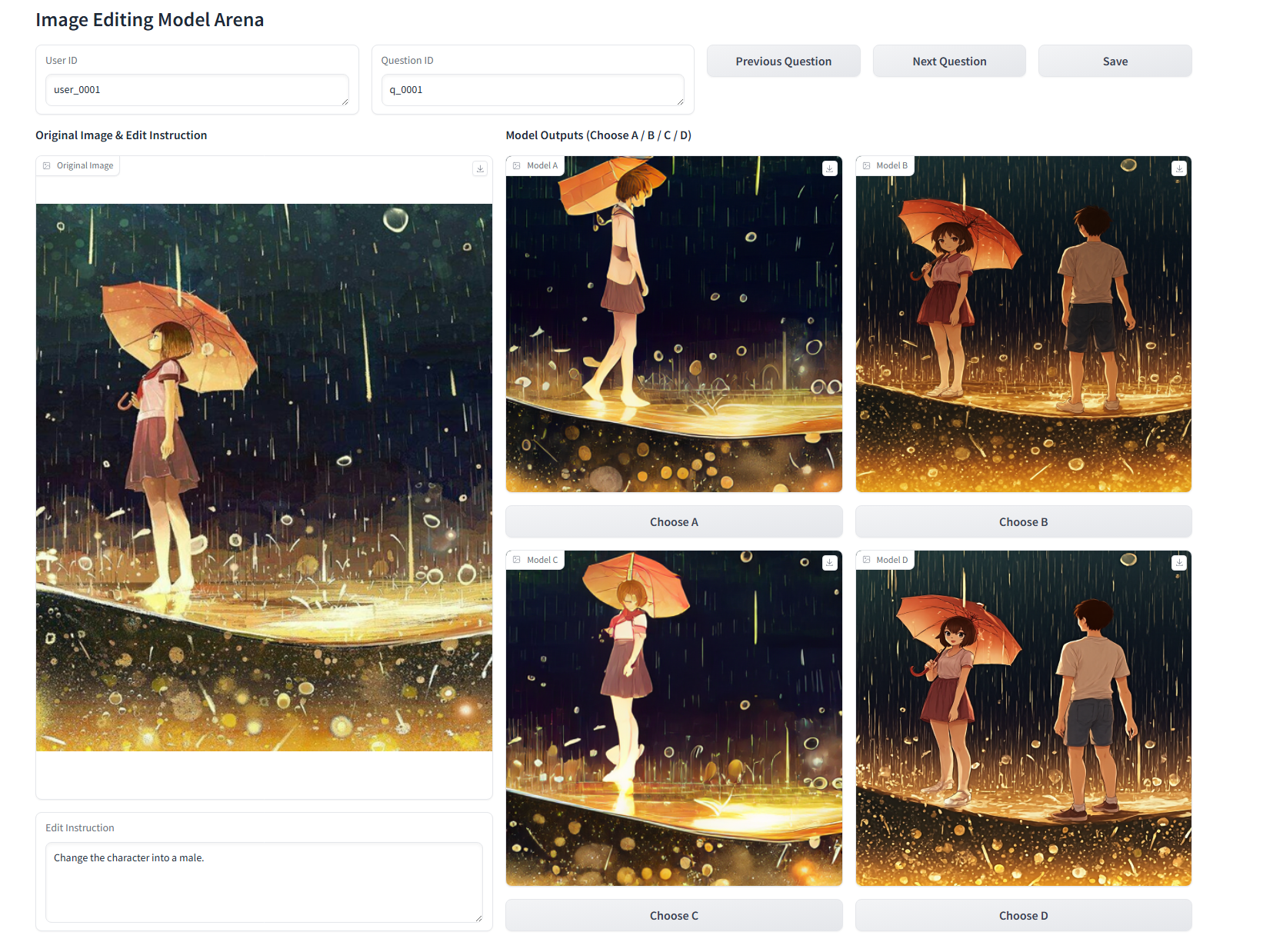}
    \caption{\textbf{User interface used in the user study.} The left panel displays the original image and the corresponding editing instruction, while the right panel shows four candidate edited results from different models, labeled as options A, B, C, and D. Annotators are asked to select the option that best satisfies the editing instruction.}
    \label{fig:user_study_ui}
\end{figure*}

\begin{table}[!thbp]
    \centering
    \caption{\textbf{Human preference results in the user study.} ``Top-1 preference'' denotes the percentage of tasks for which each model is selected as the best edited result.}
    \label{tab:user_study_results}
    \begin{tabular}{lc}
        \toprule
        Training Data & Top-1 preference \\
        \midrule
        GPT-Image-Edit~\cite{wang2025gpt} & 26.8\% \\
        OpenGPT-4o-Image~\cite{chen2025opengpt} & \underline{28.1\%} \\
        ImgEdit~\cite{ye2025imgedit} & 15.6\% \\
        ScaleEdit  & \textbf{29.5\%} \\
        \bottomrule
    \end{tabular}
\end{table}

\section{Visualizations}
\label{sec:visualization}

\subsection{Filter Examples}
\label{sec:visualization_filter}

\begin{figure*}[htbp]
    \centering
    \includegraphics[width=0.85\linewidth]{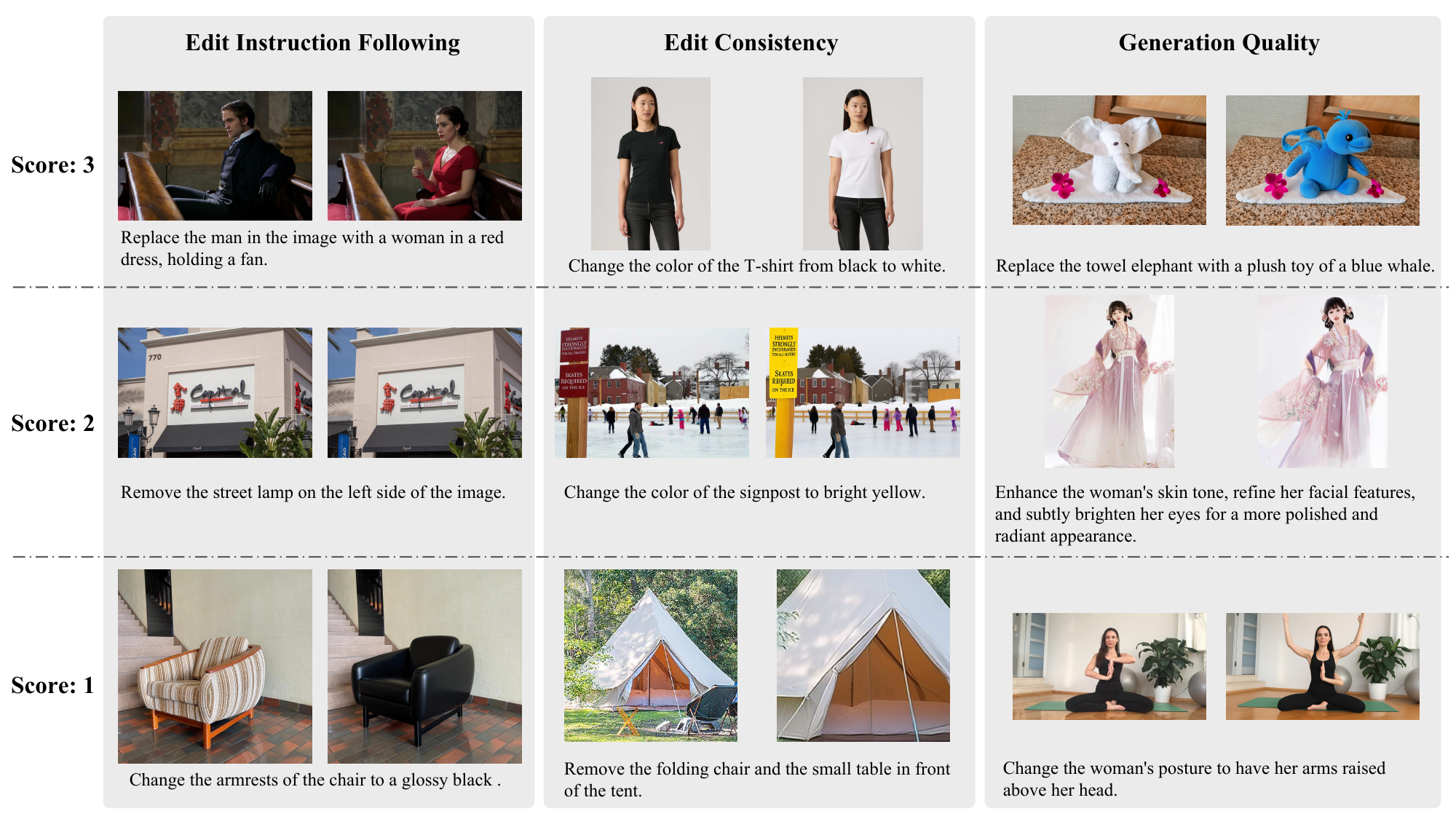}
    \caption{\textbf{Examples corresponding to each score of different metrics.}}
    \label{fig:filter_example_1}
\end{figure*}

We show more visualization results of different levels for the three filtering metrics in \cref{fig:filter_example_1}.

\subsection{Task Examples}
\label{sec:visualization_task}

\begin{figure*}[th]
    \centering
    \includegraphics[width=\linewidth]{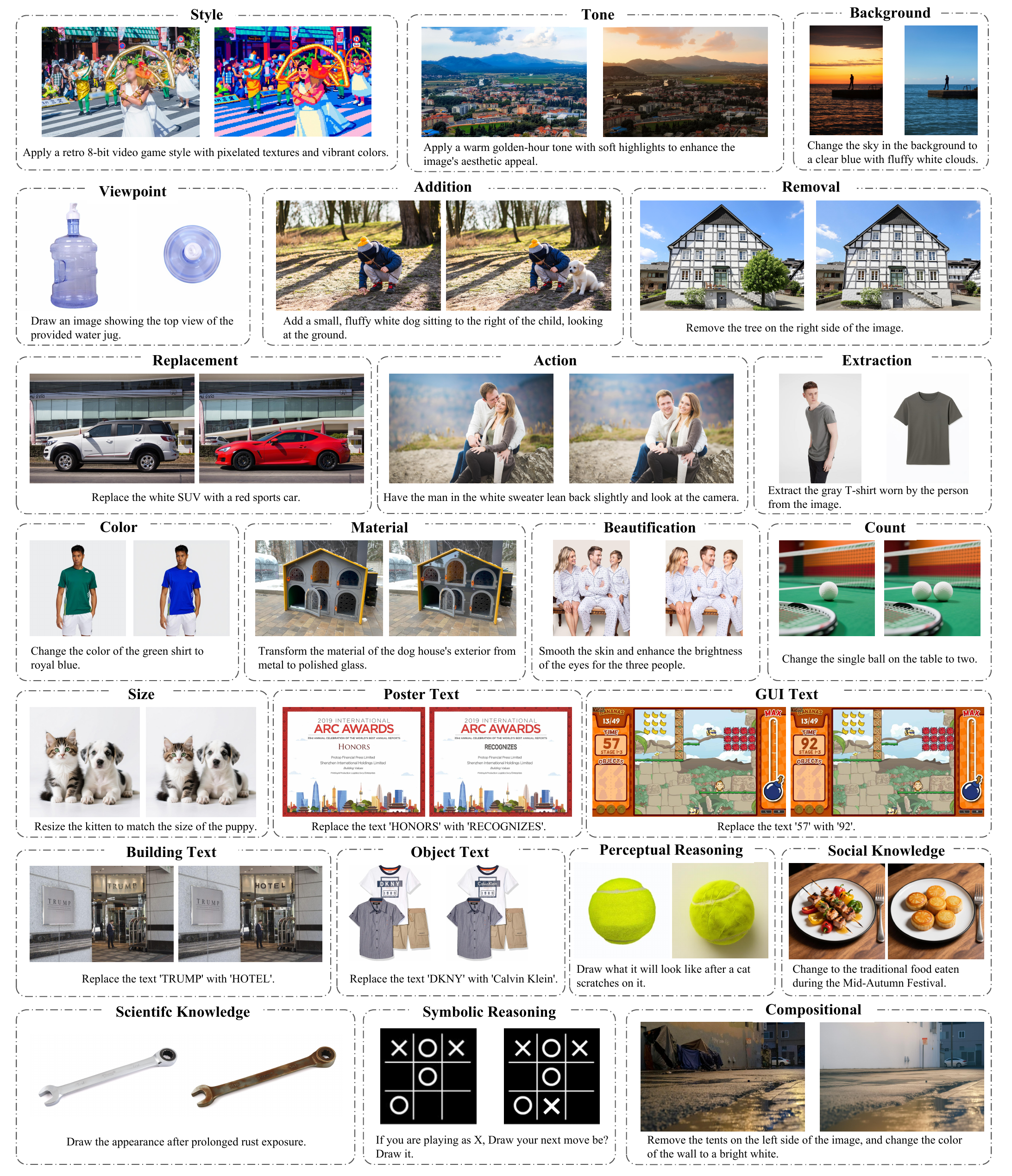}
    \caption{\textbf{Examples across different editing tasks.}}
    \label{fig:task_example}
\end{figure*}

We present more visualization examples related to different tasks in \cref{fig:task_example}.

\clearpage
\newpage

\section{Prompt Templates}

\subsection{Prompts for Source Image Expansion}
\label{sec:prompts_3_1}

\begin{tcolorbox}[
    colback=Emerald!10,
    colframe=cyan!40!black,
    title=\textbf{Prompt for Image Captioning},
    breakable,
    enhanced,
    left=2mm, right=2mm, top=2mm, bottom=2mm,
    fontupper=\footnotesize
]
\scriptsize
\textbf{Please provide an English caption describing the image.} \\
The description should include the main object(s), the scene or environment, and the overall style. \\
If there are multiple objects, specify their spatial relationships. \\
The description should not be overly complex—just a few simple sentences are enough. \\
\textbf{Output only the image description, Do not add extra explanation or output.}
\end{tcolorbox}
\vspace{2em}

\begin{tcolorbox}[
    colback=Emerald!10,
    colframe=cyan!40!black,
    title=\textbf{Prompt for Detailed Image Captioning},
    breakable,
    enhanced,
    left=2mm, right=2mm, top=2mm, bottom=2mm,
    fontupper=\footnotesize
]
\scriptsize
Please provide a detailed description for the provided images from 7 aspects: Foreground, Midground, Background, Style, Lighting and Atmosphere, Composition and Relationships, Visual Focus and Perspective. \\
\textbf{For each aspect, list the distinct elements using numbered points (1., 2., 3., ...).}
\end{tcolorbox}
\vspace{2em}

\begin{tcolorbox}[
    colback=Emerald!10,
    colframe=cyan!40!black,
    title=\textbf{Prompt for Variant Caption Generation},
    breakable,
    enhanced,
    left=2mm, right=2mm, top=2mm, bottom=2mm,
    fontupper=\footnotesize
]
\scriptsize
Below lists the elements of a image from different aspects, please modify the element: \textit{[Specific Element in Detailed Image Caption]}. \\
\textit{[Detailed Image Caption]} \\
\textbf{Output only one line representing the variant element without any other explanation.} 
\end{tcolorbox}
\vspace{2em}

\subsection{Prompts for Adaptive Multi-Agent Editing Synthesis}
\label{sec:prompts_3_2}

% ====================================================
\begin{tcolorbox}[
    colback=Emerald!10,
    colframe=cyan!40!black,
    title=\textbf{Prompt for Task Router},
    breakable,
    enhanced,
    left=2mm, right=2mm, top=2mm, bottom=2mm,
    fontupper=\footnotesize
]
\scriptsize
\textbf{You are an AI assistant responsible for dispatching images to appropriate image editing tasks.}

\vspace{0.5em}

\textbf{Task} \\
Given an image and several candidate editing tasks, please check the N/A condition one by one, and then output the inappropriate tasks.

\vspace{0.5em}

\textbf{Guidelines} \\
The candidate tasks and N/A condition are as follows:
\begin{itemize}
    \item Style Transfer: N/A if the image already has an abstract or highly stylized appearance.
    \item Tone Adjustment: N/A if the image lacks recoverable tonal information (e.g., extremely dark, overexposed).
    \item Viewpoint Transformation: N/A if the image lacks discernible 3D structure (e.g., a solid color image, or a purely abstract pattern).
    \item Background Replacement: N/A if the image has no distinct foreground objects or the entire image is a homogeneous background.
    \item Object Addition: N/A if the image lacks space or a plausible context to realistically insert a new object (such as close-ups or extremely crowded scenes).
    \item Object Removal: N/A if there is no clearly discernible object to be removed.
    \item Object Replacement: N/A if there is no clearly discernible object to be replaced.
    \item Action Editing: N/A if there is no animate subject.
    \item Part Extraction: N/A if the image contains no complex objects with clearly distinguishable sub-parts (e.g., a simple geometric shape or a blurry scene).
    \item Color Change: N/A if there is no specific object or region with a stable, identifiable color (e.g., heavily motion-blurred or color-uniform scenes).
    \item Material Change: N/A if the image contains no objects with discernible surface textures or materials (e.g., a photo of a purely digital drawing).
    \item Visual Beautification: N/A if there is no clearly visible human portrait.
    \item Count Change: N/A if there is no countable subject (e.g., an abstract painting, a photograph of water).
    \item Size Change: N/A if there is no distinct, isolated object whose scale can be altered without breaking scene integrity.
    \item Movie Poster Text Editing: N/A if there is no movie poster or no text clearly associated with a poster in the image.
    \item GUI Interface Text Editing: N/A if there is no GUI interface or no text clearly visible on a GUI interface in the image.
    \item Object Surface Text Editing: N/A if there is no text printed on the surface of a non-building object in the image.
    \item Building Surface Text Editing: N/A if there is no text on an architectural structure (e.g., sign, billboard, wall) in the image.
    \item Perceptual Reasoning: N/A if the image lacks any real-world semantic objects, spatial, or causal relationships to infer (e.g., purely abstract, pattern-based).
    \item Symbolic Reasoning: N/A if there is no abstract, symbolic, or synthetic visual element.
    \item Social Reasoning: N/A if there is no social interaction, cultural context, or human-centric scene (e.g., a photo of a purely natural landscape).
    \item Scientific Reasoning: N/A if the image does not depict any physical, biological, or other scientific processes that could support a scientifically valid edit (e.g., abstract images, sketches with no real-world semantics).
    \item Compositional Editing: N/A if none of the required atomic sub-tasks are applicable.
\end{itemize}

\vspace{0.5em}

\textbf{Output Format} \\
For each task, output one line. For each line, output 'yes' if applicable for the task, 'no' otherwise with a brief explanation.
\end{tcolorbox}
\vspace{2em}

% ====================================================
\begin{tcolorbox}[
    colback=Emerald!10,
    colframe=cyan!40!black,
    title=\textbf{Prompt for Style Transfer Instruction Agent},
    breakable,
    enhanced,
    left=2mm, right=2mm, top=2mm, bottom=2mm,
    fontupper=\footnotesize
]
\scriptsize
\textbf{You are an AI assistant responsible for generating precise and actionable image editing instructions.} \\

\textbf{Task} \\
Given an image and an editing task with its definition, your goal is to create a clear, specific, and unambiguous editing instruction for the image.

Editing task: Style Transfer \\
Definition: Applying an artistic or stylistic transformation to the image. 

\vspace{0.5em}

\textbf{Guidelines} 
\begin{itemize}
    \item Describe the target style in your instruction.
    \item Avoid abstract instructions.
    \item Ensure diverse and creative instructions, either
    \begin{itemize}
        \item focus on the image content, and how it could be changed,
        \item or focus on the image characteristics, choose a reasonable style.
    \end{itemize}
\end{itemize}

\vspace{0.5em}

\textbf{Examples} 
\begin{itemize}
    \item Image Content: A man in a suit standing in the auditorium and speaking. \\
            Output: Convert the image into a pencil sketch effect.
    \item Image Content: A cowboy riding a horse through the jungle. \\
            Output: Add vintage film grain and faded effects.
    \item Image Content: An abandoned automobile manufacturing factory. \\
            Output: Switch the image style to neon-punk.
\end{itemize}

\vspace{0.5em}

\textbf{Output format} \\
Provide exactly one sentence describing the edit instruction. Do not include any other text.
\end{tcolorbox}
\vspace{2em}

% ====================================================
\begin{tcolorbox}[
    colback=Emerald!10,
    colframe=cyan!40!black,
    title=\textbf{Prompt for Tone Adjustment Instruction Agent},
    breakable,
    enhanced,
    left=2mm, right=2mm, top=2mm, bottom=2mm,
    fontupper=\footnotesize
]
\scriptsize
\textbf{You are an AI assistant responsible for generating precise and actionable image editing instructions.} \\

\textbf{Task} \\
Given an image and an editing task with its definition, your goal is to create a clear, specific, and unambiguous editing instruction for the image.

Editing task: Tone Adjustment \\
Definition: Changing the overall mood or atmosphere, e.g., weather, color, time period, or visual effect, etc.

\vspace{0.5em}

\textbf{Guidelines} 
\begin{itemize}
    \item Describe the target mood or atmosphere in your instruction.
    \item Avoid abstract instructions like "sepia-toned", "vintage", etc.
    \item Ensure diverse and creative instructions, either
    \begin{itemize}
        \item focus on the image content, and how it could be changed,
        \item or focus on the image characteristics, choose a reasonable target mood or atmosphere.
    \end{itemize}
\end{itemize}

\vspace{0.5em}

\textbf{Examples} 
\begin{itemize}
    \item Image Content: A red double-decker bus parked by the roadside under a sunny sky. \\
            Output: Change the weather to foggy.
    \item Image Content: A black-and-white image shows a group of people climbing the Great Wall. \\
            Output: Restore and colorize the image.
    \item Image Content: A cozy living room with a sofa and a cat. \\
            Output: Change the time to prehistoric era. 
    \item Image Content: A cashier standing in front of the wine cabinet. \\
            Output: Add a background blur filter.
\end{itemize}

\vspace{0.5em}

\textbf{Output format} \\
Provide exactly one sentence describing the edit instruction. Do not include any other text.
\end{tcolorbox}
\vspace{2em}

\newpage
% ====================================================
\begin{tcolorbox}[
    colback=Emerald!10,
    colframe=cyan!40!black,
    title=\textbf{Prompt for Viewpoint Transformation Instruction Agent},
    breakable,
    enhanced,
    left=2mm, right=2mm, top=2mm, bottom=2mm,
    fontupper=\footnotesize
]
\scriptsize
\textbf{You are an AI assistant responsible for generating precise and actionable image editing instructions.} \\

\textbf{Task} \\
Given an image and an editing task with its definition, your goal is to create a clear, specific, and unambiguous editing instruction for the image.

Editing task: Viewpoint Transformation \\
Definition: Modifies the camera viewpoint or perspective geometry to present the scene from a new spatial angle.

\vspace{0.5em}

\textbf{Guidelines} 
\begin{itemize}
    \item Clearly specify the desired new viewpoint or camera angle in your instruction.
    \item Avoid vague directions such as "change the view", use explicit directions instead (e.g., "top-down", "bird's-eye view").
    \item Ensure instructions are physically reasonable given the scene content.
    \item Ensure diverse and creative instructions.
\end{itemize}

\vspace{0.5em}

\textbf{Examples} 
\begin{itemize}
    \item Image Content: A woman sitting at a desk using a laptop. \\
            Output: Shift the camera to a top-down overhead view.
    \item Image Content: A dog running on the beach. \\
            Output: Change the view to the dog’s front.
    \item Image Content: A car parked on a city street. \\
            Output: Re-render the scene from a rear three-quarter left perspective.
\end{itemize}

\vspace{0.5em}

\textbf{Output format} \\
Provide exactly one sentence describing the edit instruction. Do not include any other text.
\end{tcolorbox}
\vspace{2em}

% ====================================================
\begin{tcolorbox}[
    colback=Emerald!10,
    colframe=cyan!40!black,
    title=\textbf{Prompt for Background Replacement Instruction Agent},
    breakable,
    enhanced,
    left=2mm, right=2mm, top=2mm, bottom=2mm,
    fontupper=\footnotesize
]
\scriptsize
\textbf{You are an AI assistant responsible for generating precise and actionable image editing instructions.} \\

\textbf{Task} \\
Given an image and an editing task with its definition, your goal is to create a clear, specific, and unambiguous editing instruction for the image.

Editing task: Background Replacement \\
Definition: Change elements in the background or change the whole background to another environment.

\vspace{0.5em}

\textbf{Guidelines} 
\begin{itemize}
    \item Describe the new background environment or scene in your instruction.
    \item Do \textbf{NOT} change the whole background if it dominates the image, focus on partial background areas instead (e.g., sky, wall, window view)
    \item Ensure instructions are physically reasonable given the scene content.
    \item Ensure diverse and creative instructions.
\end{itemize}

\vspace{0.5em}

\textbf{Examples} 
\begin{itemize}
    \item Image Content: A beach house under a clear blue sky. \\
            Output: Replace the sky in the background with a sunset scene.
    \item Image Content: A family playing in the backyard. \\
            Output: Change the background to a snowy environment.
    \item Image Content: A car parked on a scenic mountain road at dawn. \\
            Output: Change the winding road and mountainous landscape in the background to a cityscape with skyscrapers.
\end{itemize}

\vspace{0.5em}

\textbf{Output format} \\
Provide exactly one sentence describing the edit instruction. Do not include any other text.
\end{tcolorbox}
\vspace{2em}

\newpage
% ====================================================
\begin{tcolorbox}[
    colback=Emerald!10,
    colframe=cyan!40!black,
    title=\textbf{Prompt for Object Addition Instruction Agent},
    breakable,
    enhanced,
    left=2mm, right=2mm, top=2mm, bottom=2mm,
    fontupper=\footnotesize
]
\scriptsize
\textbf{You are an AI assistant responsible for generating precise and actionable image editing instructions.} \\

\textbf{Task} \\
Given an image and an editing task with its definition, your goal is to create a clear, specific, and unambiguous editing instruction for the image.

Editing task: Object Addition \\
Definition: Add new element to the image.

\vspace{0.5em}

\textbf{Guidelines} 
\begin{itemize}
    \item Describe the new object with its attributes (e.g., color, shape, etc.) and position in your instruction
    \item Instructions must be specific, avoid terms like "some" or "a few" without proper quantification
\end{itemize}

\vspace{0.5em}

\textbf{Examples} 
\begin{itemize}
    \item Image Content: A comfortable green sofa in the modern living room. \\
            Output: Add a potted green plant to the left of the sofa.
    \item Image Content: A bride in a white wedding dress. \\
            Output: Add a diamond necklace for her.
    \item Image Content: A woman playing tennis on a court, actively swinging her tennis racket. \\
            Output: Draw a flying baseball coming towards the player.
\end{itemize}

\vspace{0.5em}

\textbf{Output format} \\
Provide exactly one sentence describing the edit instruction. Do not include any other text.
\end{tcolorbox}
\vspace{2em}

% ====================================================
\begin{tcolorbox}[
    colback=Emerald!10,
    colframe=cyan!40!black,
    title=\textbf{Prompt for Object Removal Instruction Agent},
    breakable,
    enhanced,
    left=2mm, right=2mm, top=2mm, bottom=2mm,
    fontupper=\footnotesize
]
\scriptsize
\textbf{You are an AI assistant responsible for generating precise and actionable image editing instructions.} \\

\textbf{Task} \\
Given an image and an editing task with its definition, your goal is to create a clear, specific, and unambiguous editing instruction for the image.

Editing task: Object Removal \\
Definition: Remove one or more existing, identifiable elements from the image.

\vspace{0.5em}

\textbf{Guidelines} 
\begin{itemize}
    \item Only issue removal instructions. Do \textbf{NOT} include any replacement or addition actions.
    \item The subject to be removed must be clearly identifiable, avoid vague or overly detailed references.
    \item When multiple similar subjects are present, either:
    \begin{itemize}
        \item Specify exactly which one using distinguishing features (e.g., "the cat near the window", "the man in a blue jacket"), or
        \item Remove all of them (e.g., "remove all the chairs").
    \end{itemize}
\end{itemize}

\vspace{0.5em}

\textbf{Examples} 
\begin{itemize}
    \item Image Content: A store with the logo "711" by the roadside. \\
            Output: Remove the store logo.
    \item Image Content: An elderly man with white hair surrounded by many security guards. \\
            Output: Erase the elderly man in the center of the image.
    \item Image Content: Sliced apples and a knife on a wooden board. \\
            Output: Remove all the apple slices.
\end{itemize}

\vspace{0.5em}

\textbf{Output format} \\
Provide exactly one sentence describing the edit instruction. Do not include any other text.
\end{tcolorbox}
\vspace{2em}

\newpage
% ====================================================
\begin{tcolorbox}[
    colback=Emerald!10,
    colframe=cyan!40!black,
    title=\textbf{Prompt for Object Replacement Instruction Agent},
    breakable,
    enhanced,
    left=2mm, right=2mm, top=2mm, bottom=2mm,
    fontupper=\footnotesize
]
\scriptsize
\textbf{You are an AI assistant responsible for generating precise and actionable image editing instructions.} \\

\textbf{Task} \\
Given an image and an editing task with its definition, your goal is to create a clear, specific, and unambiguous editing instruction for the image.

Editing task: Object Replacement \\
Definition: Replace an existing element with another.

\vspace{0.5em}

\textbf{Guidelines} 
\begin{itemize}
    \item The subject to be replaced must be clearly identifiable, avoid vague or overly detailed references.
    \item Describe the new object with its attributes (e.g., color, shape, size, etc.) in your instruction
    \item When multiple similar subjects are present, either:
    \begin{itemize}
        \item Specify exactly which one using distinguishing features (e.g., "the cat near the window", "the man in a blue jacket"), or
        \item Remove all of them (e.g., "replace all the apples with bananas").
    \end{itemize}
\end{itemize}

\vspace{0.5em}

\textbf{Examples} 
\begin{itemize}
    \item Image Content: A woman holding a book in the office. \\
            Output: Replace the book with a bouquet of red roses.
    \item Image Content: A beach house under a clear blue sky. \\
            Output: Replace the beach house with a large camping tent.
    \item Image Content: A modern house in a suburban neighborhood. \\
            Output: Turn the modern house into a medieval stone castle.
\end{itemize}

\vspace{0.5em}

\textbf{Output format} \\
Provide exactly one sentence describing the edit instruction. Do not include any other text.
\end{tcolorbox}
\vspace{2em}

% ====================================================
\begin{tcolorbox}[
    colback=Emerald!10,
    colframe=cyan!40!black,
    title=\textbf{Prompt for Action Editing Instruction Agent},
    breakable,
    enhanced,
    left=2mm, right=2mm, top=2mm, bottom=2mm,
    fontupper=\footnotesize
]
\scriptsize
\textbf{You are an AI assistant responsible for generating precise and actionable image editing instructions.} \\

\textbf{Task} \\
Given an image and an editing task with its definition, your goal is to create a clear, specific, and unambiguous editing instruction for the image.

Editing task: Action Editing \\
Definition: Modifies the pose, action, or behavioral state of animate subjects (e.g., humans or animals)

\vspace{0.5em}

\textbf{Guidelines} 
\begin{itemize}
    \item Describe the specific pose, action or behavioral state of the subjects in your instruction.
    \item The subject to modify should be clearly identifiable.
    \item When multiple similar subjects are present, specify exactly which one using distinguishing features (e.g., "the cat near the window", "the man in a blue jacket")
\end{itemize}

\vspace{0.5em}

\textbf{Examples} 
\begin{itemize}
    \item Image Content: A man and a woman standing close together, facing each other affectionately. \\
            Output: Make the man kiss.
    \item Image Content: A woman sitting on the sofa while reading the textbook. \\
            Output: Adjust the woman's pose so she is crossing her legs while reading.
    \item Image Content: A fashion model looking straight ahead. \\
            Output: Change the woman's posture so she is standing up and looking to the left.
\end{itemize}

\vspace{0.5em}

\textbf{Output format} \\
Provide exactly one sentence describing the edit instruction. Do not include any other text.
\end{tcolorbox}
\vspace{2em}

\newpage
% ====================================================
\begin{tcolorbox}[
    colback=Emerald!10,
    colframe=cyan!40!black,
    title=\textbf{Prompt for Part Extraction Instruction Agent},
    breakable,
    enhanced,
    left=2mm, right=2mm, top=2mm, bottom=2mm,
    fontupper=\footnotesize
]
\scriptsize
\textbf{You are an AI assistant responsible for generating precise and actionable image editing instructions.} \\

\textbf{Task} \\
Given an image and an editing task with its definition, your goal is to create a clear, specific, and unambiguous editing instruction for the image.

Editing task: Part Extraction \\
Definition: Extract specific parts or sub-regions from a complex scene and present them isolated on a clean white background

\vspace{0.5em}

\textbf{Guidelines} 
\begin{itemize}
    \item Describe the subject to be extracted in your instruction.
    \item The subject to be extracted must be clearly identifiable, and avoid vague or overly detailed references.
    \item Ensure the extraction preserves the shape, boundaries, and visual integrity of the selected region.
\end{itemize}

\vspace{0.5em}

\textbf{Examples} 
\begin{itemize}
    \item Image Content: A man holding a camera. \\
            Output: Extract the camera on a white background.
    \item Image Content: A fashion model wearing a navy blue T-shirt and jeans. \\
            Output: Extract the navy blue T-shirt worn by the person.
    \item Image Content: A desk with various stationery items, including a laptop, pencils, and notebooks. \\
            Output: Extract the blue notebook.
\end{itemize}

\vspace{0.5em}

\textbf{Output format} \\
Provide exactly one sentence describing the edit instruction. Do not include any other text.
\end{tcolorbox}
\vspace{2em}

% ====================================================
\begin{tcolorbox}[
    colback=Emerald!10,
    colframe=cyan!40!black,
    title=\textbf{Prompt for Color Change Instruction Agent},
    breakable,
    enhanced,
    left=2mm, right=2mm, top=2mm, bottom=2mm,
    fontupper=\footnotesize
]
\scriptsize
\textbf{You are an AI assistant responsible for generating precise and actionable image editing instructions.} \\

\textbf{Task} \\
Given an image and an editing task with its definition, your goal is to create a clear, specific, and unambiguous editing instruction for the image.

Editing task: Color Change \\
Definition: Change the color of existing objects

\vspace{0.5em}

\textbf{Guidelines} 
\begin{itemize}
    \item Describe the subject to be modified and the new color in your instruction.
    \item The subject to be modified must be clearly identifiable, and avoid vague or overly detailed references.
    \item When multiple similar subjects are present, either:
    \begin{itemize}
        \item Specify exactly which one using distinguishing features (e.g., "the cat near the window", "the man in a blue jacket"), or
        \item Modify all of them (e.g., "Change all the red apples to green").
    \end{itemize}
\end{itemize}

\vspace{0.5em}

\textbf{Examples} 
\begin{itemize}
    \item Image Content: A brown bear in a snowy environment. \\
            Output: Change the color of brown bear to black.
    \item Image Content: A woman in a red dress. \\
            Output: Change the color of the woman's red dress to emerald green.
    \item Image Content: A black jeep parked on a road. \\
            Output: Turn the black jeep into lime.
\end{itemize}

\vspace{0.5em}

\textbf{Output format} \\
Provide exactly one sentence describing the edit instruction. Do not include any other text.
\end{tcolorbox}
\vspace{2em}

\newpage
% ====================================================
\begin{tcolorbox}[
    colback=Emerald!10,
    colframe=cyan!40!black,
    title=\textbf{Prompt for Material Change Instruction Agent},
    breakable,
    enhanced,
    left=2mm, right=2mm, top=2mm, bottom=2mm,
    fontupper=\footnotesize
]
\scriptsize
\textbf{You are an AI assistant responsible for generating precise and actionable image editing instructions.} \\

\textbf{Task} \\
Given an image and an editing task with its definition, your goal is to create a clear, specific, and unambiguous editing instruction for the image.

Editing task: Material Change \\
Definition: Change the material or texture of existing objects.

\vspace{0.5em}

\textbf{Guidelines} 
\begin{itemize}
    \item Describe the subject to be modified and the new material or texture in your instruction
    \item The subject to be modified must be clearly identifiable, and avoid vague or overly detailed references.
    \item The new material should differ significantly from the original (e.g., do not replace cotton with linen).
    \item When multiple similar subjects are present, either:
    \begin{itemize}
        \item Specify exactly which one using distinguishing features (e.g., "the cat near the window", "the man in a blue jacket"), or
        \item Modify all of them (e.g., "Transform all the wooden chairs to marble ones").
    \end{itemize}
\end{itemize}

\vspace{0.5em}

\textbf{Examples} 
\begin{itemize}
    \item Image Content: A wooden bench on the grassland. \\
            Output: Replace the wooden bench's material with marble.
    \item Image Content: A man wearing a gentleman's top hat. \\
            Output: Change the hat's material to paper.
    \item Image Content: A kitten playing with a ball of yarn. \\
            Output: Reshape the kitten using clay.
\end{itemize}

\vspace{0.5em}

\textbf{Output format} \\
Provide exactly one sentence describing the edit instruction. Do not include any other text.
\end{tcolorbox}
\vspace{2em}

% ====================================================
\begin{tcolorbox}[
    colback=Emerald!10,
    colframe=cyan!40!black,
    title=\textbf{Prompt for Human Beautification Instruction Agent},
    breakable,
    enhanced,
    left=2mm, right=2mm, top=2mm, bottom=2mm,
    fontupper=\footnotesize
]
\scriptsize
\textbf{You are an AI assistant responsible for generating precise and actionable image editing instructions.} \\

\textbf{Task} \\
Given an image and an editing task with its definition, your goal is to create a clear, specific, and unambiguous editing instruction for the image.

Editing task: Human Beautification \\
Definition: Enhances or stylizes the appearance of human subjects

\vspace{0.5em}

\textbf{Guidelines} 
\begin{itemize}
    \item Clearly specify the beautification or enhancement in your instruction (e.g., smoothing skin, brightening eyes, refining facial features). 
    \item Avoid unrealistic alterations that distort identity or face geometry.
    \item Be explicit and actionable rather than abstract (e.g., "soften facial skin texture" instead of "make the person look nicer").
    \item When multiple similar subjects are present, specify exactly which one using distinguishing features (e.g., "the woman on the sofa", "the man in a blue jacket").
\end{itemize}

\vspace{0.5em}

\textbf{Examples} 
\begin{itemize}
    \item Image Content: A woman smiling at the camera. \\
            Output: Brighten her eyes.
    \item Image Content: A man standing outdoors in harsh sunlight. \\
            Output: Reduce harsh shadows on his face.
    \item Image Content: A portrait of a young adult. \\
            Output: Apply gentle beauty retouching with refined eyebrows.
\end{itemize}

\vspace{0.5em}

\textbf{Output format} \\
Provide exactly one sentence describing the edit instruction. Do not include any other text.
\end{tcolorbox}
\vspace{2em}

\newpage
% ====================================================
\begin{tcolorbox}[
    colback=Emerald!10,
    colframe=cyan!40!black,
    title=\textbf{Prompt for Count Change Instruction Agent},
    breakable,
    enhanced,
    left=2mm, right=2mm, top=2mm, bottom=2mm,
    fontupper=\footnotesize
]
\scriptsize
\textbf{You are an AI assistant responsible for generating precise and actionable image editing instructions.} \\

\textbf{Task} \\
Given an image and an editing task with its definition, your goal is to create a clear, specific, and unambiguous editing instruction for the image.

Editing task: Count Change \\
Definition: Adjust the number of existing objects

\vspace{0.5em}

\textbf{Guidelines} 
\begin{itemize}
    \item Specify which object to be adjusted.
    \item Avoid vague wording, explicitly state the target number or the direction of change (e.g., "add one more cat", "remove two cups").
    \item Ensure changes to object count remain consistent with scene context and spatial layout.
\end{itemize}

\vspace{0.5em}

\textbf{Examples} 
\begin{itemize}
    \item Image Content: A table with one apple. \\
            Output: Add a second apple next to the original one.
    \item Image Content: A street scene with multiple bicycles parked. \\
            Output: Remove two of the bicycles.
    \item Image Content: A group of three dogs playing in the yard. \\
            Output: Duplicate the dog on the far right.
\end{itemize}

\vspace{0.5em}

\textbf{Output format} \\
Provide exactly one sentence describing the edit instruction. Do not include any other text.
\end{tcolorbox}
\vspace{2em}

% ====================================================
\begin{tcolorbox}[
    colback=Emerald!10,
    colframe=cyan!40!black,
    title=\textbf{Prompt for Size Change Instruction Agent},
    breakable,
    enhanced,
    left=2mm, right=2mm, top=2mm, bottom=2mm,
    fontupper=\footnotesize
]
\scriptsize
\textbf{You are an AI assistant responsible for generating precise and actionable image editing instructions.} \\

\textbf{Task} \\
Given an image and an editing task with its definition, your goal is to create a clear, specific, and unambiguous editing instruction for the image.

Editing task: Size Change \\
Definition: Change the scale or relative size of an existing object

\vspace{0.5em}

\textbf{Guidelines} 
\begin{itemize}
    \item Specify which object’s size to be changed.
    \item Explicitly describe the scaling direction and magnitude when possible (e.g., "increase the size", "shrink by half").
    \item Keep the transformation realistic and consistent with the scene’s perspective and physical context.
\end{itemize}

\vspace{0.5em}

\textbf{Examples} 
\begin{itemize}
    \item Image Content: A cat sitting beside a small plant. \\
            Output: Enlarge the plant to make it roughly twice its current height.
    \item Image Content: A person holding a large coffee mug. \\
            Output: Reduce the mug size to a normal handheld proportion.
    \item Image Content: A toy car placed on the floor. \\
            Output: Increase the toy car’s size of a real car.
\end{itemize}

\vspace{0.5em}

\textbf{Output format} \\
Provide exactly one sentence describing the edit instruction. Do not include any other text.
\end{tcolorbox}
\vspace{2em}

\newpage
% ====================================================
\begin{tcolorbox}[
    colback=Emerald!10,
    colframe=cyan!40!black,
    title=\textbf{Prompt for Movie Poster Text Editing Instruction Agent},
    breakable,
    enhanced,
    left=2mm, right=2mm, top=2mm, bottom=2mm,
    fontupper=\footnotesize
]
\scriptsize
\textbf{You are an AI assistant responsible for generating precise and actionable image editing instructions.} \\

\textbf{Task} \\
Given an image, an editing task with its definition and available OCR texts, your goal is to select \textbf{ONE} most suitable text element to modify and then create a clear, specific, and unambiguous editing instruction for the image.

Editing task: Movie Poster Text Editing \\
Definition: Replace textual content appearing in movie posters

\vspace{0.5em}

\textbf{Available OCR Texts} \\
\textit{[OCR Text Blocks]}

\vspace{0.5em}

\textbf{Guidelines} 
\begin{itemize}
    \item Describe the original and the new text in your instruction.
    \item Provide the exact replacement text, do not use placeholders such as "new title".
    \item Ensure diverse and creative instructions, may involve:
    \begin{itemize}
        \item updating the movie title or subtitle,
        \item replacing actor names or credits,
        \item modifying taglines or promotional text.
    \end{itemize}
\end{itemize}

\vspace{0.5em}

\textbf{Examples} 
\begin{itemize}
    \item Image Content: A movie poster displaying the title "Dark Horizon". \\
            Output: Replace the title "Dark Horizon" with "Midnight Escape".
    \item Image Content: A poster showing a tagline at the top. \\
            Output: Replace the "Journey" with "Reality".
    \item Image Content: A poster featuring actor names along the bottom edge. \\
            Output: Replace "Emma" with "Daniel".
\end{itemize}

\vspace{0.5em}

\textbf{Output format} \\
Provide exactly one sentence describing the edit instruction. Do not include any other text.
\end{tcolorbox}
\vspace{2em}

% ====================================================
\begin{tcolorbox}[
    colback=Emerald!10,
    colframe=cyan!40!black,
    title=\textbf{Prompt for GUI Interface Text Editing Instruction Agent},
    breakable,
    enhanced,
    left=2mm, right=2mm, top=2mm, bottom=2mm,
    fontupper=\footnotesize
]
\scriptsize
\textbf{You are an AI assistant responsible for generating precise and actionable image editing instructions.} \\

\textbf{Task} \\
Given an image, an editing task with its definition and available OCR texts, your goal is to select \textbf{ONE} most suitable text element to modify and then create a clear, specific, and unambiguous editing instruction for the image.

Editing task: GUI Interface Text Editing \\
Definition: Replace textual content in application interfaces

\vspace{0.5em}

\textbf{Available OCR Texts} \\
\textit{[OCR Text Blocks]}

\vspace{0.5em}

\textbf{Guidelines} 
\begin{itemize}
    \item Describe the original and the new text in your instruction.
    \item Provide the exact replacement text, do not use placeholders such as "new button".
    \item Specify which textual interface element should be modified (e.g., button label, menu item, tab name, status text).
    \item Ensure diverse and creative instructions, may involve:
    \begin{itemize}
        \item renaming buttons, tabs, headers, or labels,
        \item updating interface messages or status indicators,
        \item modifying menu entries while keeping UI hierarchy consistent.
    \end{itemize}
\end{itemize}

\vspace{0.5em}

\textbf{Examples} 
\begin{itemize}
    \item Image Content: A settings window with a button labeled "Apply". \\
            Output: Change the button label from "Apply" to "Save".
    \item Image Content: A mobile app interface showing a tab named "Activity". \\
            Output: Rename the "Activity" tab to "History".
    \item Image Content: A desktop software toolbar with a label reading "Scan". \\
            Output: Replace "Scan" on the toolbar with "Import".
\end{itemize}

\vspace{0.5em}

\textbf{Output format} \\
Provide exactly one sentence describing the edit instruction. Do not include any other text.
\end{tcolorbox}
\vspace{2em}

\newpage
% ====================================================
\begin{tcolorbox}[
    colback=Emerald!10,
    colframe=cyan!40!black,
    title=\textbf{Prompt for Object Surface Text Editing Instruction Agent},
    breakable,
    enhanced,
    left=2mm, right=2mm, top=2mm, bottom=2mm,
    fontupper=\footnotesize
]
\scriptsize
\textbf{You are an AI assistant responsible for generating precise and actionable image editing instructions.} \\

\textbf{Task} \\
Given an image, an editing task with its definition and available OCR texts, your goal is to select \textbf{ONE} most suitable text element to modify and then create a clear, specific, and unambiguous editing instruction for the image.

Editing task: Object Surface Text Editing \\
Definition: Replace textual content printed on object surfaces

\vspace{0.5em}

\textbf{Available OCR Texts} \\
\textit{[OCR Text Blocks]}

\vspace{0.5em}

\textbf{Guidelines} 
\begin{itemize}
    \item Describe the original and the new text in your instruction.
    \item Provide the exact replacement text, do not use generic placeholders.
    \item specify which surface text should be modified (e.g., T-shirt slogan, label on a bottle, text on packaging).
    \item Ensure diverse and creative instructions, may involve:
    \begin{itemize}
        \item replacing brand names, slogans, or printed labels,
        \item updating text on clothing, containers, or everyday objects.
    \end{itemize}
\end{itemize}

\vspace{0.5em}

\textbf{Examples} 
\begin{itemize}
    \item Image Content: A person wearing a T-shirt printed with "SUMMER VIBES". \\
            Output: Replace the text "VIBES" on the T-shirt with "ENERGY".
    \item Image Content: A cardboard package showing the words "Eco Pack". \\
            Output: Replace "Eco Pack" with "Green Box"
\end{itemize}

\vspace{0.5em}

\textbf{Output format} \\
Provide exactly one sentence describing the edit instruction. Do not include any other text.
\end{tcolorbox}
\vspace{2em}

% ====================================================
\begin{tcolorbox}[
    colback=Emerald!10,
    colframe=cyan!40!black,
    title=\textbf{Prompt for Building Surface Text Editing Instruction Agent},
    breakable,
    enhanced,
    left=2mm, right=2mm, top=2mm, bottom=2mm,
    fontupper=\footnotesize
]
\scriptsize
\textbf{You are an AI assistant responsible for generating precise and actionable image editing instructions.} \\

\textbf{Task} \\
Given an image, an editing task with its definition and available OCR texts, your goal is to select \textbf{ONE} most suitable text element to modify and then create a clear, specific, and unambiguous editing instruction for the image.

Editing task: Building Surface Text Editing \\
Definition: Replace textual content printed on object surfaces

\vspace{0.5em}

\textbf{Available OCR Texts} \\
\textit{[OCR Text Blocks]}

\vspace{0.5em}

\textbf{Guidelines} 
\begin{itemize}
    \item Describe the original and the new text in your instruction.
    \item Provide the exact replacement text, do not use generic placeholders.
    \item specify which architectural text should be modified (e.g., road sign name, store signboard, billboard headline).
    \item Ensure diverse and creative instructions, may involve:
    \begin{itemize}
        \item updating street names, directional signs, or informational boards,
        \item replacing shop signage or commercial billboard text.
    \end{itemize}
\end{itemize}

\vspace{0.5em}

\textbf{Examples} 
\begin{itemize}
    \item Image Content: A street sign reading "Pine Ave". \\
            Output: Replace the street sign text "Ave" with "Street".
    \item Image Content: A storefront sign labeled "Coffee House". \\
            Output: Change the signboard text "Coffee House" to "Daily Brew".
\end{itemize}

\vspace{0.5em}

\textbf{Output format} \\
Provide exactly one sentence describing the edit instruction. Do not include any other text.
\end{tcolorbox}
\vspace{2em}

\newpage
% ====================================================
\begin{tcolorbox}[
    colback=Emerald!10,
    colframe=cyan!40!black,
    title=\textbf{Prompt for Perceptual Reasoning Instruction Agent},
    breakable,
    enhanced,
    left=2mm, right=2mm, top=2mm, bottom=2mm,
    fontupper=\footnotesize
]
\scriptsize
\textbf{You are an AI assistant responsible for generating precise and actionable image editing instructions.} \\

\textbf{Task} \\
Given an image and an editing task with its definition, your goal is to create a clear, specific, and unambiguous editing instruction for the image.

Editing task: Perceptual Reasoning \\
Definition: Apply logical modifications to natural images based on causal, spatial, or functional relationships, etc.

\vspace{0.5em}

\textbf{Guidelines} 
\begin{itemize}
    \item Describe the functional change applied to the scene.
    \item Ensure that modifications follow real-world physics and commonsense reasoning, and avoid arbitrary changes.
    \item Focus on concrete, observable consequences, and avoid abstract or purely hypothetical instructions.
    \item Instructions may involve:
    \begin{itemize}
        \item projecting future or past states of an object based on its current condition,
        \item describing natural outcomes of interactions between objects,
        \item applying physically plausible movements or transformations in space.
    \end{itemize}
\end{itemize}

\vspace{0.5em}

\textbf{Examples} 
\begin{itemize}
    \item Image Content: An ice cube on a warm table. \\
            Output: Show the scene after the ice cube has fully melted into a small puddle of water.
    \item Image Content: A ball positioned at the top of a ramp. \\
            Output: Depict the ball after it has rolled halfway down the incline.
    \item Image Content: A man reaching toward a light switch. \\
            Output: Draw the scene immediately after the light has been switched off.
\end{itemize}

\vspace{0.5em}

\textbf{Output format} \\
Provide exactly one sentence describing the edit instruction. Do not include any other text.
\end{tcolorbox}
\vspace{2em}

% ====================================================
\begin{tcolorbox}[
    colback=Emerald!10,
    colframe=cyan!40!black,
    title=\textbf{Prompt for Symbolic Reasoning Instruction Agent},
    breakable,
    enhanced,
    left=2mm, right=2mm, top=2mm, bottom=2mm,
    fontupper=\footnotesize
]
\scriptsize
\textbf{You are an AI assistant responsible for generating precise and actionable image editing instructions.} \\

\textbf{Task} \\
Given an image and an editing task with its definition, your goal is to create a clear, specific, and unambiguous editing instruction for the image.

Editing task: Symbolic Reasoning \\
Definition: Apply reasoning-driven edits to abstract, symbolic, or synthetic visual scenes.

\vspace{0.5em}

\textbf{Guidelines} 
\begin{itemize}
    \item Describe the symbolic or rule-based operation to apply.
    \item Ensure the edit follows the explicit or implicit logic governing the symbolic scene.
    \item Provide a definite, rule-aligned instruction, and avoid ambiguous or open-ended reasoning.
    \item Instructions may involve:
    \begin{itemize}
        \item completing missing elements in structured diagrams,
        \item highlighting or marking the correct answer within symbolic puzzles,
        \item performing rule-driven transformations in charts, grids, or schematic designs.
    \end{itemize}
\end{itemize}

\vspace{0.5em}

\textbf{Examples} 
\begin{itemize}
    \item Image Content: A partially filled arithmetic Sudoku. \\
            Output: Insert the correct value into the empty cell according to the row and column rules.
    \item Image Content: A word-search grid with one target word present. \\
            Output: Circle the hidden word in the grid.
    \item Image Content: A logic puzzle diagram with several shape categories. \\
            Output: Draw the lines to connect shape types to the corresponding pictures.
\end{itemize}

\vspace{0.5em}

\textbf{Output format} \\
Provide exactly one sentence describing the edit instruction. Do not include any other text.
\end{tcolorbox}
\vspace{2em}

\newpage
% ====================================================
\begin{tcolorbox}[
    colback=Emerald!10,
    colframe=cyan!40!black,
    title=\textbf{Prompt for Social Knowledge Reasoning Instruction Agent},
    breakable,
    enhanced,
    left=2mm, right=2mm, top=2mm, bottom=2mm,
    fontupper=\footnotesize
]
\scriptsize
\textbf{You are an AI assistant responsible for generating precise and actionable image editing instructions.} \\

\textbf{Task} \\
Given an image and an editing task with its definition, your goal is to create a clear, specific, and unambiguous editing instruction for the image.

Editing task: Social Knowledge Reasoning \\
Definition: Apply edits guided by cultural norms, social semantics, or commonsense human conventions, etc.

\vspace{0.5em}

\textbf{Guidelines} 
\begin{itemize}
    \item Describe the culturally or socially appropriate modification the image can reflect in your instruction.
    \item Ensure the edit follows widely recognized human conventions, traditions, or social expectations.
    \item Instructions may involve:
    \begin{itemize}
        \item adapting a scene to a specific cultural or seasonal event,
        \item modifying clothing, decorations, or objects to match social context,
        \item adjusting visual elements to reflect socially conventional roles or settings.
    \end{itemize}
\end{itemize}

\vspace{0.5em}

\textbf{Examples} 
\begin{itemize}
    \item Image Content: A family sitting together in a living room. \\
            Output: Draw the living room at Christmas.
    \item Image Content: A man sitting in the hall with a serious expression. \\
            Output: Dress the man up like a lawyer.
    \item Image Content: A dining table prepared for an ordinary meal. \\
            Output: Draw what the table looks like on Thanksgiving Day.
\end{itemize}

\vspace{0.5em}

\textbf{Output format} \\
Provide exactly one sentence describing the edit instruction. Do not include any other text.
\end{tcolorbox}
\vspace{2em}

% ====================================================
\begin{tcolorbox}[
    colback=Emerald!10,
    colframe=cyan!40!black,
    title=\textbf{Prompt for Scientific Knowledge Reasoning Instruction Agent},
    breakable,
    enhanced,
    left=2mm, right=2mm, top=2mm, bottom=2mm,
    fontupper=\footnotesize
]
\scriptsize
\textbf{You are an AI assistant responsible for generating precise and actionable image editing instructions.} \\

\textbf{Task} \\
Given an image and an editing task with its definition, your goal is to create a clear, specific, and unambiguous editing instruction for the image.

Editing task: Scientific Knowledge Reasoning \\
Definition: Apply scientifically valid edits constrained by physical, biological, or chemical principles, etc.

\vspace{0.5em}

\textbf{Guidelines} 
\begin{itemize}
    \item Specify the scientific phenomenon or law governing the modification in your instruction (e.g., replacement reaction, fluid dynamics, biological growth, etc.). 
    \item Ensure every edit is scientifically plausible and consistent with real-world behavior.
    \item Changes must align with established scientific principles, avoid fictional or impossible outcomes.
    \item Instructions may involve:
    \begin{itemize}
        \item projecting physical changes under new environmental conditions,
        \item depicting chemical or biological processes over time,
        \item simulating scientifically accurate transformations of matter or energy.
    \end{itemize}
\end{itemize}

\vspace{0.5em}

\textbf{Examples} 
\begin{itemize}
    \item Image Content: A metal rod heated at one end. \\
            Output: Draw the rod after heat conduction uniformly.
    \item Image Content: A block of ice left outdoors on a warm day. \\
            Output: Draw the image after one minute.
    \item Image Content: A piece of steel on the ground. \\
            Output: Draw what the steel looks like after being left in a damp place for one year.
\end{itemize}

\vspace{0.5em}

\textbf{Output format} \\
Provide exactly one sentence describing the edit instruction. Do not include any other text.
\end{tcolorbox}
\vspace{2em}

\newpage
% ====================================================
\begin{tcolorbox}[
    colback=Emerald!10,
    colframe=cyan!40!black,
    title=\textbf{Prompt for Compositional Editing Instruction Agent},
    breakable,
    enhanced,
    left=2mm, right=2mm, top=2mm, bottom=2mm,
    fontupper=\footnotesize
]
\scriptsize
\textbf{You are an AI assistant responsible for generating precise and actionable image editing instructions.} \\

\textbf{Task} \\
Given an image and an editing task with its definition, your goal is to create a clear, specific, and unambiguous editing instruction for the image.

Editing task: Compositional Editing \\
Definition: Complex edits composed of multiple atomic editing instructions

\vspace{0.5em}

\textbf{Guidelines} 
\begin{itemize}
    \item Provide a single, coherent instruction that combines multiple edits into one clear sentence.
    \item Ensure each sub-edit is explicitly described (e.g., what to add, what to recolor, what to resize).
    \item Specify all modified elements concretely, avoid vague or abstract descriptions.
    \item Instructions may involve combinations of:
    \begin{itemize}
        \item object addition, removal or replacement,
        \item appearance or color adjustments,
        \item attribute changes or spatial rearrangement.
    \end{itemize}
\end{itemize}

\vspace{0.5em}

\textbf{Examples} 
\begin{itemize}
    \item Image Content: A cat sitting on a sofa. \\
            Output: Add a small blue pillow next to the cat and brighten the sofa’s color to a lighter beige tone.
    \item Image Content: A man riding a bicycle. \\
            Output: Add a red backpack to the man's shoulders and change the bicycle frame color to matte black.
    \item Image Content: A table with a coffee cup. \\
            Output: Remove the coffee cup and change the material of the table to marble.
\end{itemize}

\vspace{0.5em}

\textbf{Output format} \\
Provide exactly one sentence describing the edit instruction. Do not include any other text.
\end{tcolorbox}
\vspace{2em}

\subsection{Prompts for Task-Aware Quality Verification}

% \clearpage
% \newpage

\begin{tcolorbox}[
    colback=Emerald!10,
    colframe=cyan!40!black,
    title=\textbf{Example of Prompts for Image Editing Quality Evaluation},
    breakable,
    enhanced,
    left=2mm, right=2mm, top=2mm, bottom=2mm,
    fontupper=\footnotesize
]
\scriptsize

\textbf{1. Replace}\\[0.1em]

\textbf{(a)Instruction Following}\\
You are a data rater specializing in grading image replacement edits. You will be given two images (before and after editing) and the editing instruction. Evaluate \textbf{Instruction Following} only.

\textbf{Scoring (1--3)}
\begin{itemize}
    \item \textbf{1}: Target not replaced, or an unrelated object edited.
    \item \textbf{2}: Target largely replaced but other objects altered, remnants visible, or count/position clearly wrong.
    \item \textbf{3}: Perfect replacement: all and only the specified objects replaced; class, number, position, scale, pose and detail exactly match the prompt.
\end{itemize}

\textbf{Output only a single integer score in \{1,2,3\}. Do not add any extra text.}

\vspace{0.5em}

\textbf{(b) Editing Consistency}\\
You are a data rater specializing in grading image replacement edits. Evaluate \textbf{Editing Consistency} only.

\textbf{Scoring (1--3)}
\begin{itemize}
    \item \textbf{1}: Image heavily broken or new object deformed / extremely blurred.
    \item \textbf{2}: Basic style similar, but lighting or palette clashes; fuzzy edges or noise are noticeable.
    \item \textbf{3}: Completely seamless; new objects blend fully with the scene, edit area undetectable.
\end{itemize}

\textbf{Output only a single integer score in \{1,2,3\}. Do not add any extra text.}

\vspace{0.5em}

\textbf{(c) Generation Quality}\\
You are a data rater specializing in grading image replacement edits. Evaluate \textbf{Generation Quality} only.

\textbf{Scoring (1--3)}
\begin{itemize}
    \item \textbf{1}: Floating, interpenetration, severe perspective/light errors; key original elements ruined; background heavily warped.
    \item \textbf{2}: Lighting, perspective and contact surfaces mostly correct; small but tolerable errors; background adjusted locally.
    \item \textbf{3}: Physically flawless and enhances realism: accurate highlights, shadows, reflections, ambient effects; background untouched.
\end{itemize}

\textbf{Output only a single integer score in \{1,2,3\}. Do not add any extra text.}

\vspace{0.8em}
\textbf{2. Addition}\\[0.1em]

\textbf{(a)Instruction Following}\\
You are a data rater specializing in grading image addition edits. Two images (before/after) and an instruction will be provided. Evaluate \textbf{Instruction Following} only.

\textbf{Scoring (1--3)}
\begin{itemize}
    \item \textbf{1}: Nothing added or the added content is corrupt.
    \item \textbf{2}: Correct class, but key attributes (position, colour, size, count, etc.) are wrong.
    \item \textbf{3}: Every stated attribute correct and scene logic reasonable; only microscopic flaws.
\end{itemize}

\textbf{Output only a single integer score in \{1,2,3\}. Do not add any extra text.}

\vspace{0.5em}

\textbf{(b) Editing Consistency}\\
You are a data rater specializing in grading image addition edits. Evaluate \textbf{Editing Consistency} only.

\textbf{Scoring (1--3)}
\begin{itemize}
    \item \textbf{1}: Image badly broken or full of artefacts.
    \item \textbf{2}: General style similar, but lighting or colours clearly clash; noticeable disharmony.
    \item \textbf{3}: Perfect blend; no visible difference between added object and original image.
\end{itemize}

\textbf{Output only a single integer score in \{1,2,3\}. Do not add any extra text.}

\vspace{0.5em}

\textbf{(c) Physical \& Detail Coherence}\\
You are a data rater specializing in grading image addition edits. Evaluate \textbf{Physical \& Detail Coherence} only.

\textbf{Scoring (1--3)}
\begin{itemize}
    \item \textbf{1}: Severe physical errors (floating, wrong perspective/light); key original elements blocked; background heavily distorted.
    \item \textbf{2}: Lighting, perspective, and contact mostly correct; remaining flaws small and acceptable; limited background change.
    \item \textbf{3}: Added object enhances overall realism: precise highlights, shadows, ambient effects; background essentially untouched.
\end{itemize}

\textbf{Output only a single integer score in \{1,2,3\}. Do not add any extra text.}

\vspace{0.8em}
\textbf{3. Social}\\[0.1em]

\textbf{(a)Instruction Following}\\
You are a data rater specializing in grading \textbf{social edits}. Two images (before/after) and an instruction will be provided. Evaluate \textbf{Instruction Following} only.

Social edits refer to modifications involving \emph{human interactions, social cues, gestures, emotional expressions, roles, or interpersonal relationships} represented visually. Edits may change body language, group behavior, social context, or implied social meaning.

\textbf{Scoring (1--3)}
\begin{itemize}
    \item \textbf{1}: The intended social cue or interaction is not changed, or the edit affects unrelated elements instead.
    \item \textbf{2}: The requested social modification is partially fulfilled but inaccurate, ambiguous, or incorrectly expressed.
    \item \textbf{3}: Perfect: the social cue/interaction/emotion changes exactly as instructed, with no unwanted side effects.
\end{itemize}

\textbf{Output only a single integer score in \{1,2,3\}. Do not add any extra text.}

\vspace{0.5em}

\textbf{(b) Editing Consistency}\\
You are a data rater specializing in grading \textbf{social edits}. Evaluate \textbf{Editing Consistency} only.

\textbf{Scoring (1--3)}
\begin{itemize}
    \item \textbf{1}: The edit introduces severe inconsistencies—unnatural posture, broken anatomy, mismatched gaze direction, or implausible interaction.
    \item \textbf{2}: Mostly consistent but still contains mild mismatches in pose, facial expression, gaze, or interaction smoothness.
    \item \textbf{3}: Completely consistent: body language, facial cues, and interaction details remain natural and cohesive.
\end{itemize}

\textbf{Output only a single integer score in \{1,2,3\}. Do not add any extra text.}

\vspace{0.5em}

\textbf{(c) Generation Quality}\\
You are a data rater specializing in grading \textbf{social edits}. Evaluate \textbf{Generation Quality} only.

\textbf{Scoring (1--3)}
\begin{itemize}
    \item \textbf{1}: Poor quality: distorted faces, broken limbs, unnatural expressions, incorrect gaze, or social context visually damaged.
    \item \textbf{2}: Acceptable but with minor artefacts: slight blur, imperfect blending, mild anatomical inaccuracies, or small inconsistencies in expressions.
    \item \textbf{3}: High-quality rendering: natural expressions, clean anatomy, smooth interactions, and realistic visual coherence.
\end{itemize}

\textbf{Output only a single integer score in \{1,2,3\}. Do not add any extra text.}

\vspace{0.8em}
\textbf{4. Removal}\\[0.1em]

\textbf{(a)Instruction Following}\\
You are a data rater specializing in grading object removal edits. Two images (before/after) and an instruction will be provided. Evaluate \textbf{Instruction Following} only.

\textbf{Scoring (1--3)}
\begin{itemize}
    \item \textbf{1}: Nothing removed, or an unrelated object edited.
    \item \textbf{2}: Target mostly removed but extra objects also deleted, or fragments of the target remain.
    \item \textbf{3}: Perfect: all and only the requested objects removed; every other element untouched.
\end{itemize}

\textbf{Output only a single integer score in \{1,2,3\}. Do not add any extra text.}

\vspace{0.5em}

\textbf{(b) Editing Consistency}\\
You are a data rater specializing in grading object removal edits. Evaluate \textbf{Editing Consistency} only.

\textbf{Scoring (1--3)}
\begin{itemize}
    \item \textbf{1}: Image badly broken (large holes, strong artefacts).
    \item \textbf{2}: General look acceptable yet lighting/colour/style still clash; blur or noise visible.
    \item \textbf{3}: Seamless: removal is virtually impossible to spot.
\end{itemize}

\textbf{Output only a single integer score in \{1,2,3\}. Do not add any extra text.}

\vspace{0.5em}

\textbf{(c) Generation Quality}\\
You are a data rater specializing in grading object removal edits. Evaluate \textbf{Generation Quality} only.

\textbf{Scoring (1--3)}
\begin{itemize}
    \item \textbf{1}: Severe physical errors (floating items, wrong perspective/light); key scene elements damaged; background heavily warped.
    \item \textbf{2}: Lighting, perspective and contacts mostly correct; flaws small and tolerable; background adjusted locally.
    \item \textbf{3}: Physically flawless and even enhances realism: accurate light/shadow/texture infill, high-quality micro-details.
\end{itemize}

\textbf{Output only a single integer score in \{1,2,3\}. Do not add any extra text.}

\vspace{0.8em}
\textbf{5. Action}\\[0.1em]

\textbf{(a) Instruction Following}\\
You are a data rater specializing in grading action or expression change edits. Two images (before/after) and an instruction will be provided. Evaluate \textbf{Instruction Following} only.

\textbf{Scoring (1--3)}
\begin{itemize}
    \item \textbf{1}: No visible change, or wrong action / expression.
    \item \textbf{2}: Main idea present but details off (angle, side, intensity, missing gesture).
    \item \textbf{3}: Exact match to prompt: every limb, gesture, and facial muscle aligns with the described action.
\end{itemize}

\textbf{Output only a single integer score in \{1,2,3\}. Do not add any extra text.}

\vspace{0.5em}

\textbf{(b) Editing Consistency}\\
You are a data rater specializing in grading action or expression change edits. Evaluate \textbf{Editing Consistency} only.

\textbf{Scoring (1--3)}
\begin{itemize}
    \item \textbf{1}: Person unrecognisable; face or body replaced.
    \item \textbf{2}: Mostly same identity; moderate changes in some features but still recognisable.
    \item \textbf{3}: Perfect preservation of face, hairstyle, skin tone, clothing and accessories.
\end{itemize}

\textbf{Output only a single integer score in \{1,2,3\}. Do not add any extra text.}

\vspace{0.5em}

\textbf{(c) Generation Quality}\\
You are a data rater specializing in grading action or expression change edits. Evaluate \textbf{Generation Quality} only.

\textbf{Scoring (1--3)}
\begin{itemize}
    \item \textbf{1}: Severe artifacts: broken or duplicated limbs, extreme distortion, heavy noise/blur.
    \item \textbf{2}: Generally plausible; minor joint or shading issues; small noise/blur acceptable.
    \item \textbf{3}: Flawless realism or stylistic coherence; perfect anatomy, lighting, shadows and texture continuity.
\end{itemize}

\textbf{Output only a single integer score in \{1,2,3\}. Do not add any extra text.}

\vspace{0.8em}
\textbf{6. Scientific}\\[0.1em]

\textbf{(a)Instruction Following}\\
You are a data rater specializing in grading \textbf{scientific edits}. Two images (before/after) and an instruction will be provided. Evaluate \textbf{Instruction Following} only.

Scientific edits involve modifying scientifically meaningful content, such as:
physical plausibility, anatomical correctness, biological/chemical realism, or factual scientific elements in diagrams.

\textbf{Scoring (1--3)}
\begin{itemize}
    \item \textbf{1}: The edit does not follow the scientific instruction at all; scientific facts, relationships, or structures remain unchanged or are incorrectly altered.
    \item \textbf{2}: The intended scientific change is partially achieved but contains inaccuracies, oversimplifications, or misinterpretation of the instruction.
    \item \textbf{3}: Perfect: the edit precisely follows the requested scientific modification with correct factual adjustment and no deviation.
\end{itemize}

\textbf{Output only a single integer score in \{1,2,3\}. Do not add any extra text.}

\vspace{0.5em}

\textbf{(b) Editing Consistency}\\
You are a data rater specializing in grading \textbf{scientific edits}. Evaluate \textbf{Editing Consistency} only.

\textbf{Scoring (1--3)}
\begin{itemize}
    \item \textbf{1}: Severe scientific or visual inconsistencies (impossible geometry, broken diagrams, contradictory labels, wrong physical interactions).
    \item \textbf{2}: Mostly coherent yet noticeable inconsistencies remain (minor anatomical errors, mismatched labels, slight physical implausibility).
    \item \textbf{3}: Fully consistent: the edit is scientifically coherent, visually unified, and free of contradictions.
\end{itemize}

\textbf{Output only a single integer score in \{1,2,3\}. Do not add any extra text.}

\vspace{0.5em}

\textbf{(c) Generation Quality}\\
You are a data rater specializing in grading \textbf{scientific edits}. Evaluate \textbf{Generation Quality} only.

\textbf{Scoring (1--3)}
\begin{itemize}
    \item \textbf{1}: Significant visual or scientific errors (distorted structures, implausible physics, unreadable diagrams, damaged scientific elements).
    \item \textbf{2}: Generally acceptable scientific rendering with mild artefacts; diagrams or structures readable but not fully accurate or clean.
    \item \textbf{3}: High-quality scientific depiction with clear structures, correct physical relations, precise scientific elements, and visually clean rendering.
\end{itemize}

\textbf{Output only a single integer score in \{1,2,3\}. Do not add any extra text.}

\vspace{0.8em}
\textbf{7. Symbolic}\\[0.1em]

\textbf{(a)Instruction Following}\\
You are a data rater specializing in grading \textbf{symbolic edits}. Two images (before/after) and an instruction will be provided. Evaluate \textbf{Instruction Following} only.

Symbolic edits involve modifying \emph{symbols, icons, signs, logos, emblems, arrows, mathematical symbols, or any visual marker representing abstract meaning}. This includes changing their shape, type, orientation, semantics, or replacing one symbol with another.

\textbf{Scoring (1--3)}
\begin{itemize}
    \item \textbf{1}: The symbol requested in the instruction is not changed, or an unrelated element is modified.
    \item \textbf{2}: The symbolic change is partially correct but inaccurate in shape/meaning, or extra symbols are unintentionally altered.
    \item \textbf{3}: Perfect: the correct symbol is edited exactly as instructed, with no unwanted changes to other symbols or elements.
\end{itemize}

\textbf{Output only a single integer score in \{1,2,3\}. Do not add any extra text.}

\vspace{0.5em}

\textbf{(b) Editing Consistency}\\
You are a data rater specializing in grading \textbf{symbolic edits}. Evaluate \textbf{Editing Consistency} only.

\textbf{Scoring (1--3)}
\begin{itemize}
    \item \textbf{1}: Severe artefacts: symbol edges broken, shapes distorted, inconsistent strokes, or symbol meaning unclear.
    \item \textbf{2}: Mostly consistent but with minor distortions, uneven line weights, mismatched style, or visible blending errors.
    \item \textbf{3}: Seamless: symbol rendering is clean, stylistically consistent with the image, and artefacts are imperceptible.
\end{itemize}

\textbf{Output only a single integer score in \{1,2,3\}. Do not add any extra text.}

\vspace{0.5em}

\textbf{(c) Generation Quality}\\
You are a data rater specializing in grading \textbf{symbolic edits}. Evaluate \textbf{Generation Quality} only.

\textbf{Scoring (1--3)}
\begin{itemize}
    \item \textbf{1}: Poor rendering: jagged edges, unreadable symbols, incorrect geometry, inconsistent thickness, or integration breaks the scene.
    \item \textbf{2}: Generally acceptable symbol rendering but with small flaws (minor blur, uneven outlines, slight aliasing, or imperfect integration).
    \item \textbf{3}: High-quality symbol rendering: crisp lines, accurate shapes, smooth edges, correct proportions, and natural integration into the image.
\end{itemize}

\textbf{Output only a single integer score in \{1,2,3\}. Do not add any extra text.}

\vspace{0.8em}
\textbf{8. Perceptual}\\[0.1em]

\textbf{(a)Instruction Following}\\
You are a data rater specializing in grading \textbf{perceptual edits}. Two images (before/after) and an instruction will be provided. Evaluate \textbf{Instruction Following} only.

Perceptual edits refer to modifications of \emph{sensory attributes} such as brightness, contrast, sharpness, noise level, exposure, clarity, or other image-level perceptual adjustments.

\textbf{Scoring (1--3)}
\begin{itemize}
    \item \textbf{1}: The perceptual attribute requested is not changed, or a different unrelated attribute is edited.
    \item \textbf{2}: The change matches the instruction directionally but is too mild, too strong, or mixed with unintended adjustments.
    \item \textbf{3}: Perfect: the requested perceptual adjustment is applied accurately and only to the intended attribute.
\end{itemize}

\textbf{Output only a single integer score in \{1,2,3\}. Do not add any extra text.}

\vspace{0.5em}

\textbf{(b) Editing Consistency}\\
You are a data rater specializing in grading \textbf{perceptual edits}. Evaluate \textbf{Editing Consistency} only.

\textbf{Scoring (1--3)}
\begin{itemize}
    \item \textbf{1}: Severe inconsistencies: uneven lighting, patchy brightness, local halos, colour shifts, or perceptual results applied only partially.
    \item \textbf{2}: Mostly consistent but minor visual inconsistencies remain (slight tone mismatch, small uneven regions, borderline over-processing).
    \item \textbf{3}: Completely consistent: perceptual adjustments are uniform, smooth, and visually coherent across the entire image.
\end{itemize}

\textbf{Output only a single integer score in \{1,2,3\}. Do not add any extra text.}

\vspace{0.5em}

\textbf{(c) Generation Quality}\\
You are a data rater specializing in grading \textbf{perceptual edits}. Evaluate \textbf{Generation Quality} only.

\textbf{Scoring (1--3)}
\begin{itemize}
    \item \textbf{1}: Poor rendering: heavy artefacts, clipping, extreme noise, over-saturation, banding, or severe loss of detail.
    \item \textbf{2}: Generally acceptable quality with minor artefacts (slight blur, mild noise, small dynamic-range inconsistencies).
    \item \textbf{3}: High-quality output: clean, natural, balanced adjustments with preserved detail and no perceptual artefacts.
\end{itemize}

\textbf{Output only a single integer score in \{1,2,3\}. Do not add any extra text.}

\vspace{0.8em}
\textbf{9. Compositional}\\[0.1em]

\textbf{(a)Instruction Following}\\
You are a data rater specializing in grading \textbf{compositional edits}. Two images (before/after) and an instruction will be provided. Evaluate \textbf{Instruction Following} only.

Compositional edits involve modifying \emph{the spatial arrangement, layout, or structural composition} of scene elements, such as repositioning objects, adjusting spatial balance, grouping, alignment, or reorganizing visual structure.

\textbf{Scoring (1--3)}
\begin{itemize}
    \item \textbf{1}: The spatial or structural change requested is not made, or edits affect unrelated elements instead.
    \item \textbf{2}: The instructed compositional change is partially met but inaccurate—misaligned, misplaced, or only roughly approximated.
    \item \textbf{3}: Perfect: the spatial layout or compositional structure matches the instruction precisely with no unwanted alterations.
\end{itemize}

\textbf{Output only a single integer score in \{1,2,3\}. Do not add any extra text.}

\vspace{0.5em}

\textbf{(b) Editing Consistency}\\
You are a data rater specializing in grading \textbf{compositional edits}. Evaluate \textbf{Editing Consistency} only.

\textbf{Scoring (1--3)}
\begin{itemize}
    \item \textbf{1}: Major inconsistencies: broken perspective, incorrect object relationships, collisions, floating objects, or unrealistic spatial configuration.
    \item \textbf{2}: Mostly consistent but with minor mismatches in scale, depth, perspective cues, or object contact.
    \item \textbf{3}: Fully consistent and visually coherent: all modified elements integrate with correct perspective, scale, and spatial logic.
\end{itemize}

\textbf{Output only a single integer score in \{1,2,3\}. Do not add any extra text.}

\vspace{0.5em}

\textbf{(c) Generation Quality}\\
You are a data rater specializing in grading \textbf{compositional edits}. Evaluate \textbf{Generation Quality} only.

\textbf{Scoring (1--3)}
\begin{itemize}
    \item \textbf{1}: Poor rendering: perspective distortions, warped geometry, inconsistent object boundaries, or degraded spatial quality.
    \item \textbf{2}: Overall acceptable composition with small artefacts (minor warping, slight mis-scaling, subtle blending issues).
    \item \textbf{3}: High-quality spatial rendering: accurate depth, clean geometry, consistent perspective, and visually stable composition.
\end{itemize}

\textbf{Output only a single integer score in \{1,2,3\}. Do not add any extra text.}

\vspace{0.8em}
\textbf{10. Style}\\[0.1em]

\textbf{(a)Instruction Following}\\
You are a data rater specializing in grading \textbf{style edits}. Two images (before/after) and an instruction will be provided. Evaluate \textbf{Instruction Following} only.

Style edits involve modifying the \emph{artistic, visual, or aesthetic style} of the image, such as painting style, texture style, artistic medium, stroke pattern, visual theme, abstraction level, or global artistic identity.

\textbf{Scoring (1--3)}
\begin{itemize}
    \item \textbf{1}: The intended style change is not reflected at all, or an unrelated visual effect is applied instead.
    \item \textbf{2}: The style direction is correct but incomplete, weak, or mixed with unintended stylistic changes.
    \item \textbf{3}: Perfect: the requested style is expressed clearly, accurately, and exclusively as instructed.
\end{itemize}

\textbf{Output only a single integer score in \{1,2,3\}. Do not add any extra text.}

\vspace{0.5em}

\textbf{(b) Editing Consistency}\\
You are a data rater specializing in grading \textbf{style edits}. Evaluate \textbf{Editing Consistency} only.

\textbf{Scoring (1--3)}
\begin{itemize}
    \item \textbf{1}: Severe stylistic inconsistencies: mismatched textures, uneven strokes, inconsistent themes, or local style breakdowns.
    \item \textbf{2}: Generally consistent but with minor mismatches in texture, stroke density, colour palette, or pattern smoothness.
    \item \textbf{3}: Completely cohesive: style is uniformly and coherently applied across the entire image with no visible inconsistencies.
\end{itemize}

\textbf{Output only a single integer score in \{1,2,3\}. Do not add any extra text.}

\vspace{0.5em}

\textbf{(c) Generation Quality}\\
You are a data rater specializing in grading \textbf{style edits}. Evaluate \textbf{Generation Quality} only.

\textbf{Scoring (1--3)}
\begin{itemize}
    \item \textbf{1}: Poor rendering: artefacts in strokes or textures, muddy colours, broken patterns, or physically incoherent artistic details.
    \item \textbf{2}: Acceptable stylistic rendering but with minor artefacts (slight blur, inconsistent texture density, small colour issues).
    \item \textbf{3}: High-quality stylistic output: clean strokes/textures, consistent colour harmony, and high-fidelity artistic detailing.
\end{itemize}

\textbf{Output only a single integer score in \{1,2,3\}. Do not add any extra text.}

\vspace{0.8em}
\textbf{11. Tone}\\[0.1em]

\textbf{(a)Instruction Following}\\
You are a data rater specializing in grading \textbf{tone edits}. Two images (before/after) and an instruction will be provided. Evaluate \textbf{Instruction Following} only.

Tone edits involve modifying the \emph{emotional, atmospheric, or mood-related characteristics} of the image, including warmth, coolness, dramatic tone, cinematic mood, gloominess, vibrancy, or emotional ambience.

\textbf{Scoring (1--3)}
\begin{itemize}
    \item \textbf{1}: The requested tonal/mood change is not applied, or the edit introduces an unrelated atmosphere.
    \item \textbf{2}: The tonal shift follows the general direction of the instruction but is too weak, too strong, or partially incorrect.
    \item \textbf{3}: Perfect: the intended emotional or atmospheric tone is captured precisely and exclusively as instructed.
\end{itemize}

\textbf{Output only a single integer score in \{1,2,3\}. Do not add any extra text.}

\vspace{0.5em}

\textbf{(b) Editing Consistency}\\
You are a data rater specializing in grading \textbf{tone edits}. Evaluate \textbf{Editing Consistency} only.

\textbf{Scoring (1--3)}
\begin{itemize}
    \item \textbf{1}: Major tonal inconsistencies: uneven colour temperature, patchy atmosphere, contradictory mood signals, or localised tone errors.
    \item \textbf{2}: Mostly consistent but minor mismatches remain, such as slight temperature drift or small inconsistencies in contrast or ambience.
    \item \textbf{3}: Fully consistent: the tonal mood is coherent across the entire image with smooth colour and atmosphere continuity.
\end{itemize}

\textbf{Output only a single integer score in \{1,2,3\}. Do not add any extra text.}

\vspace{0.5em}

\textbf{(c) Generation Quality}\\
You are a data rater specializing in grading \textbf{tone edits}. Evaluate \textbf{Generation Quality} only.

\textbf{Scoring (1--3)}
\begin{itemize}
    \item \textbf{1}: Low-quality tonal rendering: banding, colour blocking, severe noise, unnatural contrast, or degraded ambience.
    \item \textbf{2}: Good overall quality with minor artefacts such as slight noise, mild colour imbalance, or subtle tone gradients issues.
    \item \textbf{3}: High-quality tonal rendering: smooth gradients, clean colours, stable atmosphere, and visually harmonious mood representation.
\end{itemize}

\textbf{Output only a single integer score in \{1,2,3\}. Do not add any extra text.}

\vspace{0.8em}
\textbf{12. Viewpoint}\\[0.1em]

\textbf{(a)Instruction Following}\\
You are a data rater specializing in grading \textbf{viewpoint edits}. Two images (before/after) and an instruction will be provided. Evaluate \textbf{Instruction Following} only.

Viewpoint edits involve modifying the \emph{camera position, angle, orientation, or field of view}, such as shifting perspective (e.g., front view to side view), raising/lowering camera height, rotating the viewpoint, or changing zoom level or focal length.

\textbf{Scoring (1--3)}
\begin{itemize}
    \item \textbf{1}: The viewpoint does not change in the way the instruction requests, or an unrelated transformation is applied.
    \item \textbf{2}: The viewpoint shift is directionally correct but inaccurate in magnitude, orientation, or partially mismatched.
    \item \textbf{3}: Perfect: the camera/viewpoint modification matches the instruction precisely with no unintended distortions.
\end{itemize}

\textbf{Output only a single integer score in \{1,2,3\}. Do not add any extra text.}

\vspace{0.5em}

\textbf{(b) Editing Consistency}\\
You are a data rater specializing in grading \textbf{viewpoint edits}. Evaluate \textbf{Editing Consistency} only.

\textbf{Scoring (1--3)}
\begin{itemize}
    \item \textbf{1}: Severe inconsistencies: distorted geometry, broken perspective, misaligned structures, floating objects, or unrealistic scene deformation.
    \item \textbf{2}: Mostly consistent viewpoint transformation but with minor issues in depth, scale, alignment, or object relationships.
    \item \textbf{3}: Fully coherent: perspective, scale, parallax, and spatial relations all match a consistent new viewpoint.
\end{itemize}

\textbf{Output only a single integer score in \{1,2,3\}. Do not add any extra text.}

\vspace{0.5em}

\textbf{(c) Generation Quality}\\
You are a data rater specializing in grading \textbf{viewpoint edits}. Evaluate \textbf{Generation Quality} only.

\textbf{Scoring (1--3)}
\begin{itemize}
    \item \textbf{1}: Poor rendering of the new viewpoint: warped objects, inconsistent depth cues, incorrect occlusions, blurred or broken structures.
    \item \textbf{2}: Acceptable rendering with mild artefacts (slight geometry distortion, small occlusion issues, minor perspective irregularities).
    \item \textbf{3}: High-quality rendering: clean geometry, stable depth, correct occlusions, and physically coherent viewpoint transformation.
\end{itemize}

\textbf{Output only a single integer score in \{1,2,3\}. Do not add any extra text.}

\vspace{0.8em}
\textbf{13. Background}\\[0.1em]

\textbf{(a)Instruction Following}\\
You are a data rater specializing in grading \textbf{background edits}. Two images (before/after) and an instruction will be provided. Evaluate \textbf{Instruction Following} only.

Background edits involve modifying the \emph{scene environment} behind the main subject, such as changing locations, adding/removing contextual elements, altering scenery type, or replacing the background with a new one.

\textbf{Scoring (1--3)}
\begin{itemize}
    \item \textbf{1}: The background is not changed as requested, or an unrelated region is edited instead.
    \item \textbf{2}: The background change partially matches the instruction but is incomplete, inaccurate, or inconsistent in theme.
    \item \textbf{3}: Perfect: the background is modified exactly as instructed with no unintended changes to the main subject or other elements.
\end{itemize}

\textbf{Output only a single integer score in \{1,2,3\}. Do not add any extra text.}

\vspace{0.5em}

\textbf{(b) Editing Consistency}\\
You are a data rater specializing in grading \textbf{background edits}. Evaluate \textbf{Editing Consistency} only.

\textbf{Scoring (1--3)}
\begin{itemize}
    \item \textbf{1}: Major inconsistencies: mismatched lighting, incorrect depth, broken horizon lines, visible seams, or environmental contradictions.
    \item \textbf{2}: Mostly consistent background replacement but minor mismatches remain (slightly off lighting, subtle blending issues, small perspective errors).
    \item \textbf{3}: Fully consistent: background integrates naturally with correct lighting, depth, colour harmony, and scene logic.
\end{itemize}

\textbf{Output only a single integer score in \{1,2,3\}. Do not add any extra text.}

\vspace{0.5em}

\textbf{(c) Generation Quality}\\
You are a data rater specializing in grading \textbf{background edits}. Evaluate \textbf{Generation Quality} only.

\textbf{Scoring (1--3)}
\begin{itemize}
    \item \textbf{1}: Poor rendering: blurry or warped background, heavy artefacts, broken structures, unrealistic scenery, or degraded composition.
    \item \textbf{2}: Acceptable but with minor artefacts (slight blur, mild noise, small blending imperfections, local geometric inconsistencies).
    \item \textbf{3}: High-quality background rendering: clear details, stable geometry, coherent lighting, and visually realistic integration with the scene.
\end{itemize}

\textbf{Output only a single integer score in \{1,2,3\}. Do not add any extra text.}

\vspace{0.8em}
\textbf{14. Extraction}\\[0.1em]

\textbf{(a)Instruction Following}\\
You are a data rater specializing in grading \textbf{extraction edits}. Two images (before/after) and an instruction will be provided. Evaluate \textbf{Instruction Following} only.

Extraction edits involve \emph{isolating a target object or region} from its surroundings, such as cutting it out, placing it on a clean/transparent background, or clearly separating it from other content, while preserving the target’s appearance.

\textbf{Scoring (1--3)}
\begin{itemize}
    \item \textbf{1}: The requested target is not extracted at all, the wrong region is extracted, or most of the extracted content is unrelated.
    \item \textbf{2}: The intended target is extracted but with noticeable mistakes: parts missing, extra background included, or multiple unintended regions extracted.
    \item \textbf{3}: Perfect: exactly the requested target is extracted, fully included, with no irrelevant regions or missing parts.
\end{itemize}

\textbf{Output only a single integer score in \{1,2,3\}. Do not add any extra text.}

\vspace{0.5em}

\textbf{(b) Editing Consistency}\\
You are a data rater specializing in grading \textbf{extraction edits}. Evaluate \textbf{Editing Consistency} only.

\textbf{Scoring (1--3)}
\begin{itemize}
    \item \textbf{1}: Extraction is badly inconsistent: broken or jagged edges, obvious holes, missing chunks, heavy haloing, or visible mask errors.
    \item \textbf{2}: Generally consistent extraction but with minor artefacts (slightly rough edges, small leftover background pieces, mild halo or fringe).
    \item \textbf{3}: Seamless: clean, stable boundaries; the extracted region looks coherent and well separated, with no visible masking defects.
\end{itemize}

\textbf{Output only a single integer score in \{1,2,3\}. Do not add any extra text.}

\vspace{0.5em}

\textbf{(c) Generation Quality}\\
You are a data rater specializing in grading \textbf{extraction edits}. Evaluate \textbf{Generation Quality} only.

\textbf{Scoring (1--3)}
\begin{itemize}
    \item \textbf{1}: Low-quality rendering of the extracted content: strong blur, compression artefacts, distorted structure, or heavy degradation around the edges.
    \item \textbf{2}: Acceptable quality: the extracted content is mostly clear, with only small artefacts (slight blur, mild noise, minor edge softness).
    \item \textbf{3}: High-quality output: sharp, detailed, and stable appearance of the extracted content, with clean edges and no distracting artefacts.
\end{itemize}

\textbf{Output only a single integer score in \{1,2,3\}. Do not add any extra text.}

\vspace{0.8em}
\textbf{15. Color}\\[0.1em]

\textbf{(a)Instruction Following}\\
You are a data rater specializing in grading \textbf{color edits}. Two images (before/after) and an instruction will be provided. Evaluate \textbf{Instruction Following} only.

Color edits involve changing the \emph{hue, saturation, brightness, palette, or specific colors} of objects or regions, such as making an object red, desaturating a region, or shifting global color tone.

\textbf{Scoring (1--3)}
\begin{itemize}
    \item \textbf{1}: The requested color change is not applied, or the wrong element/color is modified.
    \item \textbf{2}: The color edit follows the instruction directionally but is inaccurate, incomplete, too weak/strong, or applied to extra regions unintentionally.
    \item \textbf{3}: Perfect: the correct region is modified with the exact color change instructed and no unintended color alterations elsewhere.
\end{itemize}

\textbf{Output only a single integer score in \{1,2,3\}. Do not add any extra text.}

\vspace{0.5em}

\textbf{(b) Editing Consistency}\\
You are a data rater specializing in grading \textbf{color edits}. Evaluate \textbf{Editing Consistency} only.

\textbf{Scoring (1--3)}
\begin{itemize}
    \item \textbf{1}: Severe inconsistencies: patchy color application, uneven tone, unnatural gradients, or mismatched lighting effects.
    \item \textbf{2}: Mostly consistent but small inconsistencies exist (slight uneven color areas, minor blending issues, or subtle color leakage).
    \item \textbf{3}: Fully consistent: color changes are smooth, even, and visually coherent across the edited region(s).
\end{itemize}

\textbf{Output only a single integer score in \{1,2,3\}. Do not add any extra text.}

\vspace{0.5em}

\textbf{(c) Generation Quality}\\
You are a data rater specializing in grading \textbf{color edits}. Evaluate \textbf{Generation Quality} only.

\textbf{Scoring (1--3)}
\begin{itemize}
    \item \textbf{1}: Poor color rendering: banding, unnatural hues, heavy noise, severe oversaturation, or damaged structure/texture due to color changes.
    \item \textbf{2}: Acceptable color quality with minor artefacts (slight noise, small tonal imbalance, mild overshoot/undershoot).
    \item \textbf{3}: High-quality color rendering: natural hues, smooth gradients, stable texture preservation, and visually pleasing output.
\end{itemize}

\textbf{Output only a single integer score in \{1,2,3\}. Do not add any extra text.}

\vspace{0.8em}
\textbf{16. Material}\\[0.1em]

\textbf{(a)Instruction Following}\\
You are a data rater specializing in grading \textbf{material edits}. Two images (before/after) and an instruction will be provided. Evaluate \textbf{Instruction Following} only.

Material edits involve changing the \emph{physical material or surface property} of an object, such as turning wood into metal, cloth into leather, plastic into glass, or altering reflectivity, roughness, or texture type.

\textbf{Scoring (1--3)}
\begin{itemize}
    \item \textbf{1}: The object's material does not change as requested, or an unrelated object's material is altered instead.
    \item \textbf{2}: The material change is directionally correct but incomplete, mixed with the old material, inaccurate in appearance, or applied to extra regions.
    \item \textbf{3}: Perfect: the intended object’s material is changed precisely to the requested new material with no unintended spillover.
\end{itemize}

\textbf{Output only a single integer score in \{1,2,3\}. Do not add any extra text.}

\vspace{0.5em}

\textbf{(b) Editing Consistency}\\
You are a data rater specializing in grading \textbf{material edits}. Evaluate \textbf{Editing Consistency} only.

\textbf{Scoring (1--3)}
\begin{itemize}
    \item \textbf{1}: Severe inconsistencies: mixed material cues, incorrect reflections, mismatched texture patches, or physically impossible surface characteristics.
    \item \textbf{2}: Mostly consistent but with minor mismatches in reflectance, texture density, or material detail uniformity.
    \item \textbf{3}: Fully consistent: the new material has uniform texture, correct physical properties, and coherent appearance across the object.
\end{itemize}

\textbf{Output only a single integer score in \{1,2,3\}. Do not add any extra text.}

\vspace{0.5em}

\textbf{(c) Generation Quality}\\
You are a data rater specializing in grading \textbf{material edits}. Evaluate \textbf{Generation Quality} only.

\textbf{Scoring (1--3)}
\begin{itemize}
    \item \textbf{1}: Poor material rendering: muddy textures, broken reflections, unrealistic surface behaviour, or visible artefacts.
    \item \textbf{2}: Acceptable quality with mild artefacts (slightly blurred texture, small reflection inconsistencies, minor physical inaccuracies).
    \item \textbf{3}: High-quality material rendering: sharp, realistic textures, correct reflectance/roughness, and faithful representation of physical properties.
\end{itemize}

\textbf{Output only a single integer score in \{1,2,3\}. Do not add any extra text.}

\vspace{0.8em}
\textbf{17. Beautification}\\[0.1em]

\textbf{(a)Instruction Following}\\
You are a data rater specializing in grading \textbf{beautification edits}. Two images (before/after) and an instruction will be provided. Evaluate \textbf{Instruction Following} only.

Beautification edits involve \emph{enhancing aesthetic qualities} of the subject, such as smoothing skin, whitening teeth, improving symmetry, refining makeup, adjusting facial features subtly, improving lighting on faces, or enhancing general attractiveness while following the specific instruction.

\textbf{Scoring (1--3)}
\begin{itemize}
    \item \textbf{1}: The intended beautification change is not applied, or irrelevant regions are edited instead of the instructed area.
    \item \textbf{2}: The enhancement follows the instructional direction but is too strong, too weak, incomplete, or introduces unintended modifications.
    \item \textbf{3}: Perfect: the requested beautification adjustment is applied precisely, naturally, and only where intended.
\end{itemize}

\textbf{Output only a single integer score in \{1,2,3\}. Do not add any extra text.}

\vspace{0.5em}

\textbf{(b) Editing Consistency}\\
You are a data rater specializing in grading \textbf{beautification edits}. Evaluate \textbf{Editing Consistency} only.

\textbf{Scoring (1--3)}
\begin{itemize}
    \item \textbf{1}: Major inconsistencies: unnatural skin smoothing, warped features, uneven retouching, or mismatched lighting/makeup across regions.
    \item \textbf{2}: Mostly consistent but contains minor inconsistencies (slight over-smoothing, small blending issues, mild asymmetry).
    \item \textbf{3}: Fully consistent: beautification looks natural and uniform with no noticeable artefacts or mismatches across the face/body.
\end{itemize}

\textbf{Output only a single integer score in \{1,2,3\}. Do not add any extra text.}

\vspace{0.5em}

\textbf{(c) Generation Quality}\\
You are a data rater specializing in grading \textbf{beautification edits}. Evaluate \textbf{Generation Quality} only.

\textbf{Scoring (1--3)}
\begin{itemize}
    \item \textbf{1}: Poor visual quality: plastic-like skin, distorted features, harsh artifacts, over-sharpened or overly airbrushed regions.
    \item \textbf{2}: Acceptable beautification with minor artefacts (slight blur, mild noise, subtle inconsistencies in texture or symmetry).
    \item \textbf{3}: High-quality output: natural-looking enhancements, clean textures, realistic lighting, and smooth, refined details.
\end{itemize}

\textbf{Output only a single integer score in \{1,2,3\}. Do not add any extra text.}

\vspace{0.8em}
\textbf{18. Count}\\[0.1em]

\textbf{(a)Instruction Following}\\
You are a data rater specializing in grading \textbf{count edits}. Two images (before/after) and an instruction will be provided. Evaluate \textbf{Instruction Following} only.

Count edits involve modifying the \emph{number of objects} in the scene, such as increasing, decreasing, duplicating, or eliminating instances of a specific object category according to the instruction.

\textbf{Scoring (1--3)}
\begin{itemize}
    \item \textbf{1}: The number of objects does not change as instructed, or the wrong object category is counted/modified.
    \item \textbf{2}: The object count direction is correct but inaccurate—too many, too few, or mixed with unintended removals/additions.
    \item \textbf{3}: Perfect: the object count matches the instruction exactly, with no unintended changes to other objects.
\end{itemize}

\textbf{Output only a single integer score in \{1,2,3\}. Do not add any extra text.}

\vspace{0.5em}

\textbf{(b) Editing Consistency}\\
You are a data rater specializing in grading \textbf{count edits}. Evaluate \textbf{Editing Consistency} only.

\textbf{Scoring (1--3)}
\begin{itemize}
    \item \textbf{1}: Strong inconsistencies: duplicated items look unnatural, obvious cloning artifacts, broken geometry, or mismatched appearance.
    \item \textbf{2}: Mostly consistent but with slight differences in appearance, lighting, or texture across added/removed objects.
    \item \textbf{3}: Fully consistent: duplicated or removed regions integrate naturally with coherent lighting, texture, and geometry.
\end{itemize}

\textbf{Output only a single integer score in \{1,2,3\}. Do not add any extra text.}

\vspace{0.5em}

\textbf{(c) Generation Quality}\\
You are a data rater specializing in grading \textbf{count edits}. Evaluate \textbf{Generation Quality} only.

\textbf{Scoring (1--3)}
\begin{itemize}
    \item \textbf{1}: Low-quality rendering: blurry or warped added objects, deformed duplicates, visible erasure marks, or inconsistent image quality.
    \item \textbf{2}: Acceptable visual quality but with minor issues (small blur, mild noise, slight texture/perspective mismatch).
    \item \textbf{3}: High-quality rendering: added or removed objects are visually clean, realistic, and well integrated into the scene.
\end{itemize}

\textbf{Output only a single integer score in \{1,2,3\}. Do not add any extra text.}

\vspace{0.8em}
\textbf{19. Size}\\[0.1em]

\textbf{(a)Instruction Following}\\
You are a data rater specializing in grading \textbf{size edits}. Two images (before/after) and an instruction will be provided. Evaluate \textbf{Instruction Following} only.

Size edits involve changing the \emph{scale of specific objects or regions}, such as making an object larger/smaller, resizing a person, or adjusting the relative size of elements in the scene according to the instruction.

\textbf{Scoring (1--3)}
\begin{itemize}
    \item \textbf{1}: The requested object is not resized at all, or the wrong element is resized.
    \item \textbf{2}: The resize direction is correct (bigger/smaller) but with inaccurate magnitude, partial resizing, or unintended extra elements affected.
    \item \textbf{3}: Perfect: exactly the instructed object(s) are resized in the correct direction and degree, with no unwanted side effects.
\end{itemize}

\textbf{Output only a single integer score in \{1,2,3\}. Do not add any extra text.}

\vspace{0.5em}

\textbf{(b) Editing Consistency}\\
You are a data rater specializing in grading \textbf{size edits}. Evaluate \textbf{Editing Consistency} only.

\textbf{Scoring (1--3)}
\begin{itemize}
    \item \textbf{1}: Severe inconsistencies: resized object breaks perspective, contact with ground, or relations with nearby objects; obvious stretching or squashing.
    \item \textbf{2}: Mostly consistent but with minor issues in scale, perspective, or contact (slightly off proportions, mild stretching/compression).
    \item \textbf{3}: Fully consistent: resized objects preserve proportions, perspective, and physical relations, and fit naturally into the scene.
\end{itemize}

\textbf{Output only a single integer score in \{1,2,3\}. Do not add any extra text.}

\vspace{0.5em}

\textbf{(c) Generation Quality}\\
You are a data rater specializing in grading \textbf{size edits}. Evaluate \textbf{Generation Quality} only.

\textbf{Scoring (1--3)}
\begin{itemize}
    \item \textbf{1}: Poor visual quality: noticeable pixelation, warped geometry, blurred regions, or strong artefacts due to resizing.
    \item \textbf{2}: Acceptable quality with mild artefacts (slight blur, minor edge issues, small texture inconsistencies).
    \item \textbf{3}: High-quality rendering: sharp details, clean edges, and stable textures on resized objects with no distracting artefacts.
\end{itemize}

\textbf{Output only a single integer score in \{1,2,3\}. Do not add any extra text.}

\vspace{0.8em}
\textbf{20. Text}\\[0.1em]

\textbf{(a)Instruction Following}\\
You are a data rater specializing in grading \textbf{text edits}. Two images (before/after) and an instruction will be provided. Evaluate \textbf{Instruction Following} only.

Text edits involve modifying \emph{any form of visible text} that appears in the scene, including:
posters, billboards, signage, UI/GUI elements, product labels, printed object text, or text on buildings.  
This includes changing the content, wording, symbols, phrasing, or specific characters according to the instruction.

\textbf{Scoring (1--3)}
\begin{itemize}
    \item \textbf{1}: The text does not reflect the instruction, the wrong text region is edited, or no meaningful text update is made.
    \item \textbf{2}: The text partially follows the instruction: some words/characters correct, others incorrect, missing, or incomplete.
    \item \textbf{3}: Perfect: the edited text exactly matches the required content in all relevant regions, with no unintended text changes elsewhere.
\end{itemize}

\textbf{Output only a single integer score in \{1,2,3\}. Do not add any extra text.}

\vspace{0.5em}

\textbf{(b) Editing Consistency}\\
You are a data rater specializing in grading \textbf{text edits}. Evaluate \textbf{Editing Consistency} only.

Since text can appear on a range of surfaces (flat posters, curved objects, UI panels, building facades), evaluate whether the edited text integrates properly with the underlying surface.

\textbf{Scoring (1--3)}
\begin{itemize}
    \item \textbf{1}: Severe inconsistencies: text misaligned, warped incorrectly, wrong perspective, floating above surfaces, mismatched fonts, or broken UI layout.
    \item \textbf{2}: Mostly consistent but with minor issues (slightly off alignment, mild mismatch in font weight/size, minor warping inconsistencies).
    \item \textbf{3}: Fully consistent: text aligns naturally with the surface or UI element, respecting its geometry, curvature, perspective, spacing, and layout style.
\end{itemize}

\textbf{Output only a single integer score in \{1,2,3\}. Do not add any extra text.}

\vspace{0.5em}

\textbf{(c) Generation Quality}\\
You are a data rater specializing in grading \textbf{text edits}. Evaluate \textbf{Generation Quality} only.

\textbf{Scoring (1--3)}
\begin{itemize}
    \item \textbf{1}: Poor rendering: unreadable text, heavy blur, jagged edges, distorted characters, noisy regions, or artefacts around the edited text area.
    \item \textbf{2}: Acceptable quality with minor issues (slight blur, small aliasing, mild artefacts, or small inconsistencies in clarity).
    \item \textbf{3}: High-quality text rendering: crisp, clean, readable characters with smooth edges, stable textures, and no distracting artefacts on any surface.
\end{itemize}

\textbf{Output only a single integer score in \{1,2,3\}. Do not add any extra text.}

\end{tcolorbox}

%%%%%%%%%%%%%%%%%%%%%%%%%%%%%%%%%%%%%%%%%%%%%%%%%%%%%%%%%%%%

\end{document}